\documentclass[mnsc,nonblindrev]{informs3} 

\OneAndAHalfSpacedXI

\usepackage{amssymb}
\usepackage{bbm}
\usepackage{amsfonts}
\usepackage{dsfont}
\usepackage{amsmath}
\usepackage{xspace}
\usepackage[english]{babel}
\usepackage{bm}
\usepackage{epsfig}
\usepackage{enumerate}
\usepackage{mathtools}
\usepackage{cases}
\usepackage[linesnumbered,ruled,vlined]{algorithm2e}

\SetCommentSty{mycommfont}
\usepackage{xcolor}
\DeclarePairedDelimiter\ceil{\lceil}{\rceil}
\DeclarePairedDelimiter\floor{\lfloor}{\rfloor}
\usepackage{comment}
\usepackage{enumitem}
\usepackage{hyperref} 
\usepackage{caption}
\usepackage{subcaption}

\usepackage{natbib}
 \bibpunct[, ]{(}{)}{,}{a}{}{,}%
 \def\bibfont{\small}%
\newcommand{\E}{{\mathds{E}}}

\newcommand{\PP}{\mathds{P}}

\newcommand{\1}{\mathds{1}}

\newcommand{\set}[1]{\ensuremath{\mathcal{#1}}}

\newcommand{\reg}{\text{Reg}}

\newcommand{\policy}[1]{\ensuremath{\pi_{\text{#1}}}}

\definecolor{revision}{RGB}{0,0,255}

\usepackage{commath}

\allowdisplaybreaks
\usepackage{natbib}
 \bibpunct[, ]{(}{)}{,}{a}{}{,}%
 \def\bibfont{\small}%

\ECRepeatTheorems

\TheoremsNumberedThrough     

\EquationsNumberedThrough    

\begin{document}




\TITLE{Learning to Price Supply Chain Contracts against a Learning Retailer}

\ARTICLEAUTHORS{%

\AUTHOR{Xuejun Zhao}
\AFF{Krannert School of Business, Purdue University\\
	\EMAIL{zhao630@purdue.edu}}

\AUTHOR{Ruihao Zhu}
\AFF{SC Johnson College of Business, Cornell University\\
	 \EMAIL{ruihao.zhu@cornell.edu}}

\AUTHOR{William Haskell}
\AFF{Krannert School of Business, Purdue University\\
	 \EMAIL{wbhaskell@gmail.com}}

} 

\ABSTRACT{%

The rise of big data analytics has automated the decision making of companies and increased supply chain agility. In this paper, we study the supply chain contract design problem faced by a data-driven supplier who needs to respond to the inventory decisions of the downstream retailer. Both the supplier and the retailer are uncertain about the market demand and need to learn about it sequentially. The goal for the supplier is to develop data-driven pricing policies with sublinear regret bounds under a wide range of retailer's inventory policies for a fixed time horizon.

To capture the dynamics induced by the retailer's learning policy, we first make a connection to nonstationary online learning by following the notion of variation budget. The variation budget quantifies the impact of the retailer's learning strategy on the supplier's decision-making environment. We then propose dynamic pricing policies for the supplier for both discrete and continuous demand. We also note that our proposed pricing policy only requires access to the support of the demand distribution, but critically, does not require the supplier to have any prior knowledge about the retailer's learning policy or the demand realizations. We examine several well known data-driven policies for the retailer, including sample average approximation, distributionally robust optimization, and parametric approaches, and show that our pricing policies lead to sublinear regret bounds in all these cases. 

At the managerial level, we answer affirmatively that there is a pricing policy with a sublinear regret bound under a wide range of retailer's learning policies, even though she faces a learning retailer and an unknown demand distribution. Our work also provides a novel perspective in data-driven operations management where the principal has to learn to react to the learning policies employed by other agents in the system.

}%


\HISTORY{}

\maketitle

\section{Introduction}


Rapid development of big data analytics has enabled data-driven supply chain management for companies in different industries. According to a survey conducted by \citet{EY} with 212 supply chain leaders from varying sections and company sizes, around $18\%$ of the respondents have already pivoted to big data analytics and $61\%$ of the respondents plan to adopt big data analytics in the next 12-36 months. Big data analytics has also automated the decision-making of companies, and strengthened the agility of the upstream supply chain (e.g., suppliers) to respond to downstream (e.g., retailers and market demand) changes. Motivated by this observation, we study the supply chain contract design problem faced by a data-driven supplier that needs to respond to a downstream retailer who is uncertain about market demand and employs big data analytics tools to make inventory decisions.

We study this problem through the lens of the supplier. The supplier (she) sells a product to a retailer (he) who faces uncertain market demand over a selling horizon of $T$ periods, where the supplier sets a wholesale price (i.e., contract) for the retailer in each period.
Then, the retailer makes a decision on the order quantity accordingly, which also determines the supplier's profit. The retailer does not know the market demand distribution in advance, and may employ a data-driven inventory learning policy that is \emph{unknown} to the supplier. The supplier does not know the market demand distribution either, and she has to sequentially balance the trade-off between exploring the retailer's response to different prices and exploiting profitable prices found so far. This situation may arise in many scenarios. For example, when selling newly introduced products, both the supplier and the retailer are uncertain about the demand of the product and thus have to learn it on the fly.

The supplier's goal is to choose the price to maximize her total profit over the selling horizon. We measure her performance through the notion of regret with respect to a clairvoyant benchmark who has the same information as the retailer (and can predict his orders) and thus chooses the optimal wholesale prices in each period.
This problem is challenging due to the following two sources of uncertainty:
\begin{enumerate}
    \item \textbf{Unknown Market Demand:} In the full information case, when both the supplier and retailer have full knowledge about the market demand distribution, the supplier can directly infer the ordering decisions from the retailer using knowledge about the market demand (assuming the retailer is profit-driven). However, when neither the supplier nor retailer has information about the market demand, the retailer has to learn the demand distribution over time, and the supplier cannot directly infer the retailer's ordering decisions in each period without knowing the retailer's observations and inventory learning policy.
    
    \item \textbf{Uncertain Retailer Inventory Learning Policy:} In addition, uncertainty on the retailer's inventory learning policy makes it particularly challenging to optimize the supplier's profit function, since the retailer can employ a variety of learning policies, and each policy is a mapping from the information received by the retailer to an order quantity. That is, the retailer makes inventory decisions as a response to the supplier's wholesale prices, the observed demand realizations, and his particular inventory learning policy. In this case, even if the supplier \textit{had} known the market demand (yet the retailer still does not know it), inferring the retailer's learning policy from his ordering decisions is not an easy task. 
    
\end{enumerate}

To this end, we ask the following main question: \textit{Does there exist a pricing policy for the supplier with a sublinear regret bound that does not require knowledge of the specific data-driven inventory learning policy used by the retailer?}
If there is such a pricing policy with a sublinear regret bound, then this policy will have no optimality gap with respect to a clairvoyant benchmark's profit asymptotically.

The setting in our paper is novel as well as relevant. The data-driven newsvendor has been studied extensively in the OM literature, but the impact of a data-driven newsvendor on its upstream supplier's decisions has not yet been thoroughly investigated. We approach the supplier's problem by formulating it as a non-stationary online optimization problem.
However, the non-stationarity in our supplier's problem is different than in conventional single agent non-stationary online problems.
In our problem, the non-stationarity lies in the retailer's inventory decisions which depend on his inventory learning policy and the information he receives. In our case, the non-stationarity of the problem is bounded sublinearly in $T$, but the non-stationarity of the retailer's decisions is not necessarily bounded sublinearly in $T$.
In addition, the supplier's online problem has a continuous decision set instead of a finite one. The literature has studied non-stationary online problems with infinitely many decisions, but with the assumption that the objective function is strongly convex or at least continuous. However, we will see that the supplier's profit function in our problem is not necessarily convex/concave or even continuous. Due to these challenges, we need a novel data-driven policy to achieve sublinear regret for the supplier.

\subsection{Our Contributions}

In this paper, we provide an affirmative answer to our main question by giving a data-driven pricing policy that achieves sublinear regret.
We emphasize that our policy does not require the supplier to have any knowledge on the past demand realizations or the retailer's inventory policy. Instead, she only uses her past interactions with the retailer, and knowledge of the support of the demand distribution.

We propose the supplier's policies for both discrete and continuous demand distributions. When the demand distribution is discrete, the supplier's profit function admits a special structure that our policy exploits. When the demand distribution is continuous, this special structure vanishes, but we give a policy that approximates the supplier's profit function and still attains sublinear regret. The details of our contributions in this paper are summarized below:
\begin{enumerate}
    \item Note that the retailer's ordering decisions depend on the supplier's wholesale prices and also on the data-driven inventory learning policies used, this can create non-stationarity in the supplier's decision environment. To capture this effect, we follow the notion of variation budget in the non-stationary bandit (see Section~\ref{sec:literature}) to quantify the difficulty of the supplier's learning problem. 
    With that, even if the retailer switches policies dynamically and/or use a mixture of them, we can encapsulate the impact through the variation budget. 
    Different than prior literature on non-stationary bandits \citep{besbes2014optimal,keskin2021nonstationary,cheung2021hedging}, we define the variation budget in terms of the Kolmogorov distance between the distributions that determine the retailer's inventory decisions. Here, the use of Kolmogorov distance turns out to be natural as it conveniently translates the variation on retailer's inventory decisions to the variation on the supplier's profit functions, and enables the development of pricing policies with provable regret bound for our setting. We also remark that Kolmogorov distance can be upper bounded by many other commonly used distance metrics or divergences, e.g., total variation distance, relative entropy, Helinger distance, Wasserstein distance, etc. \citep{gibbs2002choosing}. 

    \item We propose a pricing policy $\pi_{\text{LUNA}}$ for the supplier that achieves sublinear regret when the market demand distribution is discrete. In this case, the supplier's profit function is discontinuous and non-stationary. 
    In spite of this, we identify special structure in the supplier's profit function to resolve the challenge. We emphasize that our policy does not require any knowledge of the variation budget or the retailer's inventory learning policy. Instead, our policy automatically adjusts to a wide range of retailer policies and variation budgets.
    
    \item When the market demand distribution is continuous, the unique structure in the supplier's profit function vanishes and one cannot directly apply $\pi_{\text{LUNA}}$. 
    To overcome this challenge, we work on an approximation of the supplier's profit function. 
    At a high level, our policy $\pi_{\text{LUNAC}}$ for continuous demand is based on an approximate profit function for the supplier which inherits the desired structure. Then, our previous policy $\pi_{\text{LUNA}}$ for discrete demand can be employed as a sub-routine for $\policy{LUNAC}$.
    
    \item We show that our proposed pricing policy leads to sublinear regret bounds for the supplier under a wide range of retailer inventory policies. We examine: (i) sample average approximation (SAA); (ii) distributionally robust optimization (DRO); and (iii) some parametric approaches (maximum likelihood estimation (MLE), operational statistics, and Bayesian estimation). Under these policies, we compute the respective variation budgets and derive the corresponding regret bounds.

    \item We also conduct numerical experiments to compare our pricing policy with several algorithms from the literature on non-stationary bandits, including the Exp3.S algorithm by \citet{besbes2014optimal}, the deterministic non-stationary bandit algorithm proposed by \citet{karnin2016multi}, and the Master+UCB1 algorithm proposed by \citet{wei2021non} where each price is treated as an arm to pull. We show that our pricing policy has the best performance among all these benchmarks. Our results demonstrate the importance of exploiting structural properties in data-driven operations.
    \item
    At the managerial level, we establish that there is an asymptotically optimal policy for the supplier even though she faces a learning retailer and an unknown (possibly non-stationary) demand distribution
    More generally, our work shows the importance of data-driven operations management where the principal has to learn to react to the learning policies employed by other agents in the system.
    These results also further support the use of wholesale price contracts in practice.
    
\end{enumerate}

\subsection{Related Works}\label{sec:literature}

\noindent\textbf{Contract Design under Uncertainty and MAB:} Supply chain contract design is a longstanding topic, we refer to 
the survey by \citet{cachon2003supply}. In particular, there is an increasing interest in studying contract design under uncertainty \citep{fu2018profit,yu2020robust}. 
We consider the design of wholesale price contracts. There have been efforts in the literature to justify the prevalence of wholesale price contracts in practice \citep{perakis2007price,kalkanci2011contract,yu2020robust}. They suggest that wholesale price contracts are arguably the most natural form of contract for us to investigate when faced with a learning retailer.



Our work lies at the interface between contract design and multi-armed bandit (MAB) problems (see \citet{bubeck2011x} and \citet{lattimore2020bandit}). 
MAB problems have also been extensively studied. In particular, they have been used to model contract design problems. For example, \citet{ho2016adaptive} study repeated principle-agent interactions where the principle offers a contract to induce the efforts of i.i.d. arriving agents.


\noindent\textbf{Dynamic Pricing and Inventory Control:} Dynamic pricing and online revenue management has been studied widely in the OM literature \citep{broder2012dynamic,ferreira2018online,keskin2019dynamic, den2020discontinuous,den2022dynamic,ban2021personalized,keskin2022data, cheung2017dynamic,jia2022online}. Also see \citet{chen2015recent} for an overview of studies on dynamic pricing. More recently, a line of works also integrate inventory control into pricing decisions (see, e.g., \cite{ChenWZ22,ChenLWZ22} and references therein). In this stream of literature, the decision maker is unknown about the demand function, and has to balance the trade-off of learning and earning while dynamically adjusting the pricing and/or inventory decisions. 

Almost all the above works focus exclusively on the stationary demand environment, but in our case, due to the retailer's learning strategy, the decision environment could be dynamically changing. In this regime, \citet{besbes2011minimax, keskin2017chasing, keskin2022data} study dynamic pricing in a non-stationary environment. \citet{keskin2021nonstationary} study the online non-stationary newsvendor problem when the \textit{$L_2-$norm} of the variation in mean demand is bounded. \citet{keskin2022data} study a dynamic joint inventory and pricing problem with perishable products where the price-demand relationship is piecewise stationary. They derive regret bound of $\Tilde{O}(T^{2/3}(\log(T))^{1/2})$ for nonparametric noise distributions and $\Tilde{O}(T^{1/2}(\log(T)))$ for parametric noise distributions, respectively. 

Unlike the previous studies whose goals are to learn the unkown demand functions, the learning in our problem is with respect to the retailer's data-driven inventory learning policies. Furthermore, the non-stationarity in our problem is mostly driven by the learning policies of the self-interested retailer.

\noindent\textbf{Non-stationary Online Learning:}
Many bandit problems are inherently non-stationary. One approach is to model the non-stationarity as a drifting environment, where some metric is used to measure the variation of the environments. The regret analysis is done by restricting to environments with bounded variation \citep{besbes2014stochastic,besbes2015non, wei2016tracking, wei2018abruptly, karnin2016multi,luo2018efficient, cheung2019learning,cheung2021hedging}. Different metrics have been considered, which result in different regret bounds. \citet{besbes2014stochastic} study a $K$-armed bandit problem where the mean reward of the arms is changing. They derive a near-optimal policy with regret $\Tilde{O}((KV)^{1/3}T^{2/3})$ when the \textit{supremum norm} of the change in mean rewards is bounded by a known variation budget $V$. \citet{besbes2015non} study non-stationary stochastic optimization problems where the cost function is convex and the \textit{supremum norm} of the deviations in the cost function in each period is bounded. \citet{chen2019nonstationary} extends the previous work to use the \textit{$L_{p,q}-$variational functional}, which better reflects local spatial and temporal changes in the objective cost functions. These works mostly require the DM to know the variation budget. In order to relax this requirement, \citet{karnin2016multi} propose a restarting algorithm for the $K-$armed bandit problem that restarts whenever a large variation in the environment has been detected by a statistical test.

We build our supplier pricing policy based on the deterministic bandit setting in \citet{karnin2016multi}.
Their algorithm is epoch-based where each epoch consists of an exploration and an exploitation phase. In the exploration phase, the algorithm samples from each arm once and observes the noiseless bandit reward. In the exploitation phase, the algorithm randomly selects an arm to sample, where the sampling distribution is calibrated to balance the trade-off between exploration and exploitation. If the variation of the sampled arm is detected to be above some detection threshold $\set{OV}_B$, then the algorithm starts the next epoch. Otherwise, the algorithm continues the exploitation phase. This algorithm relaxes the assumption that the DM knows the variation budget by sequentially decreasing the detection threshold $\set{OV}_B$ in the exploitation phase. 

In another approach, one can model non-stationarity in a piecewise fashion where the bandit remains stationary in each interval and varies across intervals.
The total number of intervals is bounded by $S$, but the start and end time of each interval is unknown to the DM. Some algorithms have been proposed for known $S$ \citep{auer2002nonstochastic,garivier2011upper,liu2018change,luo2018efficient, cao2019nearly} and unknown $S$ \citep{karnin2016multi,luo2018efficient,auer2018adaptively, auer2019achieving,keskin2022data,chen2019new,besson2019generalized}. We note the difference between this approach for non-stationarity and the first one based on a variation budget. In the first approach, only a constraint on the total variation is imposed and the total number of intervals (where the bandit is stationary) can be linear in $T$ as long as the total variation is bounded. On the other hand, the second approach requires the number of intervals to be bounded, but the variation within intervals can be substantial. Nevertheless, \citet{wei2021non} generalize many reinforcement learning algorithms that work optimally in stationary environments to work optimally in non-stationary environments without any knowledge of the variation budget $V$ or the total number of changes $S$.
We also refer to \citet{zhou2021regime,auer2008near} for a discussion of the Markovian bandit and \citet{chen2020learning} for bandits with seasonality.

The non-stationary bandit is especially relevant to revenue management and dynamic pricing. \citet{cheung2019learning,cheung2021hedging} propose a sliding window upper confidence bound algorithm for the linear bandit where the \textit{Euclidean norm} of the variation in the cost coefficients is upper bounded (but the upper bound is unknown to the DM). Their results cover advertisement allocation, dynamic pricing, and traffic network routing. 

\noindent\textbf{Multi-Agent Learning:} There is a rich literature on multi-agent learning, particularly focusing on online simultaneous games and online Stackelberg games. See \citet{zhang2021multi} for an overview on multi-agent reinforment learning. In particular, \citet{birge2021interfere} consider a platform on which multiple sellers offer products, where sellers' pricing decisions are incentivized by the platform's contract, and both the sellers and the platform do not have full knowledge about the demand price relationship. Unlike ours where the retailer has more information on market demand than the supplier and the latter needs to leverage her interactions with the former to learn the market demand and maximize profit, they focus on the information advantage of the platform over the sellers and study whether and when the platform should release its information advantage.  

\subsection{Organization}
This work is organized as follows. 
In Section~\ref{sec:problem}, we introduce the problem formulation which consists of the supplier's dynamic pricing problem and the class of retailer inventory learning policies. In Section~\ref{sec:stationary}, we present a preliminary analysis of the regret of the supplier's pricing policy when the retailer has full knowledge about the demand distribution. Then, in Section~\ref{sec:retailer-learning} we develop the supplier's pricing policy and its regret upper bound under a learning retailer. We first develop the pricing policy for discrete distributions and then extend it to continuous distributions. In Section~\ref{sec:retailer-learning-examples}, we study several examples of retailer inventory policies under which our pricing policy achieves sublinear regret. In Section~\ref{sec:numerical} we conduct numerical experiments and we conclude the paper in Section~\ref{sec:conclusion}.

\section{Problem Formulation}\label{sec:problem}

Throughout, we let $[N] \triangleq \{1,\ldots,N\}$ be the running index for any integer $N \geq 1$. We adopt the asymptotic notations $O(\cdot)$, $o(\cdot)$, $\Omega(\cdot)$, and $\Theta(\cdot)$ \citep{cormen2022introduction}. When logarithmic factors are omitted, we use $\tilde O(\cdot)$, $\tilde o(\cdot)$, $\tilde \Omega(\cdot)$, and $\tilde \Theta(\cdot)$.
We write `max' instead of `sup' and `min' instead of `inf'. When the optimal solution to the optimization problem does not exist, an ``optimal solution'' means an $\epsilon-$optimal solution for $\epsilon > 0$ arbitrarily small. 

We consider a wholesale price contract between one supplier (she) and one retailer (he) for a single product, where the retailer faces random demand.
Let $c$ be the supplier's unit production cost and $s$ be the retailer's unit selling price.
We use $\mathcal{W} = [0,s]$ to denote the set of admissible wholesale prices, i.e., the supplier cannot sell for more than the retailer selling price (we extend to the case where the supplier's set of admissible decisions $\set{W}$ has finite cardinality in the appendix). Notice that the supplier will gain a negative profit if she sells for less than her production cost $c$, however we allow this possibility since occasionally pricing for less than $c$ may help the supplier explore.

The supplier and retailer interact over a series of time periods indexed by $t \in [T]$ with $T\geq 1$.
Let $\xi_t$ be the random demand in period $t \in [T]$ with support $\Xi\subset \mathbb R_+$.
Let $\mathcal P(\Xi)$ be the set of probability distributions on $\Xi$. We denote the cumulative distribution function (CDF) of demand $\xi_t$ as $F_t \in \mathcal P(\Xi)$ and whose density is $f_t$ (if it exists).
We introduce the shorthand $F_{1:t}\triangleq (F_i)^t_{i=1}$ for $t \in [T]$ for the sequences of true market demand distributions.

In each period $t$, the supplier first offers the retailer the wholesale price $w_t \in \set{W}$. Then, the retailer observes $w_t$, determines his order quantity $q_t$, and the supplier earns profit $\varphi(w_t; q_t) \triangleq (w_t - c) q_t$.
Finally, demand $\xi_t$ is realized and the retailer earns profit $R(q_t;w_t,\xi_t) \triangleq s\min\{q_t,\xi_t\} - w_t q_t$.

The supplier only has access to past wholesale prices and corresponding retailer order quantities. We define $\set{G}_t \triangleq \left\{ (w_i, q_i) \right\}_{i=1}^t$ to be the history of prices and order quantities \textit{by the end of} period $t$ (we let $\set{G}_0 \triangleq \emptyset$).
The supplier's (possibly randomized) pricing policy is a sequence of mappings from $\mathcal G_t$ to the set of probability distributions on $\mathcal W$ (denoted $\mathcal P(\set W)$). We denote the supplier's (possibly randomized) pricing policy by $\pi \triangleq (\pi_t)^T_{t=1}$ where $\pi_1 \in \mathcal P(\set{W})$ and $\pi_t : \set{G}_{t-1} \rightarrow \mathcal P(\set{W})$ for all $t \geq 2$. The wholesale prices under $\pi$ then follow:
\begin{subequations}
\label{equ:supplier-decision rule}
\begin{align}
    w_1 \sim \pi_1, &\\
    w_t \sim \pi_t\left(\set{G}_{t-1} \right), & \quad \forall t \geq 2.
\end{align}
\end{subequations}

Now we characterize the retailer's policy.
Let $\set{H}_t \triangleq \left\{ (w_i, q_i, \xi_i) \right\}^t_{i = 1}$ be the retailer's information \textit{by the end of} period $t$ which consists of the history of wholesale prices, order quantities, and demand realizations up to period $t$ (we simply let $\set{H}_0 \triangleq \emptyset$). Then, the retailer has access to information $\set{H}_{t-1} \cup \{w_{t}\}$ right before his ordering decision is made. 
We let $\mu = (\mu_t)^T_{t =1}$ denote the retailer's (possibly randomized) inventory learning policy, where $\mu_1 : w_1 \rightarrow \set{P}(\Xi)$ and $\mu_t : \set{H}_{t-1} \cup \{w_{t}\} \rightarrow \set{P}(\Xi)$ for $t \in [2, T]$.
The retailer's order quantities under $\mu$ are then determined by:
\begin{subequations}
\label{equ:retailer-decision rule}
\begin{align}
    q_1^\mu \sim \mu_1(w_1), &\\
    q_t^\mu \sim \mu_t\left(\set{H}_{t-1} \cup \{w_{t}\} \right), & \quad \forall t \geq 2.
\end{align}
\end{subequations}
We write $q_t^\mu(w_t; \set{H}_{t-1})$ to denote the retailer's period $t \in [T]$ response to wholesale price $w_t$ under policy $\mu$.

We measure the performance of the supplier's pricing policy $\pi$ in terms of its \textit{dynamic} regret.
We use a clairvoyant benchmark who can predict the retailer's true order quantity given any wholesale price, and the clairvoyant does not necessarily know the true market demand distribution.
For example, if the benchmark has the demand data received by the retailer and knows the policy in use by the retailer, then it can perfectly predict the retailer's order quantity given any wholesale price.

Since the retailer's policy is unknown, we identify a class $\set{M}$ (which depends on $T$ and other model parameters, we will specify $\set{M}$ shortly) of reasonable retailer policies.
If the retailer's policy is allowed to be completely arbitrary, then we cannot always expect to get a sublinear regret for the supplier.
We then consider the supplier's worst-case regret over policies in $\mu\in\set{M}$.
Let 
\begin{equation}\label{equ:optimal-price-benchmark}
    w^*_t \in\arg\max_{w\in\set{W}} (w - c)q_t^\mu(w; \set{H}_{t-1})
\end{equation}
be the benchmark's optimal wholesale price with full knowledge of how the retailer will respond under $q_t^\mu$.
Then, the regret in period $t$ when the supplier prices at $w_t$ is $(w^*_t-c)q_t^\mu(w^*_t; \set{H}_{t-1}) - (w_t-c)q_t^\mu(w_t; \set{H}_{t-1})$.
The overall dynamic regret over the entire planning horizon is then:
$$
\reg(\pi, T) \triangleq \max_{\mu \in\set{M}}\E\left[\sum^T_{t=1} \left( (w^*_t-c)q_t^\mu(w^*_t; \set{H}_{t-1}) - (w_t - c)q_t^\mu(w_t; \set{H}_{t-1} \right) \right],
$$
where the expectation is taken with respect to both the supplier and retailer's possibly randomized policies, and the underlying random demand.
The clairvoyant benchmark in the dynamic regret is able to adjust its strategy dynamically in response to the non-stationarity of the retailer response functions $q_t^\mu(\cdot; \set{H}_{t-1})$.

Dynamic regret is a stronger concept than stationary regret. In the definition of the stationary regret, the clairvoyant benchmark must set the same wholesale price $w^*\in\arg\max_{w\in\set{W} }\sum^T_{t=1}(w-c)q_t^\mu(w; \set{H}_{t-1})$ for the entire planning horizon and the supplier's stationary regret is
$$
\reg_{stat}(\pi, T) \triangleq \max_{\mu \in\set{M}} \E\left[\sum^T_{t=1} \left((w^*-c)q_t^\mu(w^*; \set{H}_{t-1}) - (w_t - c)q_t^\mu(w_t; \set{H}_{t-1}) \right) \right].
$$
It is immediate that the stationary regret is always upper bounded by the dynamic regret.

\subsection{Retailer Model}


Now we present a specific model for how the retailer makes his ordering decisions.
For demand distribution $F \in \mathcal P(\Xi)$, given wholesale price $w$ the retailer's expected profit from ordering $q$ is $\E_{F}\left[R(q;w,\xi)\right]$.
If the retailer believes the demand distribution is $F$, then his best response to wholesale price $w$ is to order
$$
q(w; F) \triangleq \arg\max_{q \geq 0}\E_{F}[R(q;w,\xi)],
$$
or equivalently
\begin{equation}\label{equ:retailer-best response}
    q(w;F) = \min\left\{q: F(q) \geq 1-w/s\right\},
\end{equation}
which maximizes his expected profit with respect to $F$.

In our setting the retailer does not know $F_{1:T}$, and has to implement some inventory learning policy as mentioned before.
We now characterize $\mu$ by supposing that the retailer's ordering decisions are all best responses to a sequence of \textit{perceived distributions}.

\begin{assumption}\label{ass:retailer-learning-1}
Let $\mu$ be the retailer's inventory learning policy. For all $t \in [T]$, there exists a \textit{perceived distribution} $\hat F_t^\mu$ that is adapted to $\set{H}_{t-1} \cup \{w_{t}\}$, such that $q_t^\mu(w_t; \mathcal H_{t-1}) = q(w_t; \hat F_t^\mu)$.
\end{assumption}
\noindent
We let $\hat F_{1:t}^\mu \triangleq (\hat F_1^\mu, \ldots, \hat F_t^\mu)$ for all $t \in [T]$ denote the sequences of perceived distributions.
We say a \textit{stationary retailer} is one who has full knowledge about the demand distribution, and the true distribution is stationary ($\hat F_t^\mu = F_t \triangleq F_0$ for all $t\in[T]$).
Otherwise, we have a \textit{learning retailer}. A learning retailer introduces non-stationarity into the supplier's decision-making environment, even if the true market demand distribution is stationary.  

Assumption~\ref{ass:retailer-learning-1} says that, at any period $t \geq 1$, the retailer's order quantity $q(w_t; \hat F_t^\mu)$ is a best response to some data-driven CDF $\hat F_t^\mu$ that only depends on the information that has been revealed to the retailer up to period $t$ (i.e., $\set{H}_{t-1}\cup w_t$). In other words, the retailer's ordering decisions and thus its entire policy are completely determined by $\hat F_{1:T}^\mu$.
This assumption is without loss of generality, since $q^\mu_t$ must be adapted to $\set{H}_{t-1}\cup w_t$ anyway.
If we are given a rule for constructing $q^\mu_t$ directly from the data, we can always find $\hat F_t^\mu$ for which $q^\mu_t$ is the best response.
There may exist more than one $\hat F_t^\mu$ in period $t$ that satisfies Assumption~\ref{ass:retailer-learning-1}.

We also assume that the retailer knows the support of the true sequence of demand distributions, and thus the retailer will construct $\hat F^\mu_t$ whose support is contained in the support of $F_t$.
\begin{assumption}\label{ass:retailer-learning-support}
    Let $\mu$ be the retailer's inventory policy. For all $t\in[T]$, the support of $\hat F^\mu_t$ is contained in the support of $F_t$. 
\end{assumption}
\noindent
In Section~\ref{sec:retailer-learning-examples} we show that many inventory policies satisfy Assumptions~\ref{ass:retailer-learning-1} and \ref{ass:retailer-learning-support}.
For example, under SAA, the retailer's perceived distribution is the empirical distribution of the observed demand samples.

\subsection{Supplier's Regret}

Since the retailer's order quantities are fully determined by $\hat F_{1:T}^\mu$, the supplier's task of minimizing regret is equivalent to learning $\hat F_{1:T}^\mu$. 
However, it is well known that if $\hat F_{1:T}^\mu$ can vary arbitrarily, then there is no pricing policy that achieves sublinear regret for the supplier.
Our main question only makes sense if we restrict the retailer's inventory policy (or equivalently, the sequence of $\hat F_{1:T}^\mu$) to belong to a reasonable class.
In this case, we expect the variation in $\hat F_{1:T}^\mu$ to be more limited since the retailer accumulates information about the demand distribution incrementally over time, and so their perceived distributions should not change too much from period to period.

We need a metric to quantify this variation in $\hat F_{1:T}^\mu$.
Recall the Kolmogorov distance $d_K$ between CDFs $F$ and $G$ with support on $\Xi \subset \mathbb R_+$ is defined by $d_K(F, G) = \max_{x \in \Xi}|F(x) - G(x)|$.
The variation in the retailer's perceived distributions from period $t$ to period $t+1$ is then $d_K(\hat F_t^\mu, \hat F_{t+1}^\mu)$, and the total variation of the sequence $\hat F_{1:T}^\mu$ is $\sum^{T-1}_{t=1}d_K(\hat F^\mu_t, \hat F^\mu_{t+1})$.
We note that $d_K$ is computable for a wide range of possible perceived distributions. 
In addition, $d_K$ is used in the well-known Kolmogorov-Smirnov test and thus has an intuitive appeal for measuring the similarity between two distributions. Furthermore, $d_K$ can be upper bounded by many other distance metrics or divergences, e.g., total variation distance, relative entropy, Helinger distance, Wasserstein distance, etc. \citep{gibbs2002choosing}. This feature of $d_K$ greatly facilitates connecting the retailer's policy with the regret analysis for the supplier.

We restrict attention to the class of retailer policies for which the total variation of $\hat F_{1:T}^\mu$ (given by $\sum^{T-1}_{t=1}d_K(\hat F_t^\mu, \hat F_{t+1}^\mu)$) is bounded.
Specifically, let $V \geq 0$ (where $V$ is a function of $T$) be a budget for the total variation and define the class of retailer policies:
\begin{multline*}
    \set{M}(V, T)\triangleq \bigg\{\mu : \text{for any }(w_t)^T_{t =1}\in\set{W},
    \text{exists } \hat F^\mu_{1:T} \text{ satisfying Assumptions \ref{ass:retailer-learning-1} and \ref{ass:retailer-learning-support} such that }\\ \sum^{T-1}_{t=1}d_K(\hat F^\mu_t, \hat F^\mu_{t+1})\leq V\bigg\}.
\end{multline*}

The set $\set{M}(V, T)$ includes all $\mu$ such that the total variation of $\hat F^\mu_{1:T}$ does not exceed $V$ for any sequence of wholesale prices. 
Notice that $\set{M}(V, T)$ also implicitly depends on $F_{1:T}$ (since the demand samples are generated from $F_{1:T}$), but we suppress this dependence for brevity. 

With some abuse of the notation, we write the supplier's profit in period $t$ as a function of $w_t$ and $\hat F^\mu_t$ as $\varphi(w_t; \hat F_t^\mu) \triangleq (w_t-c)q(w_t; \hat F_t^\mu)$, when the retailer orders optimally based on the perceived distribution $\hat F_t^\mu$.
The supplier's learning problem can then be framed in terms of the sequence of her profit functions $\{\varphi(w_t; \hat F_t^\mu)\}_{t=1}^T$.

The previous learning literature always considers bounded variation of the profit functions, and proposes learning algorithms specific to this type of variation \citep{besbes2014stochastic,besbes2014optimal,besbes2015non}. In the following example, we show that the variation of $\hat F_{1:T}^\mu$ does not directly translate into the variation of $\{\varphi(w_t; \hat F_t^\mu)\}_{t=1}^T$.
Thus, these previously proposed learning algorithms do not apply to our setting. 

\begin{example}\label{exm:1}
Let $c = 0$ and $s = 1$.
Let $\hat F_t^\mu$ be the CDF of a Bernoulli random variable which takes values $0$ and $1$ with probabilities $p_t$ and $1-p_t$, respectively, for all $t\in[T]$. Let 
\begin{equation*}
    p_t = \begin{cases}
    \frac{1}{2}-\epsilon,&\text{ for $t$ odd},\\
    \frac{1}{2} + \epsilon,&\text{ for $t$ even},
    \end{cases}
\end{equation*}
where $\epsilon = 1/T$. It is straightforward to see that $\sum^{T-1}_{t=1}d_K(\hat F_t^\mu, \hat F_{t+1}^\mu)\leq 2T\epsilon = 2$ but that 
$$
\sum^{T-1}_{t=1}\max_{w\in[c,s]} |\varphi(w; \hat F_t^\mu) - \varphi(w; \hat F_{t+1}^\mu)|= \frac{T}{2}.
$$
We see that even if $\hat F_{1:T}^\mu$ has constant total variation that does not grow in $T$, the sequence of profit functions $(\varphi(w; \hat F_t^\mu))^T_{t=1}$ can have variation that grows linearly in $T$. 
\end{example}

For the specific class $\set{M}(V, T)$ of retailer policies, the overall regret is then:
$$
\reg(\pi, T) \triangleq \max_{\mu \in\set{M}(V, T)}\E\left[\sum^T_{t=1} \left( (w^*_t-c)q(w^*_t;\hat F^\mu_t) - (w_t - c)q(w_t; \hat F^\mu_t) \right) \right],
$$
where the expectation is taken with respect to the supplier's possibly randomized policy and any randomization in $\hat F^\mu_t$ (i.e., $\hat F^\mu_t$ is random in general because it usually depends on the random demand realizations, and the retailer may also use a randomized policy).
For comparison, we define the stationary regret to be:
$$
\reg_{stat}(\pi, T) \triangleq \max_{\mu \in\set{M}(V, T)} \E\left[\sum^T_{t=1} \left((w^*-c)q(w^*;\hat F^\mu_t) - (w_t - c)q(w_t; \hat F^\mu_t) \right) \right].
$$

If the demand distribution is stationary, then intuitively we expect $\hat F_{1:T}^\mu$ to exhibit some type of convergence since the retailer is accumulating more information about the same distribution.
Yet, our results also apply when the true market demand distribution is changing over time. For example, the demand distribution may have a seasonal pattern. In this case, $\hat F_{1:T}^\mu$ may still have bounded variation that is sublinear in $T$.

\section{Stationary Retailer}
\label{sec:stationary}

We first consider the special case of the supplier's problem for a stationary retailer to help us understand the structure of the supplier's profit function. 

\begin{assumption}\label{ass:stationary-true-demand}
    The true market demand distribution is stationary, i.e., $\hat F_t^\mu = F_t = F_0$ for all $t\in[T]$, and the retailer has full knowledge of $F_0$.
\end{assumption}

Since $\set{W}$ is a continuum of allowable prices, this setting is a continuous bandit treating each price as an arm to pull. The literature typically imposes some structure on the DM's objective function, e.g., convexity or unimodality, Lipchitz continuity  or Holder continuity, etc. (see \citet{kleinberg2013bandits,besbes2015non}). However, the supplier's profit function is not necessarily continuous in our setting. Nevertheless, we will show that we can find pricing policies that have sublinear regret bounds for general demand distributions.



We make the following boundedness assumption on the demand distribution to continue.

\begin{assumption}\label{ass:retailer-learning-3}
The sequence $F_{1:T}$ has bounded support on $\Xi = [0,\bar\xi]$ for $0 < \bar\xi < \infty$, and both the retailer and the supplier know $\bar\xi$.
\end{assumption}

Under Assumption~\ref{ass:retailer-learning-3}, we can always relate the values of the supplier's profit function at $w_t$ and $w_t'$. Without loss of generality, suppose $w_t'<w_t$, then for any demand distribution $F$ we have
\begin{equation}\label{equ:supplier-profit-property}
\begin{split}
    \varphi(w_t; F) - \varphi(w'_t; F)
    =&\,(w_t-c)q(w_t; F) - (w'_t-c)q(w'_t; F)\\
    \leq &\,(w_t-c)q(w_t; F) - (w'_t-c)q(w_t; F)\\
    \leq &\,(w_t - w'_t)\bar\xi.
\end{split}
\end{equation}
This inequality holds regardless of whether the true distribution is discrete or continuous. In particular, we can discretize $\set{W}$ with a finite set of prices, and then bound sub-optimality of this discretization using Eq.~\eqref{equ:supplier-profit-property}.

We consider the following simple pricing policy for the stationary retailer that we denote by $\pi_{\text{stat}}$. The supplier first discretizes $\set{W}$ into $\ceil{\sqrt{T}}$ equally sized intervals, and then takes $\mathcal{\bar W}_{\ceil{\sqrt{T}}}$ to be the wholesale prices at the breakpoints of these intervals. In the first $\ceil{\sqrt{T}}$ periods, the supplier sets each price in $\mathcal{\bar W}_{\ceil{\sqrt{T}}}$ once and collects the corresponding profit. Let $w^*_{stat} \in \arg\max_{w\in\mathcal{\bar W}_{\ceil{\sqrt{T}}}}\varphi(w; F_0)$ be the wholesale price in $\mathcal{\bar W}_{\ceil{\sqrt{T}}}$ with the highest profit.
The supplier then sets $w_t = w^*_{stat}$ for all remaining periods $t = T-\ceil{\sqrt{T}} + 1, \ldots, T$.

Since $\hat F_{1:T}$ is stationary in this case, the supplier's profit function does not change from period to period.
The dynamic and stationary regret coincide and are:
$$
\reg(\pi_{\text{stat}}, T) = \sum^T_{t=1}\E\left[(w^*-c)q(w^*;F_0) - (w_t - c)q(w_t; F_0)\right],
$$
where $w^* \in \max_{w\in\set{W}}(w-c)q(w;F_0)$.
Our proposed policy $\pi_{\text{stat}}$ gives the supplier $O(\sqrt{T})$ regret.
We emphasize that this result holds for both discrete and continuous demand distributions.

\begin{theorem}\label{thm:bound-stationary retailer}
Suppose Assumptions~\ref{ass:stationary-true-demand} and \ref{ass:retailer-learning-3} hold.
For all $T \geq 1$, we have $\reg(\pi_{\text{stat}}, T) = O(\sqrt{T})$.
\end{theorem}

\section{Learning Retailer}\label{sec:retailer-learning}

We now discuss the supplier's problem with a learning retailer. The supplier's profit function is generally multi-modal, and now it is also changing shape since the retailer is updating his perceived distributions. Towards solving the supplier's problem in this setting, we start with the case of a discrete demand distribution and then extend to the continuous case via an approximation argument. In both cases, we find a pricing policy for the supplier with a sublinear regret bound.

\subsection{Discrete Demand Distributions}\label{sec:retailer-learning-discrete}

We first assume that the demand in all time periods has common finite support.

\begin{assumption}\label{assu:demand_discrete}
The sequence $F_{1:T}$ has common support on $M$ points: $y_M(\triangleq\bar\xi)>y_{M-1}>y_{M-2}>\cdots>y_{1}\geq 0$.
The support $\set{Y}_M \triangleq \{y_m\}_{m \in [M]}$ is known to both the supplier and the retailer.
\end{assumption}
\noindent
Combined with Assumption~\ref{ass:retailer-learning-support}, the sequence of perceived distributions $\hat F_{1:T}^\mu$ also has support on $\set{Y}_M$ under this assumption. Let $p_{t,m} \triangleq \hat F_t^\mu(y_m)$ for $m \in [M]$ denote the values of the retailer's perceived distribution in period $t \in [T]$. The retailer's order quantity given by Eq.~\eqref{equ:retailer-best response} is then
\begin{equation}\label{equ:retailer-order-stationary}
    \begin{split}
       q_t = \begin{cases}
       y_1, &\text{ if } 0\leq 1 - w_t/s \leq p_{t,1},\\
        y_m,&\text{ if } p_{t,m-1} < 1- w_t/s \leq p_{t,m}, \text{ for }m\in\{2, \ldots, M\}.
       \end{cases} 
    \end{split}
\end{equation}
We see that the retailer's order quantity is a piecewise constant function of the wholesale price $w_t$. According to Eq.~\eqref{equ:retailer-order-stationary}, the retailer's order quantity will always be in the support of the discrete distribution and satisfy: 
\begin{equation}\label{equ:retailer-learning-order}
q_t\in \set{Y}_M.
\end{equation}
Consequently, the supplier's profit function in period $t$ is:
\begin{equation}\label{equ:retailer-learning-profit}
\varphi(w_t; \hat F_t^\mu) = \begin{cases}
(w_t - c)y_1, &\text{ if } 0 \leq 1 - w_t/s \leq p_{t,1},\\
(w_t - c)y_m, &\text{ if } p_{t, m-1}< 1 - w_t/s \leq p_{t,m} \text{ for }m \in[M].
\end{cases}
\end{equation}
We see it is a piecewise linear function of $w_t$, with discontinuities at the breakpoints $\{s(1-p_{t,m})\}_{m\in[M]}$.

We propose a pricing policy for the supplier called $\pi_{\text{LUNA}}$, where LUNA stands for \textbf{L}earning \textbf{U}nder a \textbf{N}on-stationary \textbf{A}gent.
It is based on the deterministic bandit algorithm proposed in \citet{karnin2016multi}.
$\pi_{\text{LUNA}}$ works by taking advantage of the special structure of the supplier's profit function, as characterized in Eq.~\eqref{equ:retailer-learning-profit}.
In particular, the performance of the pricing policy depends on accurate estimation of the probabilities $\bm p_t \triangleq \left(p_{t,1}, \ldots, p_{t,M}\right)$.
$\pi_{\text{LUNA}}$ indirectly estimates $\bm p_t$ by observing the retailer's order quantity and the supplier's profit through Eq.~\eqref{equ:retailer-learning-profit}.
A good policy is expected to maintain a somewhat accurate estimate of $\bm p_t$, but must also hedge against large variation in $\bm p_t$.
If $\bm p_t$ has varied a lot, then the optimal wholesale price may be different, and a large regret will be incurred if the supplier fails to adapt. 

The full details of $\pi_{\text{LUNA}}$ are presented in Algorithm~\ref{alg:retailer-learning}.
It consists of multiple epochs, where each epoch consists of an exploration phase followed by an exploitation phase.
We let $i \geq 1$ index epochs, but we usually omit dependence on $i$ except when necessary since each epoch follows the same pattern.
Let $\tau_{i+1}^0$ denote the last period of epoch $i \geq 1$ (where $\tau_1^0 = 0$). Then, epoch $i \geq 1$ covers periods $t \in [\tau_i^0 + 1, \ldots, \tau_{i+1}^0]$.
Each time $\pi_{\text{LUNA}}$ starts the exploration phase of a new epoch, it discards all previous information and estimates $\bm p_t$ from scratch
(since $\bm p_t$ is non-stationary, it is the estimate for some particular period $t$). It also constructs a set of exploratory wholesale prices, tries each price once, and records the optimal exploratory price which led to the highest observed supplier profit.

Once it has an initial estimate of $\bm p_t$ and an optimal exploratory wholesale price has been found, $\pi_{\text{LUNA}}$ will enter the exploitation phase.
It first constructs a new set of wholesale prices for the exploitation phase. Then, in each period, a wholesale price is drawn randomly from this set according to some distribution which balances the exploration-exploitation trade-off. Most of the time, $\pi_{\text{LUNA}}$ prices at the nearly optimal wholesale price found in the exploration phase, while also occasionally detecting whether the previously identified optimal wholesale price is no longer optimal. If that is the case, then $\pi_{\text{LUNA}}$ quantifies a lower bound on the variation of $\hat F_t^\mu$ in the current epoch and begins the next epoch.

\subsubsection{Exploration Phase of $\pi_{\text{LUNA}}$}
The goal of the exploration phase is to obtain an initial estimate of $\bm p_t$ and to find the optimal wholesale price corresponding to this initial estimate.
As a first step, $\pi_{\text{LUNA}}$ discretizes $\set{W}$ into $K+1$ equal-length intervals with $K \geq 1$ equally spaced wholesale prices (where $K$ is an input parameter to be specified later, which is the same for every epoch). We then let $\set{\bar W}_K \triangleq \{\bar w_k\}^K_{k =1}$ be the set of exploratory prices where:
\begin{equation}\label{equ:luna-discrete-wk}
    \bar w_k \leftarrow (k-1)\frac{(s-c)}{K} + c,\, \forall k\in[K].
\end{equation}
Then in each period $\tau^0_i + k$ for all $k \in [K]$, $\pi_{\text{LUNA}}$ sets the wholesale price $\bar w_k$ and the corresponding retailer order quantity is $q(\bar w_k; \hat F_{\tau^0_i + k}^\mu)$.
Upon setting $\bar w_k$, $\pi_{\text{LUNA}}$ earns profit $\varphi_k$ given by:
\begin{equation}\label{equ:luna-discrete-varphik}
\varphi_k \leftarrow (\bar w_k - c)q(\bar w_k; \hat F_{\tau^0_i + k}^\mu).
\end{equation}
Let $k^*\in \arg\max_{k\in[K]}\varphi_{k}$, so $\bar w_{k^*}$ is the wholesale price that maximizes the observed profit among $\set{\bar W}_K$. By Eq.~\eqref{equ:retailer-learning-order}, we must have $q(\bar w_{k^*}; \hat F_{\tau^0_i + k^*}^\mu) \in \set{Y}_M$, so we let $m^* \in [M]$ be such that $y_{m^*} \triangleq q(\bar w_{k^*}; \hat F_{\tau^0_i + k^*}^\mu)$. With $\bar w_{k^*}$ and $y_{m^*}$ in hand, we begin the exploitation phase of epoch $i$.

\subsubsection{Exploitation Phase of $\pi_{\text{LUNA}}$}


If the retailer is stationary, then $\bar w_{k^*}$ will remain nearly optimal for the rest of the planning horizon (this is exactly the supplier's pricing policy for a stationary retailer, see Section~\ref{sec:stationary}).
However, as the retailer is also learning the demand distribution, we expect $\hat F_t^\mu$ to vary. If $\hat F_t^\mu$ has varied a lot and the supplier still prices at $\bar w_{k^*}$, then she is likely to suffer a large regret. The supplier's pricing policy has to balance between exploitation (stick to the optimal $\bar w_{k^*}$ found in the exploration phase) and exploration (hedge against the risk that $\hat F_t^\mu$ has changed a lot since the exploration phase). 

We show that this balance can be achieved during the exploitation phase by randomly choosing from a carefully chosen finite set of prices. This set of prices is constructed based on the structure of the supplier's profit function through Eq.~\eqref{equ:retailer-learning-profit}, and it will be different for each period $t$.
We first construct this set of prices, and then explain the intuition behind it.

We allow the pricing policy in period $t$ to have a margin of sub-optimality $\Delta_t > 0$ compared with $\varphi_{k^*}$, the highest profit observed in the exploration phase.
The sequence $\{\Delta_t\}_{t \geq 1}$ will be chosen to be decreasing in $t$ (in particular, we will take $\Delta_t = O(\sqrt{1/t})$).
Then, in each period $t$, we construct a set of prices based on $\Delta_t$ to sample from.

There are two cases where $\bar w_{k^*}$ becomes sufficiently sub-optimal to end the current epoch, either: (i) the supplier's profit at some $w\neq \bar w_{k^*}$ has increased a lot since the exploration phase; or (ii) the supplier's profit at $\bar w_{k^*}$ has decreased a lot since the exploration phase.
We discuss the details of these two cases separately.

\paragraph{Case one:}

In the first case, some $w\in\set{W}$ with $w\neq \bar w_{k^*}$ now earns greater profit for the supplier than $\bar w_{k^*}$. This $w$ can be an arbitrary member of $\set{W}$ as long as $w\neq \bar w_{k^*}$. However, we cannot check every price in $\set{W}$, so we construct a specialized finite set of prices to check as follows.

We recall that $q^\mu_t\in \set{Y}_M$ holds for all $t\in[T]$ under Assumption~\ref{assu:demand_discrete}, so we ask the question: \textit{Suppose the retailer's order quantity is $y_m \in \set{Y}_M$ for some $m\in[M]$, then what wholesale price (denoted by $w^t_m$) would give the supplier a profit that is equal to $\varphi_{k^*} + \Delta_t$?} If the retailer's order quantity $q(w^t_m; \hat F_t^\mu)$ under $w^t_m$ turns out to be larger than (smaller than) $y_m$, then $w^t_m$ will give a higher (lower) profit than $\varphi_{k^*} + \Delta_t$. 
We construct a set of wholesale prices in this way corresponding to each $y_m \in \set{Y}_M$.
In period $t$, for each $m \in [M]$, we set the corresponding wholesale prices according to:
\begin{equation}\label{equ:luna-discrete-wm}
    (w^t_m - c)y_m = \varphi_{k^*} + \Delta_t + \frac{y_m s}{K},\text{ which gives }w^t_m \triangleq \Big(\varphi_{k^*} + \Delta_t +  \frac{y_m s}{K} \Big)/y_m + c,
\end{equation}
where the term $\frac{y_m s}{K}$ is introduced to account for the error introduced by discretizing $\set{W}$ to $\set{\bar W}_K$. 

If the supplier prices at $w^t_m$ in period $t$, and if $q(w^t_m; \hat F_t^\mu) \geq y_m$, then $\bar w_{k^*}$ is no longer nearly optimal since the optimality gap now exceeds $\Delta_t$ (i.e., we have $(w^t_m - c)y_m\geq \varphi_{k^*} + \Delta_t$).
We summarize this discussion in Lemma~\ref{lem:retailer-learning-wm}.
\begin{lemma}\label{lem:retailer-learning-wm}
If $q(w^t_m; \hat F_t^\mu)\geq y_m$, then $p_{t, m-1}\leq 1-w^t_m/s$ and $\varphi(w^t_m; \hat F_t^\mu) \geq (w^t_m - c)y_m = \varphi_{k^*} + \Delta_t + \frac{y_m s}{K}$.
Otherwise, if $q(w^t_m; \hat F_t^\mu)< y_m$, then $p_{t, m-1}\geq 1-w^t_m/s$ and $\varphi(w^t_m; \hat F_t^\mu) < (w^t_m - c)y_m = \varphi_{k^*} + \Delta_t + \frac{y_m s}{K}$.
\end{lemma}
\noindent
In addition, if $q(w^t_m; \hat F_t)\geq y_m$, then $\hat F_t^\mu$ has varied a lot since the exploration phase.
\begin{lemma}\label{lem:epoch-variation-1}
If $q(w^t_m; \hat F_t^\mu)\geq y_m$, then $\sum_{j\in[\tau^0_i+1, t-1]}d_{K}(\hat F_j^\mu, \hat F_{j+1}^\mu) \geq \Delta_t/(s\, \bar\xi)$.
\end{lemma}





\paragraph{Case two:}

In the second case, the supplier's profit from pricing at $\bar w_{k^*}$ has decreased a lot since the exploration phase.
This can happen if the retailer's order quantity $q(\bar w_{k^*}; \hat F_{t}^\mu)$ in period $t$ during the exploitation phase is much smaller than the order quantity $q(\bar w_{k^*}; \hat F_{\tau^0_i + k^*}^\mu)$ observed during the exploration phase. 
Since $\bar w_{k^*}$ is the optimal price found during the exploration phase, the policy should price frequently at $\bar w_{k^*}$ to exploit what is best. However, pricing at $\bar w_{k^*}$ does not give useful information about the variation in $\hat F_t$. Even if $\hat F_t$ has only varied by a small amount, the profit at $\bar w_{k^*}$ can still change drastically (recall Example~\ref{exm:1}). If we restart the epoch each time the profit at $\bar w_{k^*}$ has decreased a lot, we will end up with too many epochs and a high overall regret.

Therefore, instead of pricing at $\bar w_{k^*}$, we determine a surrogate price $w^t_0$ which achieves two purposes: (i) the profit at $w^t_0$ is not much lower than the profit at $\bar w_{k^*}$, so we can still exploit the optimality of $\bar w_{k^*}$ from the exploration phase (see Eq.~\eqref{equ:lower-bound-w0}); and (ii) unlike $\bar w_{k^*}$, when the profit at $w^t_0$ is sufficiently low, we can quantify a lower bound on the variation of $\hat F^\mu_t$ (see Lemma~\ref{lem:epoch-variation-2}) and correctly restart the epoch. 
We define the surrogate price $w^t_0$ in period $t$ to satisfy:
\begin{equation*}
     (w^t_0 - c)y_{m^*} = \varphi_{k^*} - \Delta_t \text{ and }w^t_0 \geq 0, \text{ otherwise } w^t_0 = 0,
\end{equation*}
which gives
\begin{equation}\label{equ:luna-discrete-w0}
    w^t_0\triangleq \max\{\bar w_{k^*} - \Delta_t/y_{m^*}, 0\}.
\end{equation}
Note we require $w^t_0\geq 0$ instead of $w^t_0\geq c$. By allowing $w^t_0 < c$, the policy is able to detect variation of $\hat F_t^\mu$ that would otherwise not be detected. 

By Eq.~\eqref{equ:supplier-profit-property}, the difference in profit between pricing at $w^t_0$ and $\bar w_{k^*}$ is lower bounded by:
\begin{equation}\label{equ:lower-bound-w0}
    \varphi(w^t_0; \hat F_t^\mu) - \varphi(\bar w_{k^*}; \hat F_t^\mu) \geq -\Delta_t.
\end{equation}
We make the following inferences based on the surrogate price.
\begin{lemma}\label{lem:retailer-learning-w0}
If $q(w^t_0; \hat F_t^\mu)\geq y_{m^*}$, then $p_{t, m^*-1}\leq 1-w^t_0/s$ and $\varphi(w^t_0; \hat F_t^\mu) \geq (w^t_0 - c)y_{m^*} \geq \varphi_{k^*} - \Delta_t$. Otherwise, if $q(w^t_0; \hat F_t^\mu)< y_{m^*}$, then $p_{t, m^*-1}\geq 1 - w^t_0/s$ and $\varphi(w^t_0; \hat F_t^\mu) < (w^t_0 - c)y_{m^*} = \varphi_{k^*} - \Delta_t$.
\end{lemma}
\noindent
If the retailer's order quantity satisfies $q(w^t_0; \hat F_t^\mu) < y_{m^*}$, then we know that $\hat F_t^\mu$ has varied a lot since the beginning of epoch $i$.
\begin{lemma}\label{lem:epoch-variation-2}
If $q(w^t_0; \hat F_t^\mu)< y_{m^*}$, then $\sum_{j\in[\tau^0_i+1, t-1]}d_{K}(\hat F_j^\mu, \hat F_{j+1}^\mu)\geq \Delta_t/(s\, \bar\xi)$.
\end{lemma}

\subsubsection{Algorithm and regret bound}

In each period $t$, we construct the set of wholesale prices $\{w^t_0, w_1^t, \ldots, w_M^t\}$, and $\policy{LUNA}$ will randomly sample from $\{w^t_0, w_1^t, \ldots, w_M^t\}$ according to a distribution that is changing over time.
Based on the discussion of the previous two cases, the exploitation phase continues until $q(w^t_m; \hat F_t^\mu)\geq y_m$ for some $m\in[M]$, or $q(w^t_0; \hat F_t^\mu) < y_{m^*}$. In both cases, $\hat F_t^\mu$ is guaranteed to have varied a lot since the exploration phase, and $\pi_{\text{LUNA}}$ starts the next epoch.
Let $\set{U}([M])$ denote the uniform distribution on $\{1, 2, \ldots, M\}$.

\begin{algorithm}[ht]
Input: Time horizon $T$, supplier production cost $c$, retailer selling price $s$, support $\set{Y}_M$, grid size $K$;\\
	Update current period $t\leftarrow 1$;\\
	Set epoch $i \leftarrow 1$ and $\tau^0_1\leftarrow 0$;\\
	\For{epoch $i = 1, 2, \cdots$}{
	\textbf{Exploration:}\\
	   Price at $\bar w_k$ (see Eq.~\eqref{equ:luna-discrete-wk}) and observe $\varphi_k$ (see Eq.~\eqref{equ:luna-discrete-varphik}) for the first $K$ periods in epoch $i$; \\
	   Let $k^*\in\arg\max_{k\in[K]}\varphi_k$ and $m^*$ be such that $y_{m^*} = q\left(\bar w_{k^*}; \hat F_{\tau^0_i + k^*}^\mu \right)$;\\
	   \textbf{Exploitation:}\\
	   In period $t$, set $\Delta_t \leftarrow \sqrt{M/(t-\tau^0_i)}$;\\
	   Compute prices $w^t_m$ for $m\in[M]$ and $w^t_0$ according to Eq.~\eqref{equ:luna-discrete-wm} and Eq.~\eqref{equ:luna-discrete-w0}, respectively;\\
	   Select wholesale price $w_t \leftarrow w_{m_t}^t$ according to the distribution
	   $$
        m_t = \begin{cases}
        0, &\text{ w.p. } 1 - \sqrt{\frac{M}{t-\tau^0_i}},\\
        \set{U}([M]), &\text{ w.p. } \sqrt{\frac{M}{t-\tau^0_i}},
        \end{cases}
	   $$\\
	   Observe retailer's order $q(w_t; \hat F_t^\mu)$;\\
	   \If{
	   $q(w_{m_t}^t; \hat F_t^\mu)\geq y_{m_t}$ for $m_t\in[M]$ or $q(w_{m_t}^t; \hat F_t^\mu)< y_{m^*}$ for $m_t = 0$
	   }{
        $\tau^0_{i+1} \leftarrow t$ and start the next epoch $i\leftarrow i+1$  ;
	   }
	}
	\caption{\textbf{L}earning \textbf{U}nder \textbf{N}on-stationary \textbf{A}gent (LUNA)}
	\label{alg:retailer-learning}
\end{algorithm}

Theorem~\ref{thm:retailer-learning} upper bounds the regret of Algorithm~\ref{alg:retailer-learning} as a function of $V$ and $K$ (notice that $\pi_{\text{LUNA}}$ does not need to know $V$, but $K$ is an input). When in addition $V$ is known, the decision maker can choose $K$ optimally as a function of $V$ to minimize the regret. 

\begin{theorem}\label{thm:retailer-learning}
Suppose Assumption~\ref{sec:retailer-learning} holds. 

(i) For all $K \geq 1$, $\reg(\pi_{\text{LUNA}} ,T) = \Tilde{O}(\bar\xi^{\frac{4}{3}}V^{\frac{1}{3}}M^{\frac{1}{3}}T^{\frac{2}{3}} + \frac{\bar\xi\, T}{K} + \bar\xi^{\frac{5}{3}}KV^{\frac{2}{3}}M^{-\frac{1}{3}}T^{\frac{1}{3}})$.

(ii) If the supplier knows $V$, then $K$ can be chosen optimally as $K^* = \ceil{T^{\frac{1}{3}}V^{-\frac{1}{3}}\bar\xi^{-\frac{1}{3}}}$, and the minimized regret is $\reg(\pi_{\text{LUNA}}, T) = \Tilde{O}(\bar\xi^{\frac{4}{3}}V^{\frac{1}{3}}M^{\frac{1}{3}}T^{\frac{2}{3}})$.

(iii) If the supplier does not know $V$, then $K$ can be chosen obliviously as $\hat K = \ceil{\bar\xi^{-\frac{1}{3}}T^{\frac{1}{3}}}$, and the regret is $\reg(\pi_{\text{LUNA}}, T) = \Tilde{O}(\bar\xi^{\frac{4}{3}}V^{\frac{1}{3}}M^{\frac{1}{3}}T^{\frac{2}{3}}
    + \bar\xi^{\frac{4}{3}}V^{\frac{2}{3}}M^{-\frac{1}{3}}T^{\frac{2}{3}})$.
\end{theorem}

According to Theorem~\ref{thm:retailer-learning}, using our pricing policy $ \policy{LUNA}$, the supplier can achieve a sublinear regret bound if $V = \tilde o(T)$ and if $V = \tilde o(\sqrt{T})$ when the supplier knows $V$ and does not know $V$ respectively.

\subsection{Proof Outline of Theorem~\ref{thm:retailer-learning} ($\pi_{\text{LUNA}}$)}

Here we overview the proof of Theorem~\ref{thm:retailer-learning}, all detailed expressions and derivations referenced here appear in Appendix~\ref{appendix:learning}.
We first do the regret analysis for a single epoch $i$ (which consists of periods $t \in [\tau_i^0 + 1, \tau_{i+1}^0]$), and then assemble these into an overall regret bound. To begin, we decompose the regret in epoch $i$ into:
\begin{equation*}
    \begin{split}
        \sum^{\tau^0_{i+1}}_{t=\tau^0_i+1}\bigg\{\varphi(w^*_t; \hat F_t^\mu) - \varphi(w_t; \hat F_t^\mu)\bigg\}
         = \sum^{\tau^0_{i+1}}_{t=\tau^0_i+1}\bigg\{\varphi(w^*_t; \hat F_t^\mu) - \varphi(w^t_0; \hat F_t^\mu)\bigg\} + \bigg\{\varphi(w^t_0; \hat F_t^\mu) - \varphi(w_t; \hat F_t^\mu)\bigg\},
    \end{split}
\end{equation*}
where the first part $\reg_i^c(\policy{LUNA}) \triangleq \sum^{\tau^0_{i+1}}_{t = \tau^0_i + 1} \{ \varphi(w^*_t; \hat F_t^\mu) - \varphi(w^t_0; \hat F_t^\mu) \}$ (the superscript `c' is for `clairvoyant') is the regret incurred by always pricing at $w^t_0$ compared to the clairvoyant benchmark, and the second part $\reg_i^0(\policy{LUNA}) \triangleq \sum^{\tau^0_{i+1}}_{t = \tau^0_i + 1} \{ \varphi(w^t_0; \hat F_t^\mu) - \varphi(w_t; \hat F_t^\mu) \}$ (the superscript `0' corresponds to the surrogate price) is the regret incurred compared with the benchmark of always pricing at the surrogate price $w^t_0$. We analyze these two parts separately. 

\paragraph{Part I of the regret}
To upper bound $\reg_i^c(\policy{LUNA})$, we define the following subset of periods of epoch $i$:
\begin{equation*}
\set{E}^i \triangleq \left\{t\in[\tau^0_i+\max\{M+2, K\} + 1, \tau^0_{i+1}]: q(w^t_m; \hat F_t^\mu) < y_m, \,\forall m\in[M], \text{ and } q(w^t_0; \hat F_t^\mu)\geq y_{m^*}\right\}.
\end{equation*}
If $t\in \set{E}^i$, then $\hat F_t^\mu$ has not varied a lot within epoch $i$ and pricing at $w^t_0$ remains nearly optimal.
On the other hand, if $t\notin \set{E}^i$, then pricing at $w^t_0$ is no longer nearly optimal either because the profit at $w^t_0$ has gone down, or the profit at some $w^t_m \ne w^t_0$ has gone up. We can further decompose
\begin{equation*}
\begin{split}
    \varphi(w^*_t; \hat F_t^\mu) - \varphi(w^t_0; \hat F_t^\mu) &= \left(\varphi(w^*_t; \hat F_t^\mu) - \varphi(w^t_0; \hat F_t^\mu)\right)\1(t\in\set{E}^i) + \left(\varphi(w^*_t; \hat F_t^\mu) - \varphi(w^t_0; \hat F_t^\mu)\right)\1(t\notin\set{E}^i),\\
\end{split}
\end{equation*}
and then upper bound these expressions separately.
First we upper bound the regret $\varphi(w^*_t; \hat F_t^\mu) - \varphi(w^t_0; \hat F_t^\mu)$ for periods $t\in\set{E}^i$.
The next result takes effect after period $\tau_i^0 + \max\{M+2, K\}$ and only applies to the exploitation phase.

\begin{lemma}\label{lem:retailer-learning-2}
For all $t\in[\tau^0_i+\max\{M+2,K\}+1, \tau^0_{i+1}]\cap \set{E}^i$, we have $\varphi(w^*_t; \hat F_t^\mu) - \varphi(w^t_0; \hat F_t^\mu)\leq 2\Delta_t + \frac{\bar\xi  s}{K} $.
\end{lemma}
\noindent
Next we upper bound $E^i \triangleq \sum^{\tau^0_{i+1}}_{t=\tau^0_i+\max\{M+2,K\}+1}\1(t\notin \set{E}^i)$, the number of periods when $t\notin \set{E}^i$ (during the exploitation phase of epoch $i$).

\begin{lemma}\label{lem:retailer-learning-Ei}
With probability at least $1 - 1/T^2$, $E^i\leq 2\log(T)\sqrt{M(\tau^0_{i+1}-1 - \tau^0_i)} + 1$.
\end{lemma}
\noindent
We combine Lemmas~\ref{lem:retailer-learning-2} and \ref{lem:retailer-learning-Ei}, and summarize the resulting bound on $\reg_i^c(\policy{LUNA})$ in Eq.~\eqref{equ:appendix-retailer-learning-4} in Appendix~\ref{appendix:learning}.

\paragraph{Part II of the regret}
To upper bound $\reg_i^0(\policy{LUNA})$, we note
$$
\varphi(w^t_0; \hat F_t^\mu) - \varphi(w_t; \hat F_t^\mu) = 
\left(\varphi(w^t_0; \hat F_t^\mu) - \varphi(w_t; \hat F_t^\mu)\right)\1(w_t\neq w^t_0)
$$
for all $t$. That is, the supplier can only incur regret with respect to the benchmark of always pricing at $w^t_0$ if $w_t\neq w^t_0$.
Let
$$
T_i(K) \triangleq \left|\{t\in[\tau^0_i + K+1, \tau^0_{i+1}]: m_t \neq 0\}\right|
$$
be the number of periods in the exploitation phase of epoch $i$ when $\policy{LUNA}$ does not select $m_t = 0$ (and price at $w^t_0$).
The next lemma upper bounds $T_i(K)$.

\begin{lemma}(\citealp[Lemma A.2]{karnin2016multi})\label{lem:retailer-learning-4}
For all $K \geq 1$, we have $T_i(K) \leq \sqrt{11\log(T)M(\tau^0_{i+1}-1 - \tau^0_i)}$ with probability at least $1-1/T^2$.
\end{lemma}
\noindent
We summarize the resulting bound for $\reg_i^0(\policy{LUNA})$ in Eq.~\eqref{equ:appendix-retailer-learning-5} in Appendix~\ref{appendix:learning}.

\paragraph{Combining the two parts of the regret}
We combine Eq.~\eqref{equ:appendix-retailer-learning-4} and Eq.~\eqref{equ:appendix-retailer-learning-5} to upper bound the regret for epoch $i$ in Eq.~\eqref{equ:appendix-retailer-learning-7}.
To derive the supplier's total regret over the entire planning horizon $T$, we also need an upper bound on $I$ (the total number of epochs).
In Lemmas~\ref{lem:epoch-variation-1} and \ref{lem:epoch-variation-2}, we showed that when an epoch ends, $\hat F_t$ has varied a lot within the current epoch. Since the total variation of $\hat F_{1:T}$ over the entire planning horizon is bounded, the total number of epochs $I$ must be bounded by the variation budget.

\begin{lemma}\label{lem:retailer-learning-number-epoch}
We have $I \leq (s\,  \bar\xi)^{\frac{2}{3}}V^{\frac{2}{3}}M^{-\frac{1}{3}}T^{\frac{1}{3}} + 1$ almost surely.
\end{lemma}
\noindent
Based on Lemma~\ref{lem:retailer-learning-number-epoch} and Eq.~\eqref{equ:appendix-retailer-learning-7}, we obtain our final regret bound in Eq.~\eqref{equ:appendix-retailer-learning-final}, concluding the proof.

\subsection{Continuous Demand Distributions}\label{subsec:retailer-learning-discretization}

We now turn to the continuous case. We may also use the upcoming approach when the support of the demand distribution is finite but very large, so we do not specifically require $F_{1:T}$ and $\hat F_{1:T}^\mu$ to have densities for this treatment.
We do suppose that Assumption~\ref{ass:retailer-learning-3} is in force throughout this subsection.

When $\hat F_t^\mu$ is continuous, under Assumption~\ref{ass:retailer-learning-3}, $q_t$ can take any value in the interval $[0,\bar\xi]$.
In contrast, when $\hat F_t^\mu$ has support on $\set{Y}_M$, the retailer's order quantity always satisfies $q_t\in\set{Y}_M$. $\policy{LUNA}$ used this fact to track the variation of $\hat F_{1:T}^\mu$ when demand has support on $\set{Y}_M$, but it is much harder to infer the behavior of $\hat F_t^\mu$ in the continuous case.
Our strategy is based on approximating $[0,\bar\xi]$ with a finite subset of equally spaced points.
Let $N \geq 1$ be the size of this subset, and let $\set{Z}_{N} \triangleq \{z_n\}_{n \in [N]}$ be equally spaced points on $[0,\bar\xi]$ defined by: $z_n = (n-1)\, \bar\xi / (N-1)$ for all $n \in[N]$.

We call our pricing policy for the continuous case \textbf{L}earning \textbf{U}nder \textbf{N}on-stationary \textbf{A}gent with \textbf{C}ontinuous Distribution (LUNAC), denoted $\pi_{\text{LUNAC}}$, which calls $\pi_{\text{LUNA}}$ as a subroutine on $\set{Z}_N$.
The details of $\pi_{\text{LUNAC}}$ are outlined in Algorithm~\ref{alg:retailer-learning-discretization}.
In each period, $\pi_{\text{LUNAC}}$ enacts the wholesale price $w_t$ suggested by $\pi_{\text{LUNA}}$ and receives the feedback $q(w_t; \hat F_t^\mu)$. $\pi_{\text{LUNAC}}$ then maps the feedback $q(w_t; \hat F_t^\mu)$ to some $z_n \in \set{Z}_N$, which is then given to $\pi_{\text{LUNA}}$, which then outputs a recommended price.

\begin{algorithm}[H]
Input: Time horizon $T$, supplier production cost $c$, retailer selling price $s$, $N \geq 1$ and $\set{Z}_N$;\\
Initialize $\pi_{\text{LUNA}}$ with $\set{Z}_N$;\\
	\While{$t\leq T$}{
	Set the wholesale price $w_t$ suggested by $\pi_{\text{LUNA}}$ and observe retailer's order $q(w_t; \hat F_t^\mu)$; \\
Find $n$ such that $z_{n-1}< q(w_t; \hat F_t^\mu)\leq z_n$ for some $2\leq n\leq N$ ($n=1$ if $q(w_t;\hat F^\mu_t) = 0$), and take $z_n$ as the feedback to $\pi_{\text{LUNA}}$. 
	}
	\caption{\textbf{L}earning \textbf{U}nder \textbf{N}on-stationary \textbf{A}gent with \textbf{C}ontinuous Distribution (LUNAC)}
	\label{alg:retailer-learning-discretization}
\end{algorithm}

The retailer's order quantities are determined by $\hat F_{1:T}^\mu$ (which may have support on all of $[0, \bar \xi]$), while we are running $\pi_{\text{LUNA}}$ as a subroutine on the finite set $\set{Z}_N$.
To analyze the behavior of the $\policy{LUNA}$ subroutine, we introduce a sequence of fictitious perceived distributions with support on $\set{Z}_N$ that are based on the retailer's actual perceived distributions.
Let $\tilde F^\mu_t$ be the fictitious distribution on $\set{Z}_N$ for period $t$, which satisfies
\begin{equation}\label{equ:retailer-learning-discretization}
\Tilde F^\mu_t(z_n) = \hat F_t^\mu(z_n),\,\forall n \in [N],
\end{equation}
for all $t\in[T]$.
We introduce the shorthand $\tilde F^\mu_{1:t}\triangleq (\tilde F^\mu_i)^t_{i=1}$ for $t \in [T]$ for the partial sequences of fictitious perceived distributions.

We will establish that, under Eq.~\eqref{equ:retailer-learning-discretization}, the wholesale prices output by $\pi_{\text{LUNAC}}$ under $\hat F_{1:t}^\mu$ coincide with the wholesale prices output by $\pi_{\text{LUNA}}$ on $\set{Z}_N$ under $\tilde F_{1:t}$.
Let $\omega$ be a sample path of the randomization of $\pi_{\text{LUNA}}$, and let $\Omega$ be the set of all such sample paths.
All the randomization in $\pi_{\text{LUNAC}}$ comes from the randomization in $\pi_{\text{LUNA}}$ on $\set{Z}_N$, so we can compare both algorithms on $\Omega$.
Let $w^{\text{LUNAC}}_t(\hat F_{1:t-1}^\mu; \omega)$ be the wholesale price output by $\pi_{\text{LUNAC}}$ given the distributions $\hat F_{1:t-1}^\mu$ under $\omega$, and let $w^{\text{LUNA}}_t( \tilde F^\mu_{1:t-1}; \omega)$ be the wholesale price output by $\pi_{\text{LUNA}}$ given the distributions $\tilde F^\mu_{1:t-1}$ under $\omega$.

\begin{lemma}\label{lem:conti-approx}
For all $t \in [T]$ and $\omega \in \Omega$, $w^{\text{LUNAC}}_t(\hat F_{1:t-1}^\mu; \omega) = w^{\text{LUNA}}_t(\tilde F^\mu_{1:t-1}; \omega)$.
\end{lemma}
\noindent
Loosely speaking, Lemma~\ref{lem:conti-approx} says that $\pi_{\text{LUNAC}}$ sets wholesale prices by approximating $\hat F_t^\mu$ with $\tilde F^\mu_t$ in each period. It then outputs the wholesale prices given by $\pi_{\text{LUNA}}$, which pretends the retailer's perceived distribution is actually $\tilde F^\mu_t$. This interpretation suggests that if $\hat F_{1:T}^\mu$ has bounded variation, then $\tilde F_{1:T}$ should have bounded variation, as shown in Lemma~\ref{lem:retailer-learning-discretize}.
\begin{lemma}\label{lem:retailer-learning-discretize}
For all $t\in[T-1]$, $d_K(\Tilde F_t^\mu, \Tilde F_{t+1}^\mu)\leq d_K(\hat F_t^\mu, \hat F_{t+1}^\mu)$.
\end{lemma}


Theorem~\ref{thm:retailer-learning-2} below bounds the regret of $\pi_{\text{LUNAC}}$ (see Algorithm~\ref{alg:retailer-learning-discretization}).
$\pi_{\text{LUNAC}}$ does not require the variation budget $V$ as an input, but we get an improved regret bound with knowledge of $V$.

\begin{theorem}\label{thm:retailer-learning-2}
Suppose Assumption~\ref{ass:retailer-learning-3} holds.


(i)
If the supplier knows $V$, then $N$ can be chosen optimally as $N^* = \ceil{\bar\xi^{-\frac{1}{4}}V^{-\frac{1}{4}}T^{\frac{1}{4}}}$, and\\
$\reg(\pi_{LUNAC}, T) = \Tilde{O}(\bar\xi^{\frac{5}{4}}V^{\frac{1}{4}}T^{\frac{3}{4}})$.

(ii)
If the supplier does not know $V$, then $N$ can be chosen obliviously as $\hat N = \ceil{\bar\xi^{-\frac{1}{4}}T^{\frac{1}{4}}}$, and $\reg(\pi_{LUNAC}, T) = \Tilde{O} (\bar\xi^{\frac{5}{4}}V^{\frac{1}{3}}T^{\frac{3}{4}} + \bar\xi^{\frac{17}{12}}V^{\frac{2}{3}}T^{\frac{7}{12}})$.
\end{theorem}


According to Theorem~\ref{thm:retailer-learning-2}, the supplier can achieve a sublinear regret bound if $V = \tilde o(T)$ when the supplier knows $V$ and if $V = \tilde o(T^{\frac{5}{8}})$ when she does not know $V$, respectively. Theorem~\ref{thm:retailer-learning} shows that the supplier has a sublinear regret bound only if $V = \tilde o(T^{\frac{1}{2}})$, when there is no approximation of $\set{W}$. The improvement from $V = \tilde o(T^{\frac{1}{2}})$ to $V = \tilde o(T^{\frac{5}{8}})$ is achieved by approximation of the distribution. The supplier indirectly controls the number of epochs, and so a discrete approximation may lead to a better regret bound when $V$ is large.
It follows that the supplier can use \policy{LUNAC} not only when the distribution is continuous, but also for discrete distributions where the supplier believes the unknown $V$ is large.

\begin{remark}
When the supplier does not know $V$, she can combine LUNAC with the BOB framework \citep{cheung2019learning,cheung2021hedging}, which we refer to as LUNAC-N. The implementation details of $\pi_{\text{LUNAC-N}}$ are presented in Appendix~\ref{appendix:LUNAC-N}, and the upper bound on the regret of $\pi_{\text{LUNAC-N}}$ is presented in Theorem~\ref{thm:retailer-learning-M-BOB}. 
\end{remark}

Before we end this section, we comment on the difference between our discretization approach and those approaches used for the continuous bandit. The decision set is continuous in a continuous bandit, and it is common to approximate the decision set with a finite set. Instead of finding the optimal decision in the original continuous decision set, an optimal decision is found from the finite set, and then the regret is established through a regularity assumption (e.g., Lipschitz or Holder continuity) on the reward/cost function. We do not directly approximate the continuous decision set.
Instead, we approximate the supplier's profit function by finding an approximate distribution $\tilde F_t$ for $\hat F_t^\mu$ (albeit both the true profit function and $\hat F_t^\mu$ are unknown). Our approach relies on the bilinearity of the supplier's profit function in $w$ and $q$, and it does not require the regularity assumptions on the objective from the continuous bandit literature.

\section{Examples of Retailer's Strategies}
\label{sec:retailer-learning-examples}

We investigate several well known data-driven retailer inventory learning policies in this section, and show that our proposed pricing policies achieve sublinear regret for all of them. We emphasize that we do not need to know the retailer's exact inventory policy to achieve sublinear regret, we only mean to illustrate that these popular inventory policies satisfy our key assumption on the total variation of $\hat F_{1:T}^\mu$. In addition, the examples in this section suppose that the retailer does not have prior knowledge of $V$. 
Instead, these examples help provide guidance on refining $V$ in practice.

\subsection{Sample Average Approximation (SAA)}

SAA is arguably the most widely studied approach for data-driven optimization \citep{levi2015data,kleywegt2002sample}.
We let $\mu_{\text{e}}$ denote the retailer's inventory policy based on SAA.
For all $t \in [T]$, let $\hat F^{\text{e}}_t$ be the empirical CDF constructed from the (not necessarily i.i.d.) demand samples $(\xi_i)^{t-1}_{i=1}$ defined by $\hat F^{\text{e}}_t(x) \triangleq \frac{1}{t-1} \sum^{t-1}_{i=1}\1(\xi_i\leq x)$ for all $x \geq 0$ (note that in period $t$, the retailer only got access to the demand realizations in the previous $t-1$ periods).
Under $\mu_{\text{e}}$, given wholesale price $w_t$, the retailer's order quantity satisfies
$q^e_t = q_t(w_t; \hat F^{\text{e}}_t)$.

We can upper bound the total variation $V$ of the sequence $\hat F^{\text{e}}_{1:T}$ for arbitrary $F_{1:T}$ (i.e., the true distribution can be changing arbitrarily).

\begin{proposition}\label{prop:retailer-learning-saa}
$\mu_{\text{e}}\in\set{M}(\log(T)+1, T)$.
\end{proposition}

Next we bound the supplier's regret under $\pi_{\text{LUNA}}$ (if the distribution has finite support) or $\pi_{\text{LUNAC}}$ (if the distribution has continuous support). The proof follows from  Theorem~\ref{thm:retailer-learning} and Proposition~\ref{prop:retailer-learning-saa}.

\begin{theorem}\label{thm:retailer-learning-saa}
Suppose the retailer follows $\mu_{\text{e}}$.

(i) Suppose $F_{1:T}$ have support on $\set{Y}_M$, then $\reg(\policy{LUNA}, T) = \Tilde{O}(\bar\xi^{\frac{4}{3}}M^{\frac{1}{3}}T^{\frac{2}{3}})$.

(ii) Suppose $F_{1:T}$ have support on $[0, \bar \xi]$, then $\reg(\pi_{\text{LUNAC}}, T) = \Tilde{O}(\bar\xi^{\frac{5}{4}}T^{\frac{3}{4}} + \bar\xi^{\frac{17}{12}}T^{\frac{7}{12}})$.
\end{theorem}
\noindent

\subsection{Distributionally Robust Optimization (DRO)}

We suppose Assumption~\ref{ass:stationary-true-demand} holds for this subsection.
The DRO approach is based on the worst-case expected profit over an uncertainty set of demand distributions.
We let $\mu_{\text{r}}$ denote the retailer's inventory learning policy based on DRO.
In each period $t$, the retailer has an uncertainty set $\set{D}_t \subset \set{P}(\Xi)$ that he believes contains the true market demand distribution.
We consider uncertainty sets which consist of distributions that are ``close'' to the empirical distribution $\hat F^{\text{e}}_t$, and we measure closeness on $\set{P}(\Xi)$ with the $\phi-$divergence.
Recall the $\phi-$divergence, denoted $d_\phi$, for distributions $F,\,G\in \set{P}(\Xi)$ with $F \ll G$ (where $F \ll G$ means $F$ is absolutely continuous with respect to $G$) is defined by $d_{\phi}(F,G) = \int_{\Xi}\phi\left(dF/dG\right)dG$ for a convex function $\phi$ such that $\phi(1) = 0$.
Let $\epsilon_t \geq 0$ be the retailer's confidence level in period $t$.
The retailer's data-driven uncertainty sets under $\mu_{\text{r}}$ are $\set{D}_{\epsilon_t}^\phi(\hat F^{\text{e}}_t) \triangleq \{F\in\set{P}(\Xi): d_\phi(F, \hat F^{\text{e}}_t)\leq \epsilon_t\}$.
A retailer who is more confident that $\hat F^{\text{e}}_t$ is close to the true distribution $F_0$ should choose a smaller $\epsilon_t$, and vice versa.

The DRO literature has proposed multiple methods for choosing the confidence level $\epsilon_t$, see the review by \citet{rahimian2019distributionally}. One way is to leverage the asymptotic or finite sample performance of the uncertainty set. In other words, we would choose $\epsilon_t$ so that the optimal value of the DRO problem gives a finite sample guarantee on the retailer's original stochastic optimization problem.

\citet{duchi2016statistics} uses the optimal value of a DRO problem based on smooth $\phi-$divergences to provide asymptotic confidence intervals for the optimal value of the original (full information) stochastic optimization problem.
We will evaluate the performance of $\pi_{\text{LUNA}}$ and $\pi_{\text{LUNAC}}$ when $\epsilon_t$ is chosen by this method.
Let $\chi^2_1$ be a Chi-squared random variable with degree of freedom one.

\begin{theorem}(\citealp[Theorem 4]{duchi2016statistics})\label{thm:retailer-learning-phi-divergence-asymptotic}
Suppose the following conditions hold:

(i) The function $\phi:\mathbb{R}_{+}\to \mathbb{R}$ is convex, three times differentiable in a neighborhood of $1$, and satisfies $\phi(1) = \phi'(1) = 0$.

(ii) There exists a measurable function $M: \Xi\to \mathbb{R}_+$ such that for all $\xi\in\Xi$, $R(\cdot ;w,\xi)$ is $M(\xi)-$Lipschitz with respect to some norm $\Vert\cdot\Vert$ on $\Xi$. 

(iii) The function $R(\cdot ;w,\xi)$ is proper and lower semi-continuous for $F_0$-almost all $\xi\in\Xi$.\\
For any $\rho \geq 0$, let $\epsilon_t = \rho/t$ for all $t \geq 1$. Then,

\begin{equation}\label{equ:retailer-learning-phi-divergence-asymptotic}
\lim_{t\to\infty}\PP\left(
\max_{q\in\Xi}\E_{F_0}[R(q;w,\xi)]\geq l_t
\right)= 1-\frac{1}{2}\PP(\chi^2_1\geq \rho),
\end{equation}
where $l_t \triangleq \max_{q\in\Xi}\inf_{F \in \set{D}_{\epsilon_t}^\phi(\hat F^{\text{e}}_t) }\E_F[R(q;w,\xi)]$.

\end{theorem}
\noindent
Theorem~\ref{thm:retailer-learning-phi-divergence-asymptotic} says that the optimal value of the retailer's problem with knowledge of $F_0$ can be lower bounded by $l_t$ (the optimal value of the DRO problem for uncertainty set $\set{D}_{\epsilon_t}^\phi(\hat F^{\text{e}}_t)$) with probability $1-\frac{1}{2}\PP(\chi^2_1\geq \rho)$, as $t\to\infty$ for confidence levels $\epsilon_t = \rho/t$.
Note that Assumption~\ref{ass:stationary-true-demand} must be satisfied for this result to hold.

Let $\alpha\in[0,1]$ be a confidence level and $\chi^2_{1,\beta}$ denote the $\beta-$quantile of the $\chi^2_1$ distribution.
Theorem~\ref{thm:retailer-learning-phi-divergence-asymptotic} suggests that in order to ensure the asymptotic coverage of the optimal value as in Eq.~\eqref{equ:retailer-learning-phi-divergence-asymptotic}, the confidence levels should be chosen as
\begin{equation}\label{equ:retailer-learning-dro-phi-theta}
    \epsilon_t = \chi_{1,1-2\alpha}^2/(t-1),\,\forall t\geq 2.
\end{equation}
Confidence levels chosen in this way are usually overly conservative, and the retailer may choose a smaller $\epsilon_t$ in practice. In this case, we get a conservative estimate of $V$ by this method.

Under $\mu_{\text{r}}$, in period $t$ the retailer orders
\begin{equation}\label{equ:retailer-learning-dro}
q^r_t\in\arg\max_{q\geq 0} \min_{F\in \set{D}_{\epsilon_t}^\phi(\hat F^{\text{e}}_t) }\,\E_{F}[R(q;w_t,\xi)].
\end{equation}
The objective in Eq.~\eqref{equ:retailer-learning-dro} is the retailer's worst-case expected profit. Then, corresponding to $q^r_t$, the perceived distribution
$$
\hat F_t^{\text{d}}  \in \arg\min_{F\in \set{D}_{\epsilon_t}^\phi(\hat F^{\text{e}}_t) } \E_F[R(q^r_t; w_t, \xi)],
$$
is the distribution in $\set{D}_{\epsilon_t}^\phi(\hat F^{\text{e}}_t)$ which attains the worst-case expected profit (which also depends on $w_t$).

We consider three widely used $\phi-$divergences: the KL-divergence $d_{KL}$ (where $\phi(x) = x\log(x)$), the $\chi^2-$distance $d_{\chi^2}$ (where $\phi(x) = (x-1)^2$), and the Hellinger distance $d_H$ (where $\phi(x) = \left(\sqrt{x}-1\right)^2$).
For these three, Proposition~\ref{prop:retailer-learning-phi-divergence}  upper bounds the total variation $V$ of the sequence $\hat F_{1:T}^{\text{d}}$ as a function of $\epsilon_{1:T} = (\epsilon_t)_{t=1}^T$.

\begin{proposition}\label{prop:retailer-learning-phi-divergence}
 Under Assumption~\ref{ass:stationary-true-demand}, suppose the retailer follows $\mu_{\text{r}}$ with confidence levels $\epsilon_{1:T}$.

(i) If $d_{\phi} = d_{KL}$, 
then $\mu_r\in\set{M}\left(\log(T) +1+ \sum^T_{t=1}\sqrt{2\epsilon_t}, T\right)$.

(ii) If $d_{\phi} = d_{\chi^2}$, then $\mu_r\in\set{M}\left(\log(T)+1 + \sum^T_{t=1}\sqrt{\epsilon_t}, T\right)$.

(iii) If $d_{\phi} = d_{H}$, then $\mu_r\in\set{M}\left(\log(T)+1 + 2\sum^T_{t=1}\epsilon_t, T\right)$.
\end{proposition}

Using the specific choice of $\epsilon_{2:T}$ in Eq.~\eqref{equ:retailer-learning-dro-phi-theta} and $\epsilon_1$ = 1, Theorem~\ref{thm:retailer-learning-dro-phi-divergence} characterizes the performance of $\pi_{\text{LUNA}}$ and $\pi_{\text{LUNAC}}$ for $d_\phi\in\{d_{KL}, d_{\chi^2}, d_H\}$. Theorem~\ref{thm:retailer-learning-dro-phi-divergence} follows directly from Theorem~\ref{thm:retailer-learning}, Theorem~\ref{thm:retailer-learning-2}, and Proposition~\ref{prop:retailer-learning-phi-divergence}.

\begin{theorem}\label{thm:retailer-learning-dro-phi-divergence}
Suppose Assumptions~\ref{ass:stationary-true-demand} and \ref{ass:retailer-learning-3} hold, and suppose the retailer follows $\mu_{\text{r}}$ where $\epsilon_{1:T}$ are chosen as in Eq.~\eqref{equ:retailer-learning-dro-phi-theta}.


(i) If $d_{\phi}\in\{d_{KL}, d_{\chi^2}\}$, then $\reg(\pi_{\text{LUNAC}},T) = \Tilde{O}\big(\bar\xi^{\frac{17}{12}}T^{\frac{11}{12}}\big)$. 

(ii) If $d_{\phi} = d_{H}$ and $F_0$ has support on $\set{Y}_M$, then $\reg(\pi_{\text{LUNA}}, T) = \Tilde{O}(\bar\xi^{\frac{4}{3}}M^{\frac{1}{3}}T^{\frac{2}{3}} + \bar\xi^{\frac{4}{3}}M^{-\frac{1}{3}}T^{\frac{2}{3}})$.

(iii) If $d_{\phi} =d_{H}$ and $F_0$ has support on $[0, \bar \xi]$, then $\reg(\pi_{\text{LUNAC}}, T)= \Tilde{O}(\bar\xi^{\frac{5}{4}}T^{\frac{3}{4}}+\bar\xi^{\frac{17}{12}}T^{\frac{7}{12}})$.
\end{theorem}


\subsection{Parametric Approach}
We continue to suppose Assumption~\ref{ass:stationary-true-demand} is in force for this subsection.
We additionally suppose that the retailer has a parametric model for $F_0$ determined by the parameter $\theta \in \mathbb R^d$. Let $\Theta \subset \mathbb R^d$ be the set of admissible parameter values, and let $\{F_\theta\}_{\theta \in \Theta}$ be the corresponding parametric family. 
If $F_0$ belongs to a parametric family, then Assumption~\ref{ass:retailer-learning-3} is not likely to be satisfied (since many parametric distributions such as the normal and exponential distributions have unbounded support). In this case, we relax to the following assumption. 

\begin{assumption}\label{ass:retailer-learning-4}
The retailer's order quantity is upper bounded by $\bar q$. That is, for any $w_t$ and $\hat F^\mu_t$, the retailer's order satisfies $q(w_t; \hat F^\mu_t) = \min\{\min\{q: \hat F^\mu_t(q) \geq 1 - w_t/s\}, \bar q\}$.
\end{assumption}
\noindent
Assumption~\ref{ass:retailer-learning-4} states that even if the supplier's price $w_t$ is low enough and the perceived distribution $\hat F^\mu_t$ has unbounded support, the retailer will not place arbitrarily large orders. Assumption~\ref{ass:retailer-learning-4} is also consistent with practical constraints, e.g., warehouse and transportation capacity.

We consider three specific methods for the parametric setting: (i) maximum likelihood estimation (MLE); (ii) operational statistics; and (iii) the parametric Bayesian approach. 

\subsubsection{Maximum likelihood Estimation (MLE)}

We focus on MLE for the exponential family, where a distribution belongs to the exponential family if its probability density function $f(x;\theta)$ for $x$ in its support can be written as:
\begin{equation}
\label{eq:exponential_family}
f(x;\theta) = h(x)\exp{\left(\eta(\theta)^T\cdot T(x) - A(\theta)\right)}.
\end{equation}
In Eq.~\eqref{eq:exponential_family}, $\eta(\theta)$ is the natural parameter, $T(x)$ is the sufficient statistic, $h(x)$ is the base measure, and $A(\theta)$ is the log-partition function which normalizes the density function.
The exponential family includes the Poisson and Categorical distributions (for discrete demand), and the Normal and Exponential distributions (for continuous demand).

We let $\mu_{\text{m}}$ denote the retailer policy based on MLE.
Under $\mu_{\text{m}}$, the retailer produces an estimate $\theta_t$ of $\theta$ in each period $t$ by maximizing the likelihood function of the past demand samples. This procedure has a special form for the exponential family. 
Let $\mu \triangleq \E_{F_\theta}[T(\xi)]$, then the MLE for $\mu$ based on demand samples $(\xi_i)_{i=1}^{t-1}$ is:
\begin{equation*}
    \mu_t = \frac{\sum^{t-1}_{i=1}T(\xi_i)}{t-1},\,t\geq 2.
\end{equation*}
One can then obtain the estimator $\theta_t$ for $\theta$ through the estimator $\mu_t$ for $\mu$ by the relationship between $\mu$ and $\theta$, which depends on the particular distribution.
Under $\mu_{\text{m}}$, in each period $t \in [T]$, the retailer's perceived distribution is the fitted distribution 
\begin{equation}\label{equ:MLE-F-hat}
    \begin{split}
    \hat F_t^{\text{m}}(x) =
    \begin{cases}
        F_{\theta_t}(x), \,& 0\leq x < \bar q;\\
    1, \,&x\geq \bar q.
    \end{cases}
    \end{split}
\end{equation}
This sequence satisfies Assumption~\ref{ass:retailer-learning-1} and the discontituity in $\hat F^m_t$ is introduced by Assumption~\ref{ass:retailer-learning-4}.
The retailer then orders $q^{\text{m}}_t = q_t(w_t, \hat F_t^{\text{m}})$ where $q_t(w_t, \hat F_t^{\text{m}}) = \min\{\min\{q: F_{\theta_t}(q) \geq 1 - w/s\}, \bar q\}$.

We will investigate the variation of $\hat F_{1:T}^m$ for some canonical distributions in the exponential family. 
Let ${\rm P}(\lambda)$ denote the Poisson distribution with mean $\lambda$; let ${\rm C}(M)$ denote the categorical distribution with support size $M$; let ${\rm E}(\lambda)$ denote the exponential distribution with rate $\lambda$; and let ${\rm N}(\mu, \sigma^2)$ denote the normal distribution with mean $\mu$ and variance $\sigma^2$.

In Proposition~\ref{prop:retailer-learning-exp-family}, we derive the total variation of $\hat F_{1:T}^m$.
Note the categorical distribution has bounded support so it automatically satisfies Assumption~\ref{ass:retailer-learning-4}.

\begin{proposition}\label{prop:retailer-learning-exp-family}
Suppose the retailer follows $\mu_{\text{m}}$.

(i) If $F_0 = {\rm P}(\lambda)$, then $\mu_m\in\set{M}\left( (\ln{(T)} + 1)\left(4\ln{(T)} + 2\lambda\right), T\right)$ with probability at least $1-1/T$. 

(ii) If $F_0 = {\rm C}(M)$, then  $\mu_m\in\set{M}\left(\ln{(T+1)}, T\right)$. 

(iii) If $F_0 = {\rm E}(\lambda)$, then $\mu_m\in\set{M}\left(16\ln{(2T^2)}-1 + \left(1 +2\ln{(2T^2)}\right)(\ln{(T)}+1), T\right)$ with probability at least $1-1/T$. 


(iv) If $F_0 = {\rm N}(\mu, \sigma^2)$, with $\sigma^2$ known and $\mu$ unknown to the retailer, then\\
$\mu_m\in \set{M}\left(1 + \frac{1}{\sigma}\sqrt{(\ln(T)+1)\left(\mu^2 + 2\sigma^2\ln{(2T^2)}\right)}, T\right)$ with probability at least $1-1/T$. 



\end{proposition}

The next result on the supplier's regret bound follows directly from Theorem~\ref{thm:retailer-learning}, Theorem~\ref{thm:retailer-learning-2}, and Proposition~\ref{prop:retailer-learning-exp-family}.

\begin{theorem}\label{thm:retailer-learning-exp-family}
Suppose the retailer follows $\mu_{\text{m}}$. 

(i) Suppose $F_0 = {\rm P}(\lambda)$, then $\reg(\policy{LUNAC}, T) = \Tilde{O}(\bar q^{\frac{5}{4}}T^{\frac{3}{4}} + \bar q^{\frac{17}{12}}T^{\frac{7}{12}})$ with probability at least $1-1/T$. 

(ii) Suppose $F_0 = {\rm C}(M)$, then $\reg(\policy{LUNA}, T) = \Tilde{O}(\bar q^{\frac{4}{3}}M^{\frac{1}{3}}T^{\frac{2}{3}})$.  

(iii) Suppose $F_0 = {\rm E}(\lambda)$, then $\reg(\policy{LUNAC}, T) = \Tilde{O}(\bar q^{\frac{5}{4}}T^{\frac{3}{4}} + \bar q^{\frac{17}{12}}T^{\frac{7}{12}})$ with probability at least $1-1/T$. 


(iv) Suppose $F_0 = {\rm N}(\mu, \sigma^2)$, then $\reg(\policy{LUNAC}, T) = \Tilde{O}(\bar q^{\frac{5}{4}}T^{\frac{3}{4}} + \bar q^{\frac{17}{12}}T^{\frac{7}{12}})$ with probability at least $1-1/T$. 



\end{theorem}

\subsubsection{Operational Statistics}

Here we suppose $F_0 = {\rm E}(\lambda)$, the exponential distribution with an unknown rate $\lambda > 0$.
\citet{liyanage2005practical,chu2008solving} propose the operational statistics approach for the retailer facing exponential demand with unknown rate.
In this approach, the retailer first specifies a class of admissible policies, then finds a policy within this class to maximize his out-of-sample expected profit. 

We let $\mu_{\text{o}}$ denote the retailer's inventory policy based on operational statistics, which is implemented as follows.
According to \citet{liyanage2005practical}, given wholesale price $w_t$ and i.i.d. demand samples $(\xi_i)_{i=1}^{t-1}$, the retailer's order quantity that maximizes his out-of-sample expected profit is
$(t-1)\left(\left(\frac{s}{w_t}\right)^{\frac{1}{t}} - 1\right)\frac{\sum^{t-1}_{i=1}\xi_i}{t-1}$ for $t\geq 2$.
The order quantity is directly determined by the data, and its derivation does not involve estimating $\lambda$. However, we still can find a sequence of distributions $\hat F^{\text{o}}_{2:T}$ satisfying Assumption~\ref{ass:retailer-learning-1} that map $w_t$ to the order quantity under operational statistics.
For all $t\geq 2$, define the rates $\lambda_t$ so that
$\frac{1}{\lambda_t} = \frac{(t-1)\left(\left(\frac{s}{w_t}\right)^{\frac{1}{t}} - 1\right) \left( \frac{\sum^{t-1}_{i=1}\xi_i}{t-1} \right)}{\ln\left(\frac{s}{w_t}\right)}$,
and then set the perceived distribution to be
$$
\hat F^{\text{o}}_t(x) = \begin{cases}
    {\rm E}(\lambda_t)(x),\,& 0\leq x<\bar q,\\
    1,\,& x\geq \bar q.
\end{cases}
$$
Then, under $\mu_{\text{o}}$ the retailer equivalently solves $q^{\text{o}}_t = q(w_t, \hat F^{\text{o}}_t)$ where $q(w_t, \hat F^{\text{o}}_t) = \min\{\min\{q: {\rm E}(\lambda_t)(q) \geq 1 - w/s\}, \bar q\}$
for all $t\geq 2$.

We now derive the total variation of $\hat F^{\text{o}}_{1:T}$.

\begin{proposition}\label{prop:retailer-learning-os}
Suppose $F_0 = {\rm E}(\lambda)$, and the retailer follows $\mu_{\text{o}}$. Then,\\
$\mu_o\in\set{M}\left(21 + 40\ln{(T)} + 4(\ln{(T)})^2, T\right)$ with probability at least $1 - 1/T$.
\end{proposition}

\begin{theorem}\label{thm:retailer-learning-os}
Suppose $F_0 = {\rm E}(\lambda)$, and the retailer follows $\mu_{\text{o}}$.
Then, $\reg(\policy{LUNAC},T) = \Tilde{O}(\bar q^{\frac{5}{4}}T^{\frac{3}{4}} + \bar q^{\frac{17}{12}}T^{\frac{7}{12}})$ with probability at least $1-1/T$.
\end{theorem}

\subsubsection{Parametric Bayesian approach}

We now suppose demand is ${\rm E}(\Lambda)$ where the rate $\Lambda$ is random and has a gamma prior distribution $f_\Lambda$ with parameter $\alpha, \beta > 0$, i.e., $f_\Lambda(\lambda) = \frac{(\beta/\lambda)^{\alpha + 1}}{\beta\Gamma(\alpha)}\exp\{-\beta/\lambda\}$.
Exponential demand distributions with a gamma prior are widely studied in the OM literature \citep{azoury1985bayes}. 

We let $\mu_{\text{b}}$ denote the retailer's inventory learning policy under the Bayesian approach.
Given $w_t$, the retailer orders 
\begin{equation}\label{equ:retailer-learning-bayesian}
    \arg \max_{q\geq 0}\,\int_{0}^{\infty} \E_{{\rm E}(\Lambda)}[R(q;w_t,\xi)]f_{\Lambda | (\xi_i)^{t-1}_{i=1}}\left(\lambda\right)d\lambda,
\end{equation}
where $f_{\Lambda | (\xi_i)^{t-1}_{i=1}}$ is the posterior density function with respect to the demand samples from the previous $t-1$ periods.
\citet{lim2006model} show that the retailer's optimal order quantity as a solution to Eq.~\eqref{equ:retailer-learning-bayesian} is $\left(\beta + \sum^{t-1}_{i=1}\xi_i\right)\left(\left(\frac{s}{w_t}\right)^{1/(\alpha+t-1)}-1\right)$.
For all $t \in [T]$, define the rates $\lambda_t$ so that
$$
\frac{1}{\lambda_t} = \frac{\left(\beta + \sum^{t-1}_{i=1}\xi_i\right)\left(\left(s / w_t \right)^{\frac{1}{\alpha+t-1}} - 1\right)}{\ln\left( s / w_t \right)},
$$
and then set
$$
\hat F^b_t  = \begin{cases}
    {\rm E}(\lambda_t)(x),\,& 0\leq x<\bar q,\\
    1,\,& x\geq \bar q.
\end{cases}
$$
Then, $\hat F^b_t$ is the retailer's perceived distribution that maps the supplier's wholesale price $w_t$ to the retailer's order quantity $q^b_t$ via $q^b_t = q(w_t, \hat F^b_t)$ where $q(w_t, \hat F^b_t) = \min\{\min\{q:  {\rm E}(\lambda_t)(q) \geq 1 - w/s\}, \bar q\}$.

We derive the total variation of $\hat F^b_{1:T}$ similarly to Proposition~\ref{prop:retailer-learning-os}, and so we omit the proof.

\begin{proposition}\label{prop:retailer-learning-bayesian}
Suppose demand is exponentially distributed with mean $1/\lambda$ and the retailer follows $\mu_{\text{b}}$.
Then, $\mu_{\text{b}} \in \set{M} ( 18(\alpha+1)+ (40\alpha+39)\ln{(T)}  + 4(\alpha+1)(\ln{(T)})^2, T)$ with probability at least $1 - 1/T$.
\end{proposition}

Next we bound the regret of $\policy{LUNAC}$ when the retailer follows $\mu_{\text{b}}$. 

\begin{theorem}\label{thm:retailer-learning-bayesian}
Suppose the retailer follows $\mu_{\text{b}}$. Then, $\reg(\policy{LUNAC}, T) = \Tilde{O}(\bar q^{\frac{5}{4}}T^{\frac{3}{4}} + \bar q^{\frac{17}{12}}T^{\frac{7}{12}})$ with probability at least $1-1/T$. 
\end{theorem}

\section{Numerical Experiments}\label{sec:numerical}

\subsection{Empirical Performance}
We evaluate the empirical performance of $\pi_{\text{LUNA}}$ when the true market demand distributions $F_{1:T}$ are discrete. In order to control the total variation of $\hat F^\mu_{1:T}$, we directly construct $\hat F_{1:T}^\mu$ as follows. For $t \in [T]$, we set $\hat F_t^\mu$ to be the CDF of a Bernoulli random variable which takes values $0$ and $1$ with probabilities $p_{t,0}$ and $p_{t,1} = 1-p_{t,0}$, respectively, and let
\begin{equation}\label{equ:exp-pt}
    p_{t,0} = \frac{1}{2} + \frac{3}{10}\sin{\frac{5V\pi t}{3T}},\, t\in[T],
\end{equation}
for fixed $V > 0$.
Then, the total variation satisfies
\begin{equation*}
    \sum^{T- 1}_{t=1}d_K(\hat F^\mu_t, \hat F^\mu_{t + 1}) = C\, V,
\end{equation*}
for some constant $C > 0$.

\begin{figure}[h]
     \begin{minipage}[t]{0.4\textwidth}
         \centering
         \includegraphics[width=\textwidth]{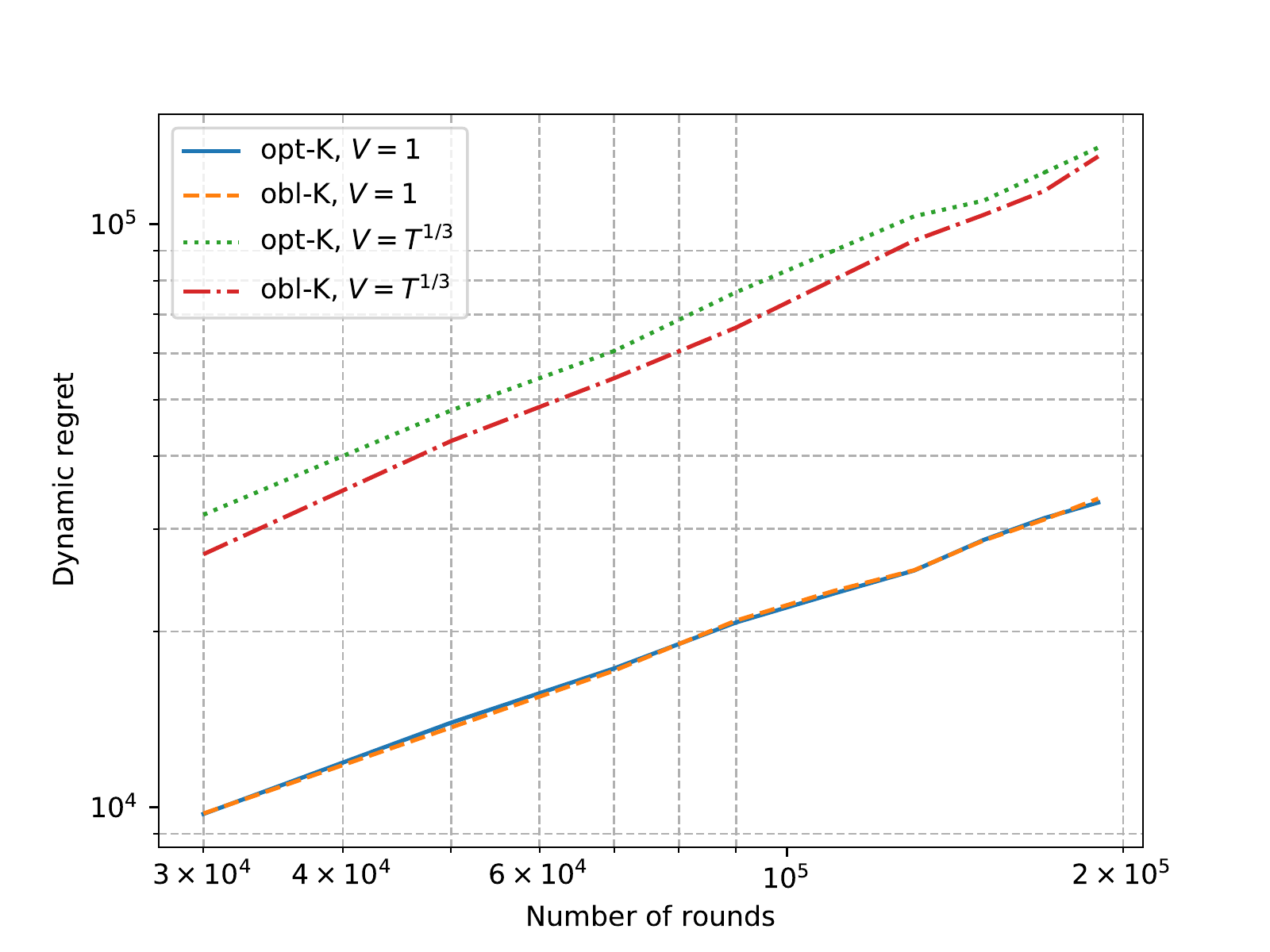}
         \caption{Performance of $\pi_{\text{LUNA}}$ for discrete distributions}
         \label{fig:exp-empirical-discrete}
     \end{minipage}
     \centering
\end{figure}

Figure~\ref{fig:exp-empirical-discrete} shows the dynamic regret of $\pi_{\text{LUNA}}$ as a function of the number of rounds when $K$ is optimally chosen (assuming the supplier knows $V$, with shorthand `opt-$K$') and obliviously chosen (assuming the supplier does not know $V$, with shorthand `obl-$K$'). We take different values of $V$ to compare the growth rate of the regret for the same pricing policy.
The regrets are plotted on a log-log scale, so the slope in this plot corresponds to the exponent of the regret, i.e., the slope is $\alpha$ if the regret grows in $\Theta(T^\alpha)$.

We see that the slope roughly matches our theoretical results (see Theorem~\ref{thm:retailer-learning}). When $V = 1$, the regret bounds corresponding to opt-$K$ and obl-$K$ overlap. 
When $V = T^{1/3}$, opt-$K$ grows more slowly than obl-$K$ but has a smaller constant term than opt-K (we did not optimize for the constant terms). Also notice that the gap in the regret bounds between opt-K and obl-$K$ is small (i.e., the regret bounds for opt-$K$ and obl-$K$ are respectively $\tilde{O}(V^{\frac{1}{3}}T^{\frac{2}{3}})$ and $\tilde{O}(V^{\frac{2}{3}}T^{\frac{2}{3}})$, and the gap is $V^{\frac{1}{3}}$). Even when $K$ is optimally chosen, obl-$K$ can have better performance than opt-$K$ within a large range of $T$ (in our case, $T\leq 2\times 10^5$) because of the constant terms dominating the regret bounds.

\subsection{Comparison between Different Algorithms}

In this subsection, we compare $\policy{LUNA}$ with some pricing policies that are designed for non-stationary bandits. Specifically, we compare $\policy{LUNA}$ with the following benchmarks:
\begin{enumerate}
    \item The Exp3.S algorithm by \citet{besbes2014optimal}, which is designed for non-stationary multi-armed bandits with known variation budget $B_T$. Notice we distinguish $B_T$ from our variation budget $V$ because $B_T$ refers to the norm of variation in the mean bandit feedback and $V$ refers to the total Kolmogorov variation in the sequence $\hat F^\mu_{1:T}$. The regret upper bound for Exp3.S is $\tilde O(d^{\frac{1}{3}}B^{\frac{1}{3}}_TT^{\frac{2}{3}})$ with $d$ arms, see \citet{besbes2014optimal}.
    
    \item The deterministic non-stationary bandit algorithm proposed by \citet{karnin2016multi} for multi-armed bandits with unknown variation budget $B_T$. The regret upper bound for this algorithm is $\tilde O(d^{\frac{1}{2}} T^{\frac{1}{2}} + d^{\frac{1}{3}} B^{\frac{1}{3}}_T T^{\frac{2}{3}})$.
    
    \item The Master+UCB1 algorithm proposed by \citet{wei2021non} for non-stationary stochastic bandits with unknown variation budget $B_T$. The regret upper bound is $\tilde O(B^{\frac{1}{3}}_TT^{\frac{2}{3}})$.
\end{enumerate}
We note that all of these pricing policies are designed for problems with finitely many admissible decisions. For fair comparison, we developed a version of $\policy{LUNA}$ that works when $\set{W}$ is finite which we call $\policy{LUNAF}$ (see Appendix~\ref{appendix:LUNAF}). Since Exp3.S requires the variation budget $B_T$ as an input, we calculate $V$ in our problem and simply let $B_T = V$.

\begin{figure}[h]
     \begin{minipage}[t]{0.4\textwidth}
         \centering
         \includegraphics[width=\textwidth]{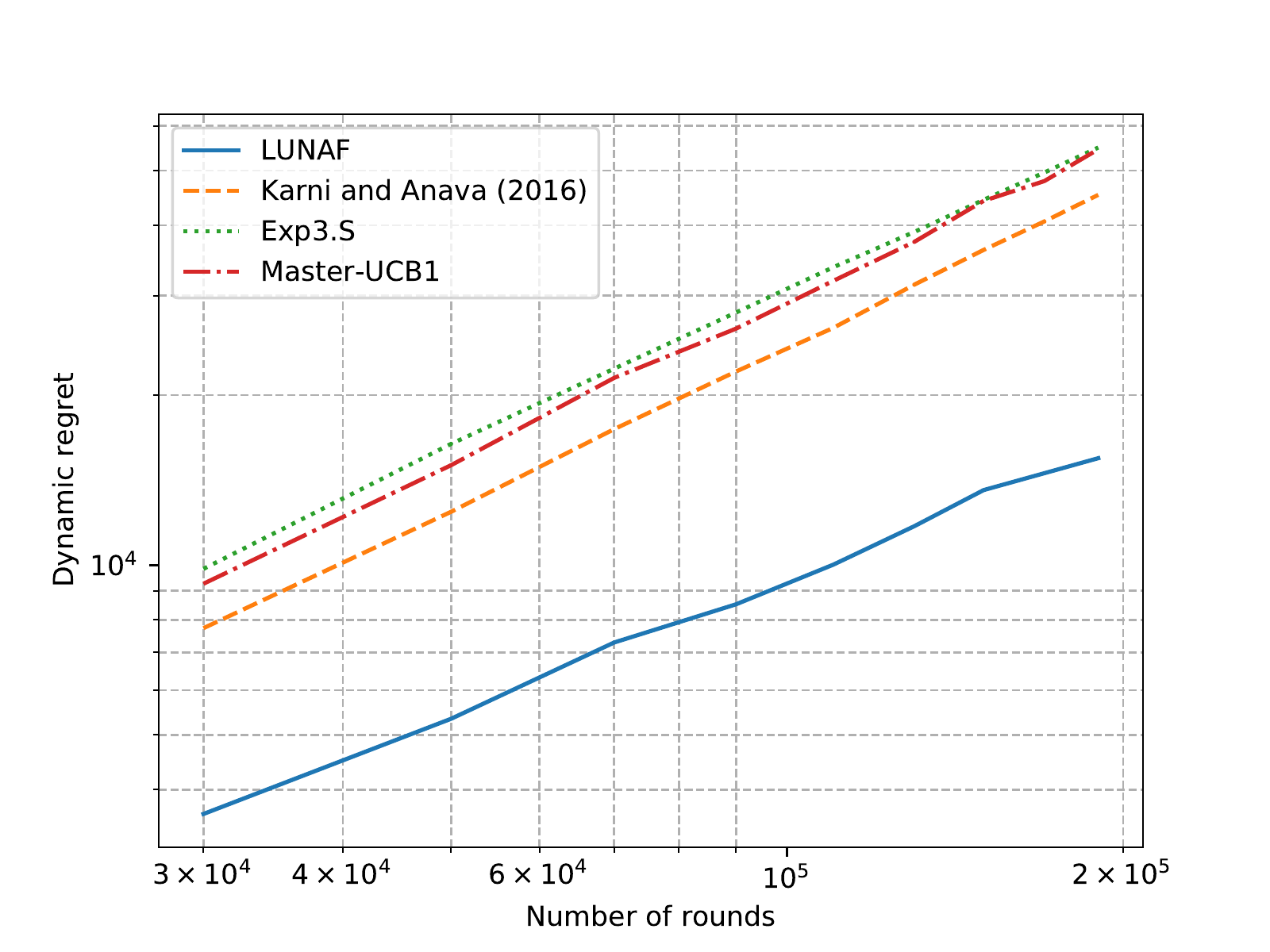}
         \caption{Performance comparison}
         \label{fig:exp-comparison-discrete}
     \end{minipage}
     \begin{minipage}[t]{0.4\textwidth}
         \centering
         \includegraphics[width=\textwidth]{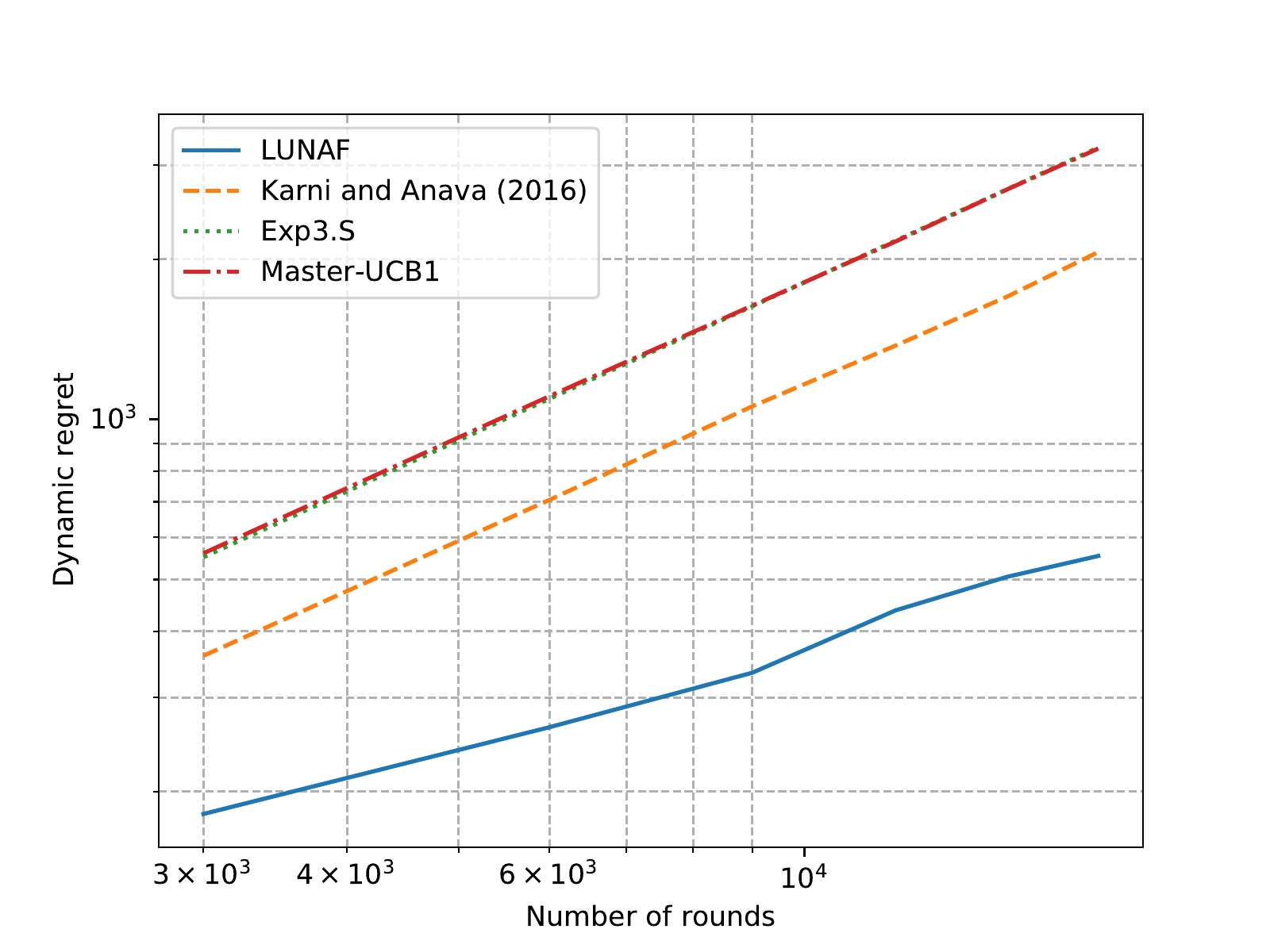}
         \caption{Performance comparison}
         \label{fig:exp-comparison-saa}
     \end{minipage}
     \centering
        \par
        \footnotesize
\emph{Notes:} (Left) $\hat F_t^\mu$ is set according to Eq.~\eqref{equ:exp-pt}. (Right) $\hat F_t$ is the CDF of the Bernoulli distribution and the retailer adopts SAA.
\end{figure}

We compare the regret of these benchmark algorithms with $\policy{LUNAF}$ in Figure~\ref{fig:exp-comparison-discrete} and Figure~\ref{fig:exp-comparison-saa}. 
In Figure~\ref{fig:exp-comparison-discrete}, we directly simulated the retailer's ordering decisions by setting $\hat F_t^\mu$ as in Eq.~\eqref{equ:exp-pt}. In Figure~\ref{fig:exp-comparison-saa}, we set the true distribution $F_t$ to be Bernoulli which takes values $0$ and $1$ with probabilities $p_t$ and $1 - p_t$ respectively where $p_t$ is also determined by Eq.~\eqref{equ:exp-pt}, i.e.,
\begin{equation*}
    p_t = \frac{1}{2} + \frac{3}{10}\sin{\frac{5V\pi t}{3T}},\, t\in[T],
\end{equation*}
in which case the true market demand distribution is non-stationary.
We also suppose the retailer follows $\mu_{\text{e}}$. In both experiments, we let $\set{W}$ contain $d = \ceil{T^{\frac{1}{2}}}$ equally spaced prices lying in $[0, s]$.

From Figures~\ref{fig:exp-comparison-discrete} and \ref{fig:exp-comparison-saa}, we see that $\policy{LUNAF}$ outperforms the benchmarks, and the performance of the benchmarks is relatively close to each other. These results suggest that the supplier benefits from using the structure of the profit function in her pricing policy, instead of applying a black box algorithm. In addition, based on the results in Figure~\ref{fig:exp-comparison-saa}, we see $\policy{LUNAF}$ still performs well even when the true market demand distribution is non-stationary. 

\subsection{Experiment on Semi-synthetic Data Set}
For our final experiment, we collected the weekly sales data of avocados in California from $2020$ to $2022$ \citep{avocado}. Avocado sales can be non-stationary and vary from month to month, see, e.g., \citet{keskin2021nonstationary}.
In order to approximate this non-stationarity, we first group the weekly sales data by month $m = \{1, \ldots, 12\}$. Then, to generate the daily sales in month $m$, we divide the weekly sales in month $m$ by $7$ and treat it as a sample of daily demand in month $m$. We repeat this procedure for all the weeks from $2020$ to $2022$ to get demand samples for each month of the year. Finally, we divide the daily sales by $1,000,000$ and round it to the nearest integer to build an approximate discrete daily demand distribution for avocados (in millions of units). Given the daily demand samples for each month, we then bootstrap the daily demand for a planning horizon of $T$ days (assuming the first day in the horizon starts on Jan $1$st). In this way, we generate random demand realizations for the retailer. 

\begin{figure}[h]
     \begin{minipage}[t]{0.4\textwidth}
         \centering
         \includegraphics[width=\textwidth]{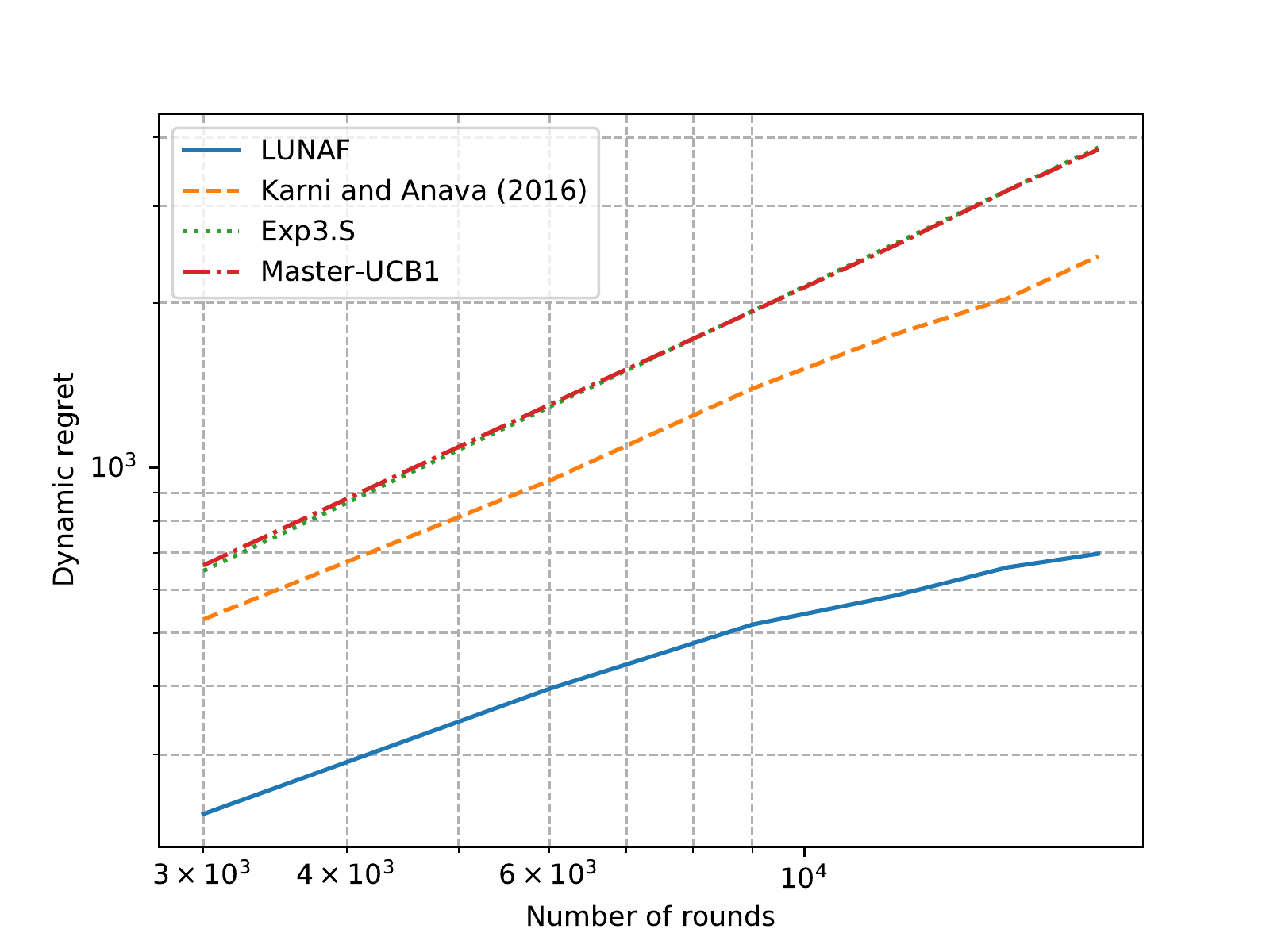}
         \caption{Performance comparison}
         \label{fig:exp-saa-realdata}
     \end{minipage}
     \centering
        \par
        \footnotesize
\end{figure}

We suppose the retailer follows $\mu_{\text{e}}$, and $\set{W}$ has cardinality $d = \ceil{T^{\frac{1}{2}}}$.
We compare the performance of the pricing policies for this setting in Figure~\ref{fig:exp-saa-realdata}.
We see that $\policy{LUNAF}$ outperforms the other policies in this setting as well. 
Exp3.S and Master-UCB1 have almost identical performance here, and the deterministic non-stationary bandit algorithm by \citet{karnin2016multi} outperforms both Exp3.S and Master-UCB1. This application further demonstrates that our pricing policy performs well even for non-stationary demand distributions.

\section{Conclusion}\label{sec:conclusion}

In this paper, we studied the supplier's pricing problem facing a retailer who is learning the demand distribution and employs data-driven inventory learning policies. We model the non-stationarity of the retailer's inventory decisions through the non-stationarity of his ``perceived" distributions. Then, we use the Kolmogorov distance to measure the variation of the retailer's perceived distributions and identify a tractable class of retailer policies.
For both discrete and continuous demand distributions, we proposed pricing policies for the supplier and derived sublinear regret upper bounds.
Our main conclusion is that the supplier can achieve asymptotically vanishing regret, even when the retailer is also learning the demand distribution, as long as the retailer's inventory policies belong to a reasonable class with bounded variation.

Much of the literature on optimization and learning in OM focuses on learning the random demand or unknown demand-price relationship. However, our work investigates the important problem of learning the learning policies implemented by a secondary agent in a multi-agent setting. 
This study brings new perspectives into learning in multi-agent problems in supply chain and inventory management, where the controller must learn to react to the learning policies by other agents in the system.

At the same time, we acknowledge some directions for future research. First, it is worth investigating information-theoretic lower bounds on the supplier's regret in our problem setting. Second, it may be possible to improve the supplier's regret bound when the non-stationarity has additional structure (e.g., seasonal demand patterns).

%


\ACKNOWLEDGMENT{%
}

%
%
%


\bibliographystyle{informs2014}
\bibliography{Ref}

\ECSwitch
\SingleSpacedXI
\ECHead{Electronic Companion}

\section{Additional material for Section~\ref{sec:stationary}}\label{appendix:stationary}

\subsection{Proof of Theorem~\ref{thm:bound-stationary retailer}}

We have
    \begin{equation*}
        \begin{split}
\reg(\pi_{\text{stat}}, T) &= \sum^T_{t=1}\E\left[(w^*-c)q(w^*;F_0) - (w_t - c)q(w_t; F_0)\right]\\
& 
 = \sum^{\ceil{\sqrt{T}}}_{t=1}\E\left[(w^*-c)q(w^*;F_0) - (w_t - c)q(w_t; F_0)\right]\\
 & + \sum^{T}_{t=\ceil{\sqrt{T}}+1}\E\left[(w^*-c)q(w^*;F_0) - ( w^*_{stat} - c)q( w^*_{stat}; F_0)\right]\\
 &\leq \ceil{\sqrt{T}}s\bar\xi + (T - \ceil{\sqrt{T}})s\bar\xi/\ceil{\sqrt{T}}\\
 &= O(\sqrt{T}),
        \end{split}
    \end{equation*}
where the first inequality follows from Eq.~\eqref{equ:supplier-profit-property} and the fact that $|w^* - w^*_{stat}|\leq s/\ceil{\sqrt{T}}$ by the discretization of $\set{W}$.

\section{Additional material for Section~\ref{sec:retailer-learning}}\label{appendix:learning}

\subsection{Proof of Lemma~\ref{lem:retailer-learning-wm}}
Suppose $q(w_m^t; \hat F_t^\mu)\geq y_m$, then it follows from Eq.~\eqref{equ:retailer-order-stationary} that $p_{t, m-1}\leq 1 - w_m^t/s$, and $\varphi(w_m^t; \hat F_t^\mu) \geq (w_m^t - c)y_m = \varphi_{k^*} + \Delta_t +\frac{y_m s}{K}$. The previous inequality follows from the fact that $q(w_m^t; \hat F_t^\mu)\geq y_m$ and the previous equality follows by construction of $w_m^t$ (see Eq.~\eqref{equ:luna-discrete-wm}).
The proof is similar for the case $q(w_m^t; \hat F_t^\mu) < y_m$.

\subsection{Proof of Lemma~\ref{lem:epoch-variation-1}}

We will show that if $q(w_m^t; \hat F_t^\mu)\geq y_m$, then 
\begin{equation}\label{equ:appendix-variation}
\max_{k\in[K]}\max_{x\in[0,\bar\xi]}|\hat F_t^\mu(x) - \hat F_{\tau^0_i + k}^\mu(x)|\geq \Delta_t/(s\,\bar\xi).
\end{equation}
It will then follow from the reverse triangle inequality that the total variation over epoch $i$ is:
\begin{equation*}
    \sum^{\tau^0_{i+1}-1}_{t=\tau^0_i +1}\max_{x\in[0,\bar\xi]}|\hat F_{t+1}^\mu(x) - \hat F_{t}^\mu (x)|\geq \Delta_t/(s \bar\xi).
\end{equation*}

Recall $(p_{\tau^0_i + k,m})_{k \in [K]}$ are the values of $\hat F_t^\mu$ at $y_m$ during the exploration phase in epoch $i$ (i.e., for periods $t\in[\tau^0_i+1, \tau^0_i + K]$). To show Eq.~\eqref{equ:appendix-variation}, we start with the following Lemma~\ref{lem:appendix-1}.
\begin{lemma}\label{lem:appendix-1}
    During the exploration phase of epoch $i$, we must have
\begin{equation}\label{equ:appendix-lemma-variation-1-1}
\max_{k\in[K]}p_{\tau^0_i + k,m-1} \geq 1 - \frac{\varphi_{k^*}/y_m + c}{s} - \frac{1}{K},
\end{equation}
for all $m\in[M]$ with $m\neq m^*$, where the term $\frac{1}{K}$ is due to the discretization error.
\end{lemma}
\proof{Proof of Lemma~\ref{lem:appendix-1}}
For a contradiction, suppose there exists some $m'\in[M]$ with $m'\neq m^*$ such that 
\begin{equation}\label{equ:appendix-contradiciton}
    s(1-p_{\tau^0_i+k, m'}) > \varphi_{k^*}/y_{m'} + c + \frac{s}{K},\,\forall k\in[K].
\end{equation}
We let $k'\in[K]$ be such that $\bar w_{k'}\in[s(1-p_{\tau^0_i + k',m'-1}) - \frac{s}{K}, s(1-p_{\tau^0_i + k',m'-1})]$ (notice that such $k'$ must exist by the discretization).  
We then have $\varphi(\bar w_{k'}; \hat F_{\tau^0_i + k'}^\mu) = (\bar w_{k'} - c)q(\bar w_{k'}; \hat F^\mu_{\tau^0_i + k'}) \geq (\bar w_{k'} - c)y_{m'} \geq (s - s\, p_{\tau^0_i + k,m'-1} - c - \frac{s}{K})y_{m'}> \varphi_{k^*}$ where the first inequality follows since $\bar w_{k'}\leq s(1 - p_{\tau^0_i + k', m' - 1})$ and then according to Eq.~\eqref{equ:retailer-learning-profit}, $q(\bar w_{k'}; \hat F^\mu_{\tau^0_i + k})\geq y_{m'}$. The second inequality follows since $\bar w_{k'} \geq s(1-p_{\tau^0_i + k',m'-1}) - \frac{s}{K}$ and the last inequality follows from Eq.~\eqref{equ:appendix-contradiciton}. Thus we have shown that Eq.~\eqref{equ:appendix-contradiciton} is a contradiction of the fact that $k^*\in\arg\max_{k\in[K]}\varphi_k$.\Halmos
\endproof

On the other hand, since $q(w_m^t; F_t)\geq y_m$ for some $m\in[M]$, we have
\begin{equation}\label{equ:appendix-lemma-variation-1-2}
    p_{t,m-1}\leq 1 - w_m^t/s = 1 - \frac{(\varphi_{k^*}+\Delta_t)/y_m + 1/K + c}{s},
\end{equation}
where the inquality follows from Lemma~\ref{lem:retailer-learning-wm} and the equality follows from Eq.~\eqref{equ:luna-discrete-wm}. Combining Eq.~\eqref{equ:appendix-lemma-variation-1-1} and Eq.~\eqref{equ:appendix-lemma-variation-1-2} gives:
\begin{equation*}
\max_{k\in[K]} (p_{\tau^0_i + k,m-1} - p_{t, m-1}) \geq \Delta_t/(s\, y_m)\geq \Delta_t/(s\bar\xi).
\end{equation*}
It follows that $\max_{k\in[K]}\max_{x\in[0,\bar\xi]}|\hat F_{\tau^i_0 + k}^\mu(x) - \hat F_t^\mu (x)|\geq \Delta_t/(s\, \bar\xi)$, and the Lemma holds.

\subsection{Proof of Lemma~\ref{lem:retailer-learning-w0}}
There are two cases: (i) $\bar w_{k^*} - \Delta_t/y_{m^*} \geq 0$; and (ii) $\bar w_{k^*} - \Delta_t/y_{m^*} < 0$.
In the first case where $\bar w_{k^*} - \Delta_t/y_{m^*} \geq 0$, we have $w^t_0 = \bar w_{k^*} - \Delta_t/y_{m^*}$ and the proof is similar to Lemma~\ref{lem:retailer-learning-wm}.
In the second case where $\bar w_{k^*} - \Delta_t/y_{m^*} < 0$, $w^t_0 = 0$ and we have $q(0; \hat F^\mu_t) = \bar\xi\geq y_{m^*}$ (the retailer will order as much as possible since the order cost is zero). In this case, we have $p_{t, m^* - 1} \leq 1- w^t_0/s = 1$ (by Eq.~\eqref{equ:retailer-order-stationary}) and $\varphi(w^t_0; \hat F^\mu_t) \geq (w^t_0 - c)y_{m^*} \geq (\bar w_{k^*} - \Delta_t/y_{m^*} - c)y_{m^*} = \varphi_{k^*} - \Delta_t$.


\subsection{Proof of Lemma~\ref{lem:epoch-variation-2}}
When $q(w_0^t; \hat F_t^\mu)< y_{m^*}$, we must have $w^t_0 > 0$. Otherwise, if  $w^t_0 = 0$, the retailer will always order $q(0; \hat F^\mu_t) =\bar\xi$ since the order cost is zero. Thus we can restrict to $\bar w_{k^*} - \Delta_t/y_{m^*} > 0$.

We will show that if $q(w_0^t; \hat F_t^\mu)< y_{m^*}$, then 
$$
\max_{k\in[K]}\max_{x\in[0,\bar\xi]}|\hat F_{\tau^0_i + k}^\mu(x) - \hat F_t^\mu(x)|\geq \Delta_t/(s\, \bar\xi),
$$
and consequently
\begin{equation*}
    \sum^{\tau^0_{i+1}-1}_{t=\tau^0_i + 1}\max_{x\in[0,\bar\xi]}|\hat F_t^\mu(x) - \hat F_{t+1}^\mu(x)|\geq \Delta_t/(s\, \bar\xi).
\end{equation*}
First notice that according to the policy implementation, we have (by the relation $y_{m^*} = q(\bar w_{k^*}; \hat F_{\tau^0_i + k^*}^\mu)$ and Eq.~\eqref{equ:retailer-learning-profit}) that:
\begin{equation}\label{equ:appendix-lemma-variation-2-1}
    p_{\tau^0_i + k^*, m^*-1}< 1- \bar w_{k^*}/s.
\end{equation}
Next, since $q(w_0^t; \hat F_t^\mu)< y_{m^*}$, we have
\begin{equation}\label{equ:appendix-lemma-variation-2-2}
p_{t, m^*-1} \geq 1 - w_0^t/s = 1 - (\bar w_{k^*} - \Delta_t/y_{m^*})/s = 1 - \bar w_{k^*}/s + \Delta_t/(s\, y_{m^*}),
\end{equation}
where the first inequality follows from Lemma~\ref{lem:retailer-learning-w0} and the first equality follows from Eq.~\eqref{equ:luna-discrete-w0}. 
Combining Eq.~\eqref{equ:appendix-lemma-variation-2-1} and Eq.~\eqref{equ:appendix-lemma-variation-2-2} gives
$$
p_{t,m^* -1} - p_{\tau^0_i+ k^*, m^* -1} \geq \Delta_t/(s\, y_{m^*})\geq \Delta_t/(s\, \bar\xi),
$$
and it follows that $\max_{k\in[K]}\max_{x\in[0,\bar\xi]}|\hat F_{\tau^i_0 + k}^\mu(x) - \hat F_t^\mu(x)|\geq \Delta_t/(s\, \bar\xi)$.

\subsection{Proof of Theorem~\ref{thm:retailer-learning}}
Here we complete the details of the proof of Theorem~\ref{thm:retailer-learning}. We use superscript $i$ to denote quantities corresponding to epoch $i$ since those quantities vary from epoch to epoch. That is, 
we use $\varphi^i_{k}$ to denote the profit observed during period $k \in [K]$ of the exploration phase of epoch $i$, and $\varphi^i_{k^*}$ to denote the optimal profit observed during the exploration phase in epoch $i$.

Abusing notation, let $\reg(\pi, \hat F^\mu_{t_1:t_2}) \triangleq \sum^{t_2}_{t=t_1}\varphi(w^*_t; \hat F_t^\mu) - \varphi(w_t; \hat F_t^\mu)$ be the regret incurred from periods $t \in [t_1,t_2]$ when the supplier follows $\pi$, given the sequence of perceived distributions $\hat F_{t_1:t_2}^\mu$.
The regret incurred in epoch $i$ (for periods $t\in[\tau^0_i + 1, \tau^0_{i+1}]$) for $\policy{LUNA}$ is then:
\begin{equation*}
        \reg\left(\policy{LUNA}, \hat F^\mu_{\tau^0_i + 1:\tau^0_{i+1}}\right) = \sum^{\tau^0_{i+1}}_{t=\tau^0_i + 1}\varphi(w^*_t; \hat F_t^\mu) - \varphi(w^t_0; \hat F_t^\mu) + \varphi(w^t_0; \hat F_t^\mu) - \varphi(w_t; \hat F_t^\mu).
\end{equation*}
We will bound the first part $\reg_i^c(\policy{LUNA}) \triangleq \sum^{\tau^0_{i+1}}_{t = \tau^0_i 
 + 1}\varphi(w^*_t; \hat F_t^\mu) - \varphi(w^t_0; \hat F_t^\mu)$ and the second part $\reg^0_i(\policy{LUNA}) \triangleq \sum^{\tau^0_{i+1}}_{t = \tau^0_i+1}\varphi(w^t_0; \hat F_t^\mu) - \varphi(w_t; \hat F_t^\mu)$ separately. 

\paragraph{Part I of the regret}
To bound the first part $\reg^c_i(\policy{LUNA})$, recall the set (for epoch $i$):
$$
\set{E}^i = \left\{t\in[\tau^0_i+\max\{M+2, K\} + 1, \tau^0_{i+1}]: q(w^t_m; \hat F_t^\mu) < y_m, \,\forall m\in[M], \text{ and } q(w^t_0; \hat F_t^\mu)\geq y_{m^*}\right\}.
$$
Then, we have
\begin{equation}\label{equ:appendix-retailer-learning-1}
\begin{split}
    &\quad\reg^c_i(\policy{LUNA})\\
    & \leq \sum^{\tau^0_{i+1}}_{t=\tau^0_i + \max\{M+2,K\}+1}\left(\varphi(w^*_t; \hat F_t^\mu) - \varphi(w^t_0; \hat F_t^\mu)\right) + s\bar\xi\max\{M+2, K\} \\
    &\leq
    \sum^{\tau^0_{i+1}}_{t = \tau^0_i+\max\{M+2,K\}+1}\left[\left(\varphi(w^*_t; \hat F_t^\mu) - \varphi(w^t_0; \hat F_t^\mu)\right)\1(t\notin \set{E}^i) + \left(\varphi(w^*_t; \hat F_t^\mu) - \varphi(w^t_0; \hat F_t^\mu)\right)\1(t\in \set{E}^i)\right]\\
&\quad\quad\quad\quad\quad\quad\quad\quad\quad\quad\quad\quad\quad\quad\quad\quad\quad\quad\quad\quad\quad\quad\quad\quad\quad\quad\quad\quad\quad\quad\quad
    +s\bar\xi\max\{M+2, K\}\\
    &\leq 
    \sum^{\tau^0_{i+1}}_{t = \tau^0_i+\max\{M+2,K\}+1}\left[s\bar\xi\1(t\notin \set{E}^i) + \left(\varphi(w^*_t; \hat F_t^\mu) - \varphi(w^t_0; \hat F_t^\mu)\right)\1(t\in \set{E}^i)\right]\\
&\quad\quad\quad\quad\quad\quad\quad\quad\quad\quad\quad\quad\quad\quad\quad\quad\quad\quad\quad\quad\quad\quad\quad\quad\quad\quad\quad\quad\quad\quad\quad
    +s\bar\xi\max\{M+2, K\}.
\end{split}
\end{equation}

It then follows from Eq.~\eqref{equ:appendix-retailer-learning-1}, Lemma~\ref{lem:retailer-learning-2}, and Lemma~\ref{lem:retailer-learning-Ei} that
\begin{equation}\label{equ:appendix-retailer-learning-4}
\begin{split}
    &\reg^c_i(\policy{LUNA})\\
    \leq \,&
    \sum^{\tau^0_{i+1}}_{t = \tau^0_i+\max\{M+2,K\}+1}\left[s\bar\xi\1(t\notin \set{E}^i) + \left(\varphi(w^*_t; \hat F_t^\mu) - \varphi(w^t_0; \hat F_t^\mu)\right)\1(t\in \set{E}^i)\right]
    +s\bar\xi\max\{M+2, K\}\\
    \leq \,&
    s\bar\xi\left[\max\{M+2,K\}+1+2\log(T)\sqrt{M(\tau^0_{i+1}-1 - \tau^0_i)}\right]\\
&\quad\quad\quad\quad\quad\quad\quad\quad\quad\quad\quad\quad\quad\quad\quad\quad\quad\quad\quad+
    \sum^{\tau^0_{i+1}}_{t = \tau^0_i+\max\{M+2,K\}+1}2\Delta_t+
    (\tau^0_{i+1}-1 - \tau^0_i) s\bar\xi \left(\frac{1}{K}\right)\\
    \leq\,&  
    s\bar\xi\left[M+K+3+2\log(T)\sqrt{M(\tau^0_{i+1}-1 - \tau^0_i)}\right] +
   4\sqrt{M(\tau^0_{i+1}-1-\tau^0_i)}+
    (\tau^0_{i+1}-1 - \tau^0_i) s\bar\xi \left(\frac{1}{K}\right)
\end{split}
\end{equation}
where the last inequality follows since
\begin{align*}
    \sum^{\tau^0_{i+1}}_{t = \tau^0_i+\max\{M+2,K\}+1}2\Delta_t = \sum^{\tau^0_{i+1}}_{t = \tau^0_i+\max\{M+2,K\}+1}2\sqrt{M/(t-\tau^0_i)}\leq 4\sqrt{M(\tau^0_{i+1}-1-\tau^0_i)}.
\end{align*}

\paragraph{Part II of the regret}
For the second part of the regret $\reg^0_i(\policy{LUNA})$, we have the equivalence
\begin{equation*}
\begin{split}
    \reg^0_i(\policy{LUNA}) & = \sum^{\tau^0_{i+1}}_{t=\tau^0_i+1}\varphi(w^t_0; \hat F_t^\mu) - \varphi(w_t; \hat F_t^\mu)\\
    & =\left(\varphi(w^t_0; \hat F_t^\mu) - \varphi(w_t; \hat F_t^\mu)\right)\1(w_t\neq w^t_0)\\
    &\leq \bar\xi s\left[K + T_i(K) \right].
\end{split}
\end{equation*}
Then, by Lemma~\ref{lem:retailer-learning-4} we have
\begin{equation}\label{equ:appendix-retailer-learning-5}
    \reg^0_i(\policy{LUNA}) \leq \bar\xi s\left[K + \sqrt{11\log(T)M(\tau^0_{i+1}-1 - \tau^0_i)} \right],
\end{equation}
with probability at least $1 - 1/T^2$.

\paragraph{Combining the two parts of the regret}

Combining Eqs.~\eqref{equ:appendix-retailer-learning-4} and \eqref{equ:appendix-retailer-learning-5}, and using the union bound, we have with probability at least $1-1/T$ that (where we let $I$ be the total number of epochs)

\begin{align}
    \begin{split}
    \label{equ:appendix-retailer-learning-7}
    &\quad\reg\left(\policy{LUNA}, \hat F_{1:T}^\mu \right)\\
        & = \sum^{T}_{t=1}\varphi(w^*_t; \hat F_t^\mu) - \varphi(w^t_0; \hat F_t^\mu) + \varphi(w^t_0; \hat F_t^\mu) - \varphi(w_t; \hat F_t^\mu)\\
        & = \sum^I_{i = 1} \reg^c_i(\policy{LUNA}) + \sum^I_{i = 1}\reg^0_i(\policy{LUNA})\\
        & \leq  s\bar\xi(M+2K+3)I+\left(2s\bar\xi\log(T)+4\right)\sum^I_{i=1}\sqrt{M(\tau^0_{i+1}-1 - \tau^0_i)}+
    \sum^I_{i=1}(\tau^0_{i+1}-1 - \tau^0_i) s\bar\xi\frac{1}{K} \\ &\quad\quad\quad\quad\quad\quad\quad\quad\quad\quad\quad\quad\quad\quad\quad\quad\quad\quad\quad\quad\quad\quad\quad\quad\quad\quad+\sum^I_{i=1}\bar\xi s\sqrt{11\log(T)M(\tau^0_{i+1}-1 - \tau^0_i)}\\
        & \leq s\bar\xi(M+2K+3)I+\left(2s\bar\xi\log(T) + 4 + \bar\xi\sqrt{11\log(T)}\right)\left(\sum^I_{i=1}\sqrt{M(\tau^0_{i+1} - 1-\tau^0_i)}\right) + \frac{T}{K}.
    \end{split}
\end{align}
It then follows that 
\begin{align}
    \begin{split}
    \label{equ:appendix-retailer-learning-final}
\reg(\policy{LUNA}, T) &= \sup_{\mu\in\set{M}(V, T)}
\E\left[\reg\left(\policy{LUNA}, \hat F_{1:T}^\mu\right)\right]\\
& \leq s\bar\xi(M+2K+3)I+\left(2s\bar\xi\log(T) + 4 + \bar\xi\sqrt{11\log(T)}\right)\left(\sum^I_{i=1}\sqrt{M(\tau^0_{i+1} - 1-\tau^0_i)}\right) + \frac{T}{K}\\
& \leq s\bar\xi(M+2K+3)I+\left(2s\bar\xi\log(T) + 4 + \bar\xi\sqrt{11\log(T)}\right)\sqrt{MTI} + \frac{T}{K}.\\
&= \Tilde{O}\left(\bar\xi^{\frac{4}{3}}V^{\frac{1}{3}}M^{\frac{1}{3}}T^{\frac{2}{3}} + \frac{\bar\xi T}{K} + \bar\xi^{\frac{5}{3}}KV^{\frac{2}{3}}M^{-\frac{1}{3}}T^{\frac{1}{3}}
\right),
    \end{split}
\end{align}
where the second inequality follows by Jensen's inequality (using $\sum^T_{i=1}(\tau^0_{i+1} - \tau^0_i - 1) = T$), and the last equality follows by Lemma~\ref{lem:retailer-learning-number-epoch}.
The rest of the argument follows by setting $K^* = \ceil{T^{\frac{1}{3}}V^{-\frac{1}{3}}\bar\xi^{-\frac{1}{3}}}$ and $\hat K = \ceil{\bar\xi^{-\frac{1}{3}}T^{\frac{1}{3}}}$.

\subsection{Proof of Lemma~\ref{lem:retailer-learning-2}}

Since $t\in \set{E}^i\cap [\tau^0_i + \max\{M+2,K\}+1,\tau^0_{i+1}]$, we have by Lemma~\ref{lem:retailer-learning-w0} that
$$
\varphi(w^t_0; \hat F_t^\mu) \geq (w^t_0 - c)y_{m^*}
 \geq \varphi^i_{k^*} - \Delta_t.
$$
At the same time, by Eq.~\eqref{equ:retailer-learning-profit}, the optimal supplier profit in period $t$ satisfies:
$$
\varphi(w^*_t;\hat F_t^\mu) = \max_{m\in[M-1]}(s-s\, p_{t,m}-c)y_{m+1}.
$$
We then have
\begin{align*}
\varphi(w^*_t; \hat F_t^\mu)
& = \max_{m\in[M-1]}(s - s\, p_{m,t} - c)y_{m+1}\\
& \leq \max_{m\in[M-1]}(w^t_{m+1} - c)y_{m+1}\\
& = \Delta_t + \frac{\bar\xi s}{K} +\varphi^i_{k^*},
\end{align*}
where the inequality follows from Lemma~\ref{lem:retailer-learning-wm}. Therefore, 
\begin{align*}
    \varphi(w^*_t; \hat F_t^\mu) - \varphi(w^t_0; \hat F_t^\mu) = \varphi(w^*_t; \hat F_t^\mu) -\varphi^i_{k^*} + \varphi^i_{k^*} -\varphi(w^t_0; \hat F_t^\mu) \leq 2\Delta_t + \frac{\bar\xi s}{K}.
\end{align*}

\subsection{Proof of Lemma~\ref{lem:retailer-learning-Ei}}

We make use of the following supporting result.

\begin{lemma}\label{lem:retailer-learning-3}
Let $(t_n)_{n=1}^N\in[\tau^0_i+\max\{M+2,K\}+1, \tau^0_{i+1}-1]$ for $N \geq 1$ be the sequence of periods where $t_n\in \set{E}^i$. Let
$$
s = \arg\min_{s' \geq 1}\left\{\sum^{s'}_{n=1}\frac{1}{\sqrt{M(t_n-\tau^0_i)}}\geq 2\log(T)\right\},
$$
then $E^i \leq s$ with probability at least $1-1/T^2$.
\end{lemma}
\noindent
The proof of Lemma~\ref{lem:retailer-learning-3} follows from Lemma~\ref{lem:retailer-learning-4} below and \citealp[Theorem A.4]{karnin2016multi}.
\begin{lemma}\label{lem:retailer-learning-4}
    If $t\notin \set{E}^i$ and $t\geq \tau^0_i + \max\{M+2, K\}+1$, then epoch $i$ ends in period $t$ with probability at least $\sqrt{\frac{1}{M(t-\tau^0_i)}}$ given that epoch $i$ has not ended before period $t$.
\end{lemma}

\proof{Proof of Lemma~\ref{lem:retailer-learning-4}}

If $t\notin \set{E}^i$, then either $q(w^t_m; \hat F^\mu_t)>y_m$ for some $m\in[M]$ holds, or $q(w^t_0; \hat F^\mu_t)<y_{m^*}$ holds. 
In the first case, where $q(w^t_m; \hat F^\mu_t)>y_m$ for some $m\in[M]$, epoch $i$ will end if $m_t = m$ which occurs with probability $\frac{1}{M}\sqrt{\frac{M}{t-\tau^0_i}} = \sqrt{\frac{1}{M(t-\tau^0_i)}}$. In other words, according to the algorithm implementation, with probability $\sqrt{\frac{1}{M(t-\tau^0_i)}}$ we choose $m_t = m$ and since $q(w^t_m; \hat F^\mu_t)>y_m$, we end the current epoch according to the pricing policy.
In the second case, if $q(w^t_0; \hat F^\mu_t)<y_{m^*}$, then epoch $i$ will end if $m_t = 0$ which occurs with probability $1-\sqrt{\frac{M}{t - \tau^0_i}}\geq \sqrt{\frac{1}{M(t-\tau^0_i)}}$. In either case, epoch $i$ will end with probability at least $\sqrt{\frac{1}{M(t-\tau^0_i)}}$ for all $t\geq \tau^0_i + \max\{M+2,K\}+1$ with $t\notin \set{E}^i$. \Halmos

\endproof

\proof{Proof of Lemma~\ref{lem:retailer-learning-Ei}}

It follows from Lemma~\ref{lem:retailer-learning-3} that with probability at least $1 - 1/T^2$, we have
$$
\frac{E^i - 1}{\sqrt{M(\tau^0_{i+1}-1 - \tau^0_i)}}\leq
\sum^{E^i-1}_{n=1}\frac{1}{\sqrt{M(t_n - \tau^0_i)}}
=
\sum^{s-1}_{n=1}\frac{1}{\sqrt{M(t_n - \tau^0_i)}}\leq 2\log(T),
$$
and so
\begin{equation*}
E^i\leq 2\log(T)\sqrt{M(\tau^0_{i+1}-1 - \tau^0_i)} + 1.
\end{equation*} \Halmos
\endproof

\subsection{Proof of Lemma~\ref{lem:retailer-learning-number-epoch}}

For $\policy{LUNA}$, when epoch $i$ ends in period $t = \tau^0_{i+1}$, we have $\Delta_t  = \Delta_{\tau^0_{i+1}} =  M^{\frac{1}{2}}(\tau^0_{i+1} - \tau^0_i - 1)^{-\frac{1}{2}}$ for every epoch $i\in[I-1]$.
It then follows that
\begin{equation*}
    \begin{split}
        V &= \sum^I_{i=1} \sum_{t = \tau^0_i+1}^{\tau^0_{i+1}-1}d_K(\hat F^\mu_{t}, \hat F^\mu_{t+1})  \geq \sum^{I-1}_{i=1} \sum_{t = \tau^0_i+1}^{\tau^0_{i+1}-1}d_K(\hat F^\mu_{t}, \hat F^\mu_{t+1}) \geq \sum^{I-1}_{i=1}\Delta_{\tau^0_{i+1}}/(s\, \bar\xi)\\
        &=\sum^{I-1}_{i=1} \frac{1}{s\, \bar\xi}M^{\frac{1}{2}}(\tau^0_{i+1} - \tau^0_i - 1)^{-\frac{1}{2}}\geq \frac{M^{\frac{1}{2}}}{s\, \bar\xi}(I-1)^{\frac{3}{2}}T^{-\frac{1}{2}}.
    \end{split}
\end{equation*}
In the above display, we drop the last epoch $I$ from the summation in the first inequality because we do not necessarily have $\sum_{t = \tau^0_I+1}^{\tau^0_{I+1}-1}d_K(\hat F^\mu_{t}, \hat F^\mu_{t+1}) \geq \Delta_{I}/(s\, \bar\xi)$, i.e., epoch $I$ does not necessarily end because $T\notin \set{E}^I$. 
The second inequality follows from Lemma~\ref{lem:epoch-variation-1} and Lemma~\ref{lem:epoch-variation-2}.
The last inequality follows since
\begin{align*}
    \sum^{I-1}_{i=1}(\tau^0_{i+1} - \tau^0_i - 1)^{-\frac{1}{2}}\geq (I-1)\left(\frac{1}{I-1}\sum^{I-1}_{i=1}(\tau^0_{i+1}-\tau^0_i - 1)\right)^{-\frac{1}{2}}\geq (I-1)\left(T/(I-1)\right)^{-\frac{1}{2}},
\end{align*}
where the first inequality follows from Jensen's inequality and the second follows since $\sum^{I-1}_{i=1}\tau^0_{i+1} - \tau^0_{i} - 1 \leq T$.
We conclude that $I \leq (s\, \bar\xi)^{\frac{2}{3}}V^{\frac{2}{3}}M^{-\frac{1}{3}}T^{\frac{1}{3}} + 1$.

\subsection{Proof of Lemma~\ref{lem:conti-approx}}

Define the mapping $f: \mathbb{R}\to \mathbb{R}$ by $f(q) = z_n$ for $q\in (z_{n-1}, z_n]$. When $q = 0$, we have $f(0) \triangleq 0$.

We make use of the following result.

\begin{lemma}\label{lem:retailer-learning-appendix-1}
    For all $t\in[T]$, $q(w_t; \tilde F^\mu_t) = f(q(w_t; \hat F_t^\mu))$.
\end{lemma}

\proof{Proof of Lemma~\ref{lem:retailer-learning-appendix-1}}

When $q(w_t; \hat F^\mu_t) = 0$,
we have $f(q(w_t; \hat F^\mu_t)) = 0$ and $\hat F^\mu_t(0) \geq 1 - w_t/s$ (by Eq.~\eqref{equ:retailer-best response}). 
Since $\tilde F^\mu_t(0) = \hat F^\mu_t(0)$ by Eq.~\eqref{equ:retailer-learning-discretization}, 
we have $\tilde F^\mu_t(0) \geq 1 - w_t/s$ and thus $q(w_t; \Tilde F^\mu_t) = 0$.

Now suppose $q(w_t; \hat F^\mu_t)\in (z_{n-1}, z_n]$ for some $n\geq 2$. Then, we have 
$\hat F^\mu_t(z_{n-1}) <1 - w_t/s$, $\hat F^\mu_t(z_n) \geq 1 - w_t/s$, and $f(q(w_t; \hat F^\mu_t)) = z_n$. By Eq.~\eqref{equ:retailer-learning-discretization}, $\tilde F^\mu_t(z_{n-1})< 1 - w_t/s$ and $\tilde F^\mu_t(z_n)\geq  1 - w_t/s$ both hold. Thus, $q(w_t; \tilde F^\mu_t) = z_n = f(q(w_t; \hat F^\mu_t)) = z_n$, and the claim holds. \Halmos

\endproof

\proof{Proof of Lemma~\ref{lem:conti-approx}}
The proof is by induction. The result clearly holds for period $t = 1$ since the first period wholesale price is fixed at $\bar w_1$ (recall $\policy{LUNAC}$ calls $\policy{LUNA}$ as a subroutine). Now suppose the claim holds up to some period $1\leq t< T$, we will prove that it holds for period $t+1$. For brevity, by the induction hypothesis we simply write $w_i \triangleq w^{\text{LUNAC}}_i(\hat F_{1:t-1}^\mu; \omega) = w^{\text{LUNA}}_i(\tilde F^\mu_{1:t-1}; \omega)$ for $i \in [t]$.

Fix the sample path $\omega$, the history of wholesale prices $(w_i)^t_{i=1}$, and perceived distributions $\hat F_{1:t-1}^\mu$. In $\policy{LUNAC}$, in each period the feedback $f(q(w_t; \hat F_t^\mu))$ is given to $\policy{LUNA}$ based on the actual order quantity $q(w_t; \hat F_t^\mu)$ (see Line 5 of Algorithm~\ref{alg:retailer-learning-discretization}). Then, the wholesale price $w^{\text{LUNAC}}_{t+1}(\hat F_{1:t}^\mu; \omega)$ output by $\policy{LUNAC}$ is the wholesale price output by $\policy{LUNA}$ given the past wholesale prices $(w_i)^t_{i=1}$ and feedback $(f(q(w_t; \hat F_{1:i}^\mu)))^t_{i=1}$. 

At the same time, according to the construction of $\tilde F^\mu_{1:t}$ (see Eq.~\eqref{equ:retailer-learning-discretization}), given any wholesale price $w_t$ we have $q(w_t; \tilde F^\mu_t) = f(q(w_t; \hat F_t^\mu))$ as shown in Lemma~\ref{lem:retailer-learning-appendix-1}.
In other words, in each period $t\in[T]$, $\policy{LUNA}$ receives the feedback $f(q(w_t; \hat F_t^\mu)) = q(w_t; \tilde F^\mu_t)$.
It then follows that $w^{\text{LUNAC}}_{t+1}(\hat F_{1:t}^\mu; \omega)$ is the price output by $\policy{LUNA}$ given past wholesale prices $(w_i)^t_{i=1}$ and orders $(q(w_t; \tilde F^\mu_t))^t_{i=1}$. Since $\policy{LUNA}$ will output $w^{\text{LUNA}}_{t+1}(\tilde F^\mu_{1:t}; \omega)$ given past wholesale prices $(w_i)^t_{i=1}$ and orders $(q(w_t; \tilde F^\mu_t))^t_{i=1}$, we have proved that $
w^{\text{LUNAC}}_{t+1}(\hat F_{1:t}^\mu; \omega) = w^{\text{LUNA}}_{t+1}(\tilde F^\mu_{1:t}; \omega)$, and the induction step holds. \Halmos

\endproof

\subsection{Proof of Lemma~\ref{lem:retailer-learning-discretize}}
By definition, we have
\begin{equation*}
    \begin{split}
        d_K(\tilde F^\mu_t, \Tilde F^\mu_{t+1}) = \max_{n\in[N]}|\tilde F^\mu_t(z_n) - \Tilde F^\mu_{t+1}(z_n)| = \max_{n\in[N]}|\hat F_t^\mu(z_n) - \hat F_{t+1}^\mu(z_n)|\leq d_K(\hat F_t^\mu, \hat F_{t+1}^\mu),
    \end{split}
\end{equation*}
where the first equality follows since $\tilde F^\mu_t$ is supported on $\set{Z}_{N}$, and the second equality follows by construction of $\tilde F^\mu_t$ in Eq.~\eqref{equ:retailer-learning-discretization}.

\subsection{Proof of Theorem~\ref{thm:retailer-learning-2}}
Recall $w^*_t$ defined in \eqref{equ:optimal-price-benchmark} is the optimal wholesale price in each period (regardless of whether the $\hat F^\mu_t$ is continuous or discrete).
We can decompose the regret as
\begin{equation}
\begin{split}
&\quad \reg(\policy{LUNAC}, \hat F^\mu_{1:T})\\
    &=\sum^T_{t=1}\E\left[\varphi(w^*_t; \hat F_t^\mu) - \varphi(w_t; \hat F_t^\mu)\right]\\
    & = \sum^T_{t=1}\E\left[(\varphi(w^*_t; \hat F_t^\mu) - \varphi(w^*_t; \tilde F^\mu_t)) - (\varphi(w_t; \hat F_t^\mu) - \varphi(w_t; \tilde F^\mu_t)) + (\varphi(w^*_t; \tilde F^\mu_t) - \varphi(w_t; \tilde F^\mu_t))\right]\\
    &\leq \sum^T_{t=1}\E\bigg[(\varphi(w^*_t; \hat F_t^\mu) - \varphi(w^*_t; \tilde F^\mu_t)) - (\varphi(w_t; \hat F_t^\mu) - \varphi(w_t; \tilde F^\mu_t)) + \Big(\sup_{w\in\set{W}}\varphi(w; \tilde F^\mu_t) - \varphi(w_t; \tilde F^\mu_t)\Big)\bigg].
\end{split}
\end{equation}
In the above display, both $\sum^T_{t=1}\E[\varphi(w^*_t; \hat F_t^\mu) - \varphi(w^*_t; \tilde F^\mu_t)]$ and $\sum^T_{t=1}\E\left[\varphi(w_t; \hat F_t^\mu) - \varphi(w_t; \tilde F^\mu_t)\right]$ represent the regret incurred by approximating $\hat F_t^\mu$ with $\tilde F^\mu_t$. According to Lemma~\ref{lem:conti-approx}, the wholesale prices $w_{1:T}$ output from $\policy{LUNAC}$ are just the pricing decisions of running the subroutine $\policy{LUNA}$ with distributions $\tilde F_{1:T}$. Thus the expression
$$
\sum^T_{t=1}\E\left[\sup_{w\in\set{W}}\varphi(w; \tilde F^\mu_t) - \varphi(w_t; \tilde F^\mu_t)\right]
$$
is the regret from running $\policy{LUNA}$ with respect to $\Tilde F_{1:T}$, i.e.,
$$
\sum^T_{t=1}\E\left[\sup_{w\in\set{W}}\varphi(w; \tilde F^\mu_t) - \varphi(w_t; \tilde F^\mu_t)\right] = \reg(\policy{LUNA}, T),
$$
see Theorem~\ref{thm:retailer-learning}. 

Based on the approximation of $\hat F_t^\mu$ with $\tilde F^\mu_t$, for any $w\in\set{W}$, we have
\begin{equation}\label{equ:retailer-learning-appendix-discretize}
    |\varphi(w, \hat F_t^\mu) - \varphi(w,\tilde F^\mu_t)| = (w -c)\left|\min\left\{q: \hat F^{\mu}_t \geq 1-w/s\right\} - \min\left\{q: \Tilde F^{\mu}_t \geq 1-w/s\right\}\right|\leq (w-c)\bar\xi/N,
\end{equation}
where the inequality follows from Eq.~\eqref{equ:retailer-learning-discretization}.
Then, we have
\begin{equation*}
    \begin{split}
&\quad\reg(\policy{LUNAC}, \hat F^\mu_{1:T})\\
& \leq \sum^T_{t=1}\E\bigg[(\varphi(w^*_t; \hat F_t^\mu) - \varphi(w^*_t; \tilde F^\mu_t)) - (\varphi(w_t; \hat F_t^\mu) - \varphi(w_t; \tilde F^\mu_t)) + \Big(\sup_{w\in\set{W}}\varphi(w; \tilde F^\mu_t) - \varphi(w_t; \tilde F^\mu_t)\Big)\bigg]\\
& \leq \sum^T_{t=1}\E\bigg[\big|\varphi(w^*_t; \hat F_t^\mu) + \varphi(w^*_t; \tilde F^\mu_t)\big| + \big|\varphi(w_t; \hat F_t^\mu) - \varphi(w_t; \tilde F^\mu_t)\big| \bigg] + \reg(\policy{LUNA}, T).
    \end{split}
\end{equation*}
With Eq.~\eqref{equ:retailer-learning-appendix-discretize} and Theorem~\ref{thm:retailer-learning}, if the supplier knows $V$, then 
\begin{equation*}
    \reg(\policy{LUNAC}, \hat F^\mu_{1:T}) = \Tilde{O}\left(\bar\xi\, T/N + \bar\xi^{\frac{4}{3}}V^{\frac{1}{3}} N^{\frac{1}{3}}T^{\frac{2}{3}} \right)
\end{equation*}
and if the supplier does not know $V$, then 
\begin{equation*}
    \reg(\policy{LUNAC}, \hat F^\mu_{1:T}) = \Tilde{O}\left(\bar\xi\, T/N + \bar\xi^{\frac{4}{3}}V^{\frac{1}{3}} N^{\frac{1}{3}}T^{\frac{2}{3}} + \bar\xi^{\frac{4}{3}}V^{\frac{2}{3}} N^{-\frac{1}{3}}T^{\frac{2}{3}}\right).
\end{equation*}

It follows that if the supplier knows $V$, then with $N^* = \ceil{\bar\xi^{-\frac{1}{4}}V^{-\frac{1}{4}}T^{\frac{1}{4}}}$, the regret is $\reg(\policy{LUNAC}, T) = \Tilde{O}(\bar\xi^{\frac{5}{4}}V^{\frac{1}{4}}T^{\frac{3}{4}})$.
If the supplier does not know $V$, then by choosing $\hat N = \ceil{\bar\xi^{-\frac{1}{4}}T^{\frac{1}{4}}}$ the regret is $\reg(\policy{LUNAC}, T) = \tilde O(\bar\xi^{\frac{5}{4}}V^{\frac{1}{3}}T^{\frac{3}{4}} + \bar\xi^{\frac{17}{12}}V^{\frac{2}{3}}T^{\frac{7}{12}})$.

\section{Additional material for Section~\ref{sec:retailer-learning-examples}}
\subsection{Proof of Proposition~\ref{prop:retailer-learning-saa}}
Recall both $\hat F_t^e = (\sum^{t-1}_{i=1}\1(\xi_i\leq x))/(t-1)$ and $\hat F_{t+1}^e = (\sum^t_{i=1}\1(\xi_i\leq x))/t$ are empirical distributions, so for any $x\in[0,\bar\xi]$ we have
\begin{equation}\label{equ:appendix-empirical}
    \left\vert\hat F_t^e(x) - \hat F_{t+1}^e(x)\right\vert
    \leq \left\vert \frac{\sum^{t-1}_{i=1}\1(\xi_i \leq x)}{t-1} -\frac{\sum^{t}_{i=1}\1(\xi_i \leq x)}{t} \right\vert= \begin{cases}
    &\frac{\sum^{t-1}_{i=1}\1(\xi_i = x)}{t(t-1)},\, \xi_t> x,\\
    & \frac{1}{t} - \frac{\sum^{t-1}_{i=1}\1(\xi_i = x)}{t(t-1)},\,\xi_t\leq x.
    \end{cases}
\end{equation}
It then follows that $d_K(\hat F_t^e, \hat F_{t+1}^e)\leq \frac{1}{t}$. Then $\sum^{T-1}_{t=1}d_K(\hat F^e_t, \hat F^e_{t+1})\leq \log(T) + 1$ and thus $\mu_e\in\set{M}(\log(T) + 1, T)$.

\subsection{Proof of Proposition~\ref{prop:retailer-learning-phi-divergence}}

(i) For any $t \in [T-1]$, we have
\begin{align*}
d_K(\hat F_t^{\text{d}}, \hat F_{t+1}^{\text{d}})&\leq d_{K}(\hat F_t^{\text{d}}, \hat F_t^{\text{e}}) + d_K(\hat F_{t+1}^{\text{e}}, \hat F_t^{\text{e}}) + d_K(\hat F_{t+1}^{\text{e}}, \hat F_{t+1}^{\text{d}})\\
&\leq \frac{1}{t} + \sqrt{d_{KL}(\hat F_t^{\text{d}}, \hat F_{t}^{\text{e}})/2} + \sqrt{d_{KL}(\hat F_{t+1}^{\text{e}}, \hat F_{t+1}^{\text{d}})/2}\\
&\leq \frac{1}{t} + \sqrt{\epsilon_t/2} + \sqrt{\epsilon_{t+1}/2},
\end{align*}
where the first inequality follows from triangle inequality, the
second inequality follows from \citet{gibbs2002choosing} (which states that $d_K(F,G)\leq \sqrt{d_{KL}(F,G)/2}$ for all $F,G\in\set{P}$ with $F\ll G$) and Eq.~\eqref{equ:appendix-empirical}.
Then, we have
$$
\sum^{T-1}_{t=1} d_K(\hat F^d_t, \hat F^d_{t+1}) \leq \sum^{T-1}_{t=1}\left(\frac{1}{t} + \sqrt{\epsilon_t/2} + \sqrt{\epsilon_{t+1}/2}\right)\leq \log{(T)} + 1 + \sum^T_{t=1}\sqrt{2\epsilon_t}.
$$

(ii) For any $t \in [T-1]$, we have
\begin{align*}
d_K(\hat F_t^{\text{d}}, \hat F_{t+1}^{\text{d}})&\leq d_{K}(\hat F_t^{\text{d}}, \hat F_t^{\text{e}}) + d_K(\hat F_{t+1}^{\text{e}}, \hat F_t^{\text{e}}) + d_K(\hat F_{t+1}^{\text{d}}, \hat F_{t+1}^{\text{e}})\\
&\leq \frac{1}{t} + \sqrt{d_{\chi^2}(\hat F_t^{\text{d}}, \hat F_t^{\text{e}})}/2 + \sqrt{d_{\chi^2}(\hat F_{t+1}^{\text{e}}, \hat F_{t+1}^{\text{d}})}/2\\
&\leq \frac{1}{t} + \sqrt{\epsilon_t}/2 + \sqrt{\epsilon_{t+1}}/2,
\end{align*}
where the second inequality follows from \citet{gibbs2002choosing}, (which states $d_K(F,G)\leq \sqrt{d_{\chi^2}(F,G)}/2$ for $F,G\in\set{P}$ with $F\ll G$).
Then, we have
$$
\sum^{T-1}_{t=1} d_K(\hat F^d_t, \hat F^d_{t+1}) \leq \sum^{T-1}_{t=1}\left(\frac{1}{t} + \sqrt{\epsilon_t}/2 + \sqrt{\epsilon_{t+1}}/2\right)\leq \log{(T)} + 1 + \sum^T_{t=1}\sqrt{\epsilon_t}.
$$

(iii) We have
\begin{align*}
d_K(\hat F_t^{\text{d}}, \hat F_{t+1}^{\text{d}})&\leq d_{K}(\hat F_t^{\text{d}}, \hat F_t^{\text{e}}) + d_K(\hat F_{t+1}^{\text{e}}, \hat F_t^{\text{e}}) + d_K(\hat F_{t+1}^{\text{d}}, \hat F_{t+1}^{\text{e}})\\
&\leq \frac{1}{t} + d_{H}(\hat F_t^{\text{d}}, \hat F_t^{\text{e}}) + d_{H}(\hat F_{t+1}^{\text{d}}, F_{t+1}^{\text{e}})\\
&\leq \frac{1}{t} + \epsilon_t + \epsilon_{t+1},
\end{align*}
where the second inequality follows from \citet{gibbs2002choosing} (which states $d_K(F,G)\leq d_H(F,G)$ for $F,G\in\set{P}$ with $F\ll G$).
Then, we have
$$
\sum^{T-1}_{t=1} d_K(\hat F^d_t, \hat F^d_{t+1})  \leq \sum^{T-1}_{t=1}\left(\frac{1}{t} + \epsilon_t + \epsilon_{t+1}\right)\leq \log{(T)} +1 + 2\sum^T_{t=1}\epsilon_t.
$$

\subsection{Proof of Proposition~\ref{prop:retailer-learning-exp-family}}
To prove Proposition~\ref{prop:retailer-learning-exp-family}, we relate the Kolmogorov distance and the total variation distance $d_{TV}$. For two probaiblity distributions $F,G\in\set{P}(\Xi)$ equipped with the $\sigma-$algebra $\set{F}$, the total variation distance $d_{TV}$ between $F$ and $G$ is defined by:
$$
d_{TV}(F,G) \triangleq \sup\left\{|F(A) - G(A)|: A\in\set{F}\right\}.
$$
According to \citet{gibbs2002choosing}, 
\begin{equation}\label{equ:retailer-learning-mle-tv}
d_{KL}(F,G)\leq d_{TV}(F,G).
\end{equation}

(i) According to \citet{adell2006exact}, the total variation between Poisson distributions $F_{\lambda_1}$ and $F_{\lambda_2}$ with means $\lambda_1$ and $\lambda_2$ respectively (we assume that $\lambda_1 \leq \lambda_2$) satisfies $d_{TV}\left(F_{\lambda_1}, F_{\lambda_2}\right) \leq |\lambda_2 - \lambda_1|$.
The MLE estimate of the mean of a Poisson distribution is
$\lambda_t = \frac{\sum^{t-1}_{i=1}\xi_i}{t-1}$, so it follows that
\begin{multline}
d_{TV}(\hat F_t^{\text{m}}, \hat F_{t+1}^{\text{m}})\leq d_{TV}(F_{\lambda_t}, F_{\lambda_{t+1}})\leq \left|\frac{\sum^t_{i=1}\xi_i}{t} - \frac{\sum^{t-1}_{i=1}\xi_i}{t-1}\right| = \left|\frac{t(\sum^{t-1}_{i=1}\xi_i) - (t-1)(\sum^{t-1}_{i=1}\xi_i + \xi_t)}{t(t-1)}\right|\\
\leq \frac{\max\left\{\sum^{t-1}_{i=1}\xi_i, (t-1)\xi_t\right\}}{t(t-1)},
\end{multline}
where the first inequality follows by recalling from Eq.~\eqref{equ:MLE-F-hat} that 
$$
\hat F_t^{\text{m}}(x) =
    \begin{cases}
        F_{\lambda_t}(x), \,& 0\leq x < \bar q;\\
    1, \,&x\geq \bar q.
    \end{cases}
$$

${\rm Poisson}(\lambda)$ distribution has the following concentration inequality:
\begin{equation}\label{equ:retailer-learning-mle-appendix-1}
\PP\left(\xi\geq \lambda + \epsilon\right)\leq \exp{\left(-\frac{\epsilon^2}{2(\lambda + \epsilon)}\right)}\text{ for }\epsilon > 0,
\end{equation}
and so $\PP\left(\xi \leq 4\ln(T) + 2\lambda\right)\geq 1 - 1/T^2$.
By the union bound, we then have 
$$
\PP\left(\xi_t \leq 4\ln(T) + 2\lambda,\,\forall t\in[T]\right) \geq 1 - \frac{1}{T}.
$$
It follows that
$$
d_{TV}(\hat F_t^{\text{m}}, \hat F_{t+1}^{\text{m}})\leq \frac{4\ln{(T)} + 2\lambda}{t}
$$
with probability at least $1 - 1/T$, and thus
\begin{align*}
    \sum^{T-1}_{t=1}d_K(\hat F_t^{\text{m}}, \hat F_{t+1}^{\text{m}})
    &\leq \sum^{T-1}_{t=1} d_{TV}(\hat F_t^{\text{m}}, \hat F_{t+1}^{\text{m}})\\
    &\leq \sum^{T-1}_{t=1}\frac{4\ln{(T)} + 2\lambda}{t}\\
    &\leq \left(4\ln{(T)} + 2\lambda\right)\sum^{T-1}_{t=1}\frac{1}{t}\\
    &\leq (\ln{(T)} + 1)\left(4\ln{(T)} + 2\lambda\right),
\end{align*}
where the first inequality follows from Eq.~\eqref{equ:retailer-learning-mle-tv}.

(ii) If the true demand distribution is the categorical distribution, and the retailer is using MLE, then $F_t = \hat F^e_t$, the empirical distribution at time $t$. The argument then follows similarly to Proposition~\ref{prop:retailer-learning-saa}.

(iii) 
Let $F_{\lambda}$ and $F_{\lambda'}$ be the CDFs of the ${\rm E}(\lambda)$ and ${\rm E}(\lambda')$ distributions, respectively, and suppose $\lambda<\lambda'$. Then we have
\begin{align*}
    d_{TV}(F_{\lambda},F_{\lambda'}) &= \frac{1}{2}\int^{\infty}_{x=0}|\lambda\exp(-\lambda x) - \lambda'\exp(-\lambda' x)|dx\\
    &=\left(\frac{\lambda}{\lambda'}\right)^{\frac{\lambda}{\lambda'-\lambda}} -\left(\frac{\lambda}{\lambda'}\right)^{\frac{\lambda'}{\lambda-\lambda'}}\\
    &\leq 1 - \frac{\lambda}{\lambda'}\\
    & = \min\{\lambda,\lambda'\}\left|\frac{1}{\lambda} - \frac{1}{\lambda'}\right|.
\end{align*}
The MLE estimator for the rate is $\lambda_t = \frac{t-1}{\sum^{t-1}_{i=1}\xi_i}$, and so
\begin{align*}
    d_{TV}(\hat F_t^{\text{m}}, \hat F_{t+1}^{\text{m}})\leq d_{TV}(F_{\lambda_t}, F_{\lambda_{t+1}})
    &\leq \lambda_{t}\left|\frac{\sum^{t-1}_{i=1}\xi_i}{t-1} - \frac{\sum^{t}_{i=1}\xi_i}{t}\right|\\
    &\leq \frac{t-1}{\sum^{t-1}_{i=1}\xi_i}\frac{\left|\sum^{t-1}_{i=1}\xi_i - t\xi_{t}\right|}{t(t-1)}\\
    &\leq \frac{1}{t} + \frac{t-1}{t}\frac{\xi_t}{\sum^{t-1}_{i=1}\xi_i}.
\end{align*}

Now according to the high probability bound for the ${\rm E}(\lambda)$ distribution, we have:
\begin{align*}
    \PP\left(\sum^t_{i=1}\xi_i \leq t/\lambda - \epsilon\right)\leq \exp\left(-\frac{\epsilon^2\lambda^2}{4t}\right),
\end{align*}
which gives
\begin{align*}
   \PP\left(\sum^t_{i=1}\xi_i \geq \frac{t}{\lambda} - \frac{\sqrt{4t\ln{(2T^2)}}}{\lambda}\right) \geq 1 - \frac{1}{2T^2}.
\end{align*}
On the other hand, according to the CDF of the exponential distribution, we have
$$
\PP\left(\xi\leq \frac{\ln{(2T^2)}}{\lambda}\right) \geq 1 - \frac{1}{2T^2}.
$$
By the union bound, we have 
\begin{equation}\label{equ:retailer-learning-parametric-exponential-1}
\PP\left(\sum^t_{i=1}\xi_i \geq \frac{t}{\lambda} - \frac{\sqrt{4t\ln{(2T^2)}}}{\lambda}\text{ and }\xi_{t}\leq \frac{\ln{(2T^2)}}{\lambda},\, \forall t\in[T-1]\right)\geq 1-\frac{1}{T}.
\end{equation}
It then follows that, with probability at least $1-1/T$, we have
\begin{align*}
    \sum^{T-1}_{t=1}d_K(\hat F_t^{\text{m}}, \hat F_{t+1}^{\text{m}})
    & = \sum^{16\ln(2T^2)-1}_{t=1}d_K(\hat F_t^{\text{m}}, \hat F_{t+1}^{\text{m}}) + \sum^{T-1}_{t=16\ln(2T^2)}d_K(\hat F_t^{\text{m}}, \hat F_{t+1}^{\text{m}})\\
    & \leq \sum^{16\ln(2T^2)-1}_{t=1}d_K(\hat F_t^{\text{m}}, \hat F_{t+1}^{\text{m}}) + \sum^{T-1}_{t=16\ln(2T^2)}\left(\frac{1}{t} + \frac{t-1}{t}\frac{\xi_{t}}{\sum^{t-1}_{i=1}\xi_i}\right)\\
    &\leq 16\ln(2T^2)-1 + \sum^{T-1}_{t=16\ln(2T^2)}\frac{1+2\ln(2T^2)}{t}\\
    &\leq 16\ln(2T^2)-1 + \sum^{T-1}_{t=1}\frac{1+2\ln(2T^2)}{t}\\
    & \leq 16\ln{(2T^2)}-1 + \left(1 +2\ln{(2T^2)}\right)(\ln{(T)}+1),
\end{align*}
where the second inequality follows from Eq.~\ref{equ:retailer-learning-parametric-exponential-1} (which states that if $t\geq 16\ln(2T^2+1)$, then
$\frac{\xi_{t}}{\sum^{t-1}_{i=1}\xi_i}\leq \frac{2\ln(2T^2)}{t}$ for all $t\geq 2$ with probability at least $1-1/T$).



(iv) Let $F_{\mu}$ and $F_{\mu'}$ be the CDFs of the ${\rm N}(\mu, \sigma^2)$ and ${\rm N}(\mu', \sigma^2)$ distributions, respectively (they have the same variance and possibly different means).
The KL-divergence between $F_{\mu}$ and $F_{\mu'}$ is 
$$
d_{KL}(F_{\mu},F_{\mu'}) = \frac{(\mu - \mu')^2}{2\sigma^2}.
$$
The MLE estimator for the mean is
\begin{align*}
    \mu_t = \frac{\sum^{t-1}_{i=1}\xi_i}{t-1}, \, t\geq 2.
\end{align*}
Thus, we have
\begin{align*}
    d_{KL}(F_{\mu_t}, F_{\mu_{t+1}}) &= \frac{1}{2\sigma^2}\left(\frac{\sum^{t-1}_{i=1}\xi_i}{t-1} - \frac{\sum^{t}_{i=1}\xi_i}{t}\right)^2\\
    &\leq \frac{1}{2\sigma^2}\left(\frac{\sum^{t-1}_{i=1}\xi_i - (t-1)\xi_{t}}{t(t-1)}\right)^2\\
    &\leq \frac{1}{2\sigma^2}\left[\left(\frac{\sum^{t-1}_{i=1}\xi_i}{t(t-1)}\right)^2 + \left(\frac{\xi_{t}}{t}\right)^2\right],\,t\geq 2.
\end{align*}
On one hand, when $\xi \sim {\rm Normal}(\mu, \sigma^2)$, we have
\begin{align*}
    \PP(\xi\geq \mu + \epsilon) \leq \exp{\left(-\epsilon^2/2\sigma^2\right)},
\end{align*}
which gives
\begin{align*}
    \PP\left(\xi\leq \sigma\sqrt{2\log(2T^2)} + \mu\right) \geq 1-1/(2T^2).
\end{align*}
On the other hand, according to Hoeffding's inequality,
\begin{equation}\label{equ:retailer-learning-appendix-normal}
    \PP\left(\sum^t_{i=1}\xi_i - t\mu\geq \epsilon  \right)\leq \exp\left(-\frac{\epsilon^2}{2t\sigma^2}\right).
\end{equation}
It then follows that 
\begin{align*}
    \PP\left(\sum^t_{i=1}\xi_i \leq t\mu + \sigma\sqrt{2t\ln{(2T^2)}}\right) \geq 1 - 1/(2T^2).
\end{align*}
Using the union bound, we have
\begin{equation}\label{equ:normal-concentration}
    \PP\left(
\sum^{t-1}_{i=1}\xi_i \leq (t-1)\mu + \sigma\sqrt{2(t-1)\ln{(2T^2)}} \text{ and }\xi_{t} \leq \sigma\sqrt{2\ln(2T^2)} + \mu,\,\forall t\geq 2
\right)\geq 1-\frac{1}{T}.
\end{equation}
We then have with probability at least $1-1/T$, for $t\geq 2$,
\begin{align*}
    d_{KL}(F_{\mu_t}, F_{\mu_{t+1}}) & \leq \frac{1}{2\sigma^2}\left[\left(\frac{\sum^{t-1}_{i=1}\xi_i}{t(t-1)}\right)^2 + \left(\frac{\xi_{t}}{t}\right)^2\right]\\
    &\leq \frac{1}{2\sigma^2}\left[\left(\frac{(t-1)\mu + \sigma\sqrt{2(t-1)\ln{(2T^2)}}}{t(t-1)}\right)^2 + \left(\frac{\mu + \sigma\sqrt{2\ln{(2T^2)}}}{t}\right)^2\right]\\
    &\leq \frac{1}{\sigma^2}\left[\frac{2\mu^2}{t^2} + \frac{2\sigma^2\ln{(2T^2)}}{t(t-1)^2} +  \frac{2\sigma^2\ln{(2T^2)}}{t^2}\right]\\
    &\leq \frac{2\mu^2 + 4\sigma^2\ln{(2T^2)}}{\sigma^2}\frac{1}{(t-1)^2}
\end{align*}
where the second inequality follows by the high probability bound Eq.~\eqref{equ:normal-concentration}. 
The third inequality follows by the inequality $(x+y)^2\leq 2x^2 + 2y^2$ for arbitrary number $x, y$.
Consequently,
\begin{align*}
    \sum^{T-1}_{t=1}d_K(\hat F_t^{\text{m}}, \hat F_{t+1}^{\text{m}}) &\leq 1 + \sum^{T-1}_{t=2}d_{K}(F_{\mu_t}, F_{\mu_{t+1}})\\
    &\leq 1 + \sum^{T-1}_{t=2}\sqrt{d_{KL}(F_{\mu_t}, F_{\mu_{t+1}})/2}\\
    &\leq 1 + \sum^{T-1}_{t=2}\frac{1}{t-1}\sqrt{\frac{\mu^2 + 2\sigma^2\ln{(2T^2)}}{\sigma^2}}\\
    &\leq 1 + \frac{1}{\sigma}\sqrt{(\ln(T)+1)\left(\mu^2 + 2\sigma^2\ln{(2T^2)}\right)}.
\end{align*}





\subsection{Proof of Proposition~\ref{prop:retailer-learning-os}}
Let $F_{\lambda}$ and $F_{\lambda'}$ be the CDFs of the ${\rm E}(\lambda)$ and ${\rm E}(\lambda')$ distributions, respectively, and suppose $\lambda < \lambda'$. The total variation between $F_\lambda$ and $F_{\lambda'}$ is
\begin{align*}
    d_{TV}(F_\lambda,F_{\lambda'}) &= \frac{1}{2}\int^{\infty}_{x=0}|\lambda\exp(-\lambda\, x) - \lambda'\exp(-\lambda' x)|dx\\
    &=\left(\frac{\lambda}{\lambda'}\right)^{\frac{\lambda}{\lambda'-\lambda}} -\left(\frac{\lambda}{\lambda'}\right)^{\frac{\lambda'}{\lambda-\lambda'}}\\
    &\leq 1 - \frac{\lambda}{\lambda'}.
\end{align*}
For all $t \in [T]$, define the function $f_t$ such that $f_t(w_t)\triangleq \frac{(t-1)\left((s/w_t)^{\frac{1}{t}}-1\right)}{\ln(s/w_t)}$. Then, we have 
$1/\lambda_t = f_t(w_t)\left(\sum^{t-1}_{i=1}\xi_i\right)/(t-1)$ and $f_t(w_t)\in [(t-1)/t,1)$ for all $w_t\in\set{W}$.
It follows that

\begin{align*}
    d_{K}(\hat F_t^{\text{o}}, \hat F_{t+1}^{\text{o}})
    \leq d_{TV}(\hat F_t^{\text{o}}, \hat F_{t+1}^{\text{o}})
    \leq d_{TV}(F_{\lambda_t}, F_{\lambda_{t+1}})
    \leq 1 - \frac{\min\{\lambda_t,\lambda_{t+1}\}}{\max\{\lambda_t,\lambda_{t+1}\}}\leq \min\{\lambda_t,\lambda_{t+1}\}\left|\frac{1}{\lambda_t}-\frac{1}{\lambda_{t+1}}\right|.
\end{align*}
Next, note that for $t\geq 2$, 
\begin{align*}
    \left|\frac{1}{\lambda_t} - \frac{1}{\lambda_{t+1}}\right| &= \left|f_t(w_t)\frac{\sum^{t-1}_{i=1}\xi_i}{t-1} - f_{t+1}(w_{t+1})\frac{\sum^{t}_{i=1}\xi_i}{t}\right|\\
    &= \left|\frac{tf_t(w_t)\sum^{t-1}_{i=1}\xi_i - (t-1)f_{t+1}(w_{t+1})\sum^{t}_{i=1}\xi_i}{t(t-1)}\right|\\
    &= \left|\frac{(t-1)(f_t(w_t) - f_{t+1}(w_{t+1}))\left(\sum^{t-1}_{i=1}\xi_i\right) + f_t(w_t)\left(\sum^{t-1}_{i=1}\xi_i\right) - (t-1)f_{t+1}(w_{t+1})\xi_{t}}{t(t-1)}\right|\\
    &\leq \frac{\left|(t-1)(f_t(w_t) - f_{t+1}(w_{t+1}))\left(\sum^{t-1}_{i=1}\xi_i\right)\right| + \left|f_t(w_t)\left(\sum^{t-1}_{i=1}\xi_i\right)\right| + \left|(t-1)f_{t+1}(w_{t+1})\xi_{t}\right|}{t(t-1)}
    \\
    &\leq \frac{2t-1}{(t-1)t^2}\left(\sum^{t-1}_{i=1}\xi_i\right) + \frac{\xi_{t}}{t},
\end{align*}
where the last inequality follows since $f_t(w_t)\in[(t-1)/t,1)$ for all $w_t\in\set{W}$.
Clearly $\min\{\lambda_t,\lambda_{t+1}\} \leq  \lambda_{t} = \frac{t-1}{f_{t}(w_{t})\left(\sum^{t-1}_{i=1}\xi_i\right)}$ and thus for $t\geq$ 2, 
\begin{align*}
    d_{K}(\hat F_t^{\text{o}}, \hat F_{t+1}^{\text{o}}) &\leq
    \lambda_t\left|\frac{1}{\lambda_t} - \frac{1}{\lambda_{t+1}}\right|\\
    &\leq 
    \frac{t-1}{f_{t}(w_{t})\left(\sum^{t-1}_{i=1}\xi_i\right)} \left(\frac{2t-1}{(t-1)t^2}\left(\sum^{t-1}_{i=1}\xi_i\right) + \frac{\xi_t}{t}\right)\\
    &\leq \frac{2t-1}{(t-1)t} + \frac{\xi_t}{\sum^{t-1}_{i=1}\xi_i}\\
    &\leq \frac{2}{t-1} + \frac{\xi_{t}}{\sum^{t-1}_{i=1}\xi_i},
\end{align*}
where the third inequality again follows because $f_t(w_t)\in[(t-1)/t,1)$ for all $w_t\in\set{W}$.
Since $(\xi_i)^T_{i=1}$ are i.i.d. ${\rm Exponential}(\lambda)$,
by Eq.~\eqref{equ:retailer-learning-parametric-exponential-1} we have
$$
\PP\left(\sum^t_{i=1}\xi_i \geq \frac{t}{\lambda} - \frac{\sqrt{4t\ln{(2T^2)}}}{\lambda}\text{ and }\xi_{t+1}\leq \frac{\ln{(2T^2)}}{\lambda},\, \forall t\in[T-1]\right)\geq 1-\frac{1}{T}.
$$
Thus, with probability at least $1 - 1/T$, we have 
\begin{align*}
    \sum^{T-1}_{t=1}d_K(\hat F_t^{\text{o}}, \hat F_{t+1}^{\text{o}})
    & \leq 1 + \sum^{T-1}_{t=2}\frac{2}{t-1} + 16\ln{(2T^2)} + \sum^{T-1}_{t=2}\frac{2\ln{(2T^2)}}{t-1}\\
    &\leq 1 + 2\ln{T} + 2 + 16\ln{(2T^2)} +2\ln{(2T^2)}(\ln{T} + 1)\\
    &\leq 21 + 40\ln{(T)} + 4(\ln{(T)})^2.
\end{align*}



\section{Additional material for Section~\ref{sec:retailer-learning-discrete}}\label{appendix:LUNAC-N}

\subsection{LUNAC-N}

When the supplier does not know $V$, we show that we can further improve the regret bound by adopting the Bandit-over-Bandit (BOB) framework proposed by \citet{cheung2019learning,cheung2021hedging} to sequentially adjust the approximation size $N$ (we call this algorithm $\policy{LUNAC-N}$).  $\pi_{\text{LUNAC-N}}$ divides the time horizon into $\ceil{T/H}$ blocks (indexed by $i$) of equal length $H$. Inside block $i$, the discretization size $N_i$ is chosen from a finite set $\set{J}\subset[H]$. Based on the chosen $N_i$ for each block, we run $\policy{LUNAC}$ for that block. After receiving the profits from each block, the algorithm sequentially adjusts the approximation size $N_i$ from block to block.

The choice of $N_i$ in each block is chosen according to the EXP3 algorithm \citep{auer2002nonstochastic} designed for the adversarial bandit. In this way, the overall procedure consists of a meta algorithm for choosing $N_i$ in each block according to the profits collected from each block, and a sub-algorithm that is run inside each block, based on the chosen $N_i$ for that block. Theorem~\ref{thm:retailer-learning-M-BOB} presents the regret bound when the supplier does not have knowledge of $V$ and $N_i$ is chosen according to $\pi_{\text{LUNAC-N}}$.

\begin{theorem}\label{thm:retailer-learning-M-BOB}
Suppose Assumption~\ref{ass:retailer-learning-3} holds, the supplier does not have knowledge of $V$, and $\{N_i\}$ are chosen according to $\policy{LUNAC-N}$. Then, $\reg(\policy{LUNAC-N}, T) = \Tilde{O}(\bar\xi^{\frac{4}{3}}V^{\frac{3}{4}}T^{\frac{1}{3}} + \bar\xi^{\frac{5}{4}}V^{\frac{1}{3}}T^{\frac{3}{4}})$.
\end{theorem}

Algorithm~\ref{alg:retailer-learning-LUNAC-N} presents the implementation details for $\policy{LUNAC-N}$ for sequentially adjusting $N$ when $V$ is unknown. We initialize the EXP3 parameters as:
\begin{equation}\label{equ:appendix-BOB-initialize-2}
    \gamma = \min\left\{1,\sqrt{\frac{(z+1)\ln{(z+1)}}{(e-1)\ceil{T/H}}}\right\},\, s_{j,1} = 1,\quad \forall j = 0,1,\ldots, z.
\end{equation}

\begin{algorithm}[H]
	Input: Time horizon $T$, production cost $c$, and selling price $s$;\\
	Initialize $H\leftarrow \floor{\bar\xi^{-\frac{1}{4}}T^{\frac{1}{4}}}, z\leftarrow \ceil{\ln{H}}, \set{J}\leftarrow \left\{H^0, \floor{H^{1/z}},\ldots,H\right\}$;\\
	Set $\gamma$ and $(s_{j,1})^z_{j=0}$ according to Eq.~\eqref{equ:appendix-BOB-initialize-2};\\
	\For{$i = 1, 2, \ldots, \ceil{T/H}$}{
	Define distributions $(\alpha_{j,i})^z_{j=0}$ as:
	\begin{equation*}
	    \alpha_{j,i} = (1-\gamma)\frac{s_{j,i}}{\sum^z_{u=0}s_{u,i}} + \frac{\gamma}{z+1}, \,\forall j=0,\ldots, z;
	\end{equation*}
	 Choose $j_i\leftarrow j$ with probability $\alpha_{j,i}$ and set $N_i\leftarrow \floor{H^{j_i/z}}$;\\
	 \For{
	 $t = (i-1)H+1, \ldots, (i\ldots H)\wedge T$
	 }{
	 Run LUNAC-N with $N_i$;
	 }
	 $\sum^{(iH)\wedge T}_{t=(i-1)H+1}\varphi(w_t;F_t)$ is the profit collected during $t\in[(i-1)H+1, (i\cdot H)\wedge T]$;\\
	 Update $s_{j, i+1}$ as:
	 \begin{align*}
	 &s_{j_i, i+1} \leftarrow s_{j_i,i}\exp{\left(\frac{\gamma}{(z+1)\alpha_{j_i,i}}\left(\frac{1}{2} + \frac{1}{2}\frac{\sum^{(iH)\wedge T}_{t=(i-1)H+1}\varphi(w_t;F_t)}{((iH)\wedge T- (i-1)H)(s-c)\bar\xi}\right)\right)},\\
	 &s_{j,i+1} \leftarrow s_{j,i},\,\text{ if }j\neq j_i.
	 \end{align*}
	}
	\caption{LUNAC-N}
	\label{alg:retailer-learning-LUNAC-N}
\end{algorithm}

\proof{Proof of Theorem~\ref{thm:retailer-learning-M-BOB}}
Let $N^\dagger$ be the optimally tuned approximation size and $w^\dagger_t$ be the corresponding wholesale price of $\policy{LUNA}$ when the approximation size satisfies $N = N^\dagger$. Notice that since each block has at most $H$ rounds, we do not necessarily have $N^\dagger = N^*$ (where $N^*$ is the optimally chosen discretization size given the supplier knows $V$, see Theorem~\ref{thm:retailer-learning-2}) since we need $N^\dagger\leq H$. The regret of running BOB on top of $\policy{LUNAC}$ can be decomposed as:
\begin{align*}
    \reg\left(\pi_{\text{LUNA-K}};\hat F_{1:T}^\mu\right) = &\sum^T_{t=1}\varphi(w_t^*; \hat F^\mu_t) - \varphi(w_t; \hat F^\mu_t)\\
    = &\sum^T_{t=1}\varphi(w_t^*; \hat F^\mu_t) -\varphi(w_t^\dagger; \hat F^\mu_t) +\varphi(w_t^\dagger; \hat F^\mu_t) -\varphi(w_t; \hat F^\mu_t)\\
    = & \sum^{\ceil{T/H}}_{i=1}\sum^{i\cdot H\wedge T}_{t = (i-1)H+1}\left\{\left(\varphi(w_t^*; \hat F^\mu_t) -\varphi(w_t^\dagger; \hat F^\mu_t)\right) +\left(\varphi(w_t^\dagger; \hat F^\mu_t) -\varphi(w_t; \hat F^\mu_t)\right)\right\}.
\end{align*}
where $\sum^{\ceil{T/H}}_{i=1}\sum^{i\cdot H\wedge T}_{t = (i-1)H+1}\left(\varphi(w_t^*; \hat F^\mu_t) -\varphi(w_t^\dagger; \hat F^\mu_t)\right)$ is the regret incurred by always discretizing at $N^\dagger$ and $\sum^{\ceil{T/H}}_{i=1}\sum^{i\cdot H\wedge T}_{t = (i-1)H+1}\left(\varphi(w_t^\dagger; \hat F^\mu_t) -\varphi(w_t; \hat F^\mu_t)\right)$ is the regret of learning $N^\dagger$. 

Let $V(i)$ be the variation in block $i$:
$$
V(i) \triangleq \sum^{(i\cdot H)\wedge T-1}_{t = (i-1)H+1} d_K(\hat F^\mu_t, \hat F^\mu_{t+1}).
$$
Then, the regret incurred by always discretizing at $N^\dagger$ can be upper bounded with:
\begin{equation*}
    \begin{split}
    &\quad\sup_{\hat F_{1:T}^\mu\in\set{M}(V, T)}\quad\sum^{\ceil{T/H}}_{i=1}\sum^{i\cdot H\wedge T}_{t = (i-1)H+1}\E\left[\varphi(w_t^*; \hat F^\mu_t) -\varphi(w_t^\dagger; \hat F^\mu_t)\right]\\
    &= \sum^{\ceil{T/H}}_{i=1}\Tilde{O}\left(\bar\xi\, H/N^\dagger + \bar\xi^{\frac{4}{3}}V(i)^{\frac{1}{3}} N^{\dagger\frac{1}{3}}H^{\frac{2}{3}} + \bar\xi^{\frac{4}{3}}V(i)^{\frac{2}{3}} N^{\dagger-\frac{1}{3}}H^{\frac{2}{3}}\right)\\
    &=\Tilde{O}\left(\bar\xi T/N^\dagger + \bar\xi^{\frac{4}{3}}V^{\frac{1}{3}} N^{\dagger\frac{1}{3}}T^{\frac{2}{3}} + \bar\xi^{\frac{4}{3}}V^{\frac{2}{3}} N^{\dagger-\frac{1}{3}}T^{\frac{1}{3}} H^{\frac{1}{3}}\right),
    \end{split}
\end{equation*}
where the first equality follows from Theorem~\ref{thm:retailer-learning-2} and the second equality follows from Holder's inequality. 

The regret of learning $N^\dagger$ follows directly from the regret of running EXP3. Since the number of blocks for EXP3 is $\ceil{T/H}$, the number of possible values for $K$ is $|\set{J}|$ and the maximum regret in each block is $(s-c)\bar\xi H$, we have
\begin{align*}
    &\,\sup_{\mu\in\set{M}(V, T)}\sum^{\ceil{T/H}}_{i=1}\sum^{i\cdot H\wedge T}_{t = (i-1)H+1}\E\left[\varphi(w_t^\dagger; F_t) -\varphi(w_t; F_t)\right]\\
    =&\,\Tilde{O}\left(\bar\xi H\sqrt{\frac{|\set{J}|T}{H}}\right)\\
    =&\,\Tilde{O}\left(\bar\xi\sqrt{|\set{J}|TH}\right).
\end{align*}
Combining these bounds, we get
\begin{align*}
    \reg(\policy{LUNAC}, T) = \Tilde{O}\left(
    \bar\xi T/N^\dagger + \bar\xi^{\frac{4}{3}}V^{\frac{1}{3}} N^{\dagger\frac{1}{3}}T^{\frac{2}{3}} + \bar\xi^{\frac{4}{3}}V^{\frac{2}{3}} N^{\dagger-\frac{1}{3}}T^{\frac{1}{3}}H^{\frac{1}{3}} + \bar\xi\sqrt{|\set{J}|TH}
    \right).
\end{align*}
Following \citet{cheung2019learning,cheung2021hedging}, we consider the set $\set{J}$ for possible choices of $N$: 
$$
\set{J} = \{H^0, \floor{H^\frac{1}{z}}, \floor{H^\frac{2}{z}}, \ldots, H\}
$$
where $z$ is some positive integer. Since the choice of $H$ cannot depend on $V$, we can set $H = \bar\xi^\epsilon T^\alpha$ for some $\alpha\in(0,1)$. We now discuss two cases depending on whether $N^* \geq H$ or not. 

\underline{Case 1}: $N^* \leq H$, then $V> T^{1-4\alpha}\bar\xi^{-1-4\epsilon}$. In this case, $N^\dagger$ can automatically adapt to the largest element in $\set{J}$ that is smaller than $N^*$ (i.e., $N^*H^{-\frac{1}{z}}\leq N^\dagger\leq N^* H^{\frac{1}{z}}$), and thus
\begin{equation}\label{equ:retailer-learning-K-BOB-appendix1}
    \begin{split}
    \reg(\policy{LUNA-N}, T) &= \Tilde{O}\left(
    \bar\xi T/(N^* H^{-\frac{1}{z}}) + \bar\xi^{\frac{4}{3}}V^{\frac{1}{3}} (N^* H^{\frac{1}{z}})^{\frac{1}{3}}T^{\frac{2}{3}} + \bar\xi^{\frac{4}{3}}V^{\frac{2}{3}} (N^* H^{-\frac{1}{z}})^{-\frac{1}{3}}T^{\frac{1}{3}}H^{\frac{1}{3}} + \bar\xi\sqrt{|\set{J}|TH}
    \right)\\
    &= \Tilde{O}\left(
    \bar\xi^{\frac{5}{4}}V^{\frac{1}{4}}T^{\frac{3}{4}} H^{\frac{1}{z}} + \bar\xi^{\frac{5}{4}}T^{\frac{3}{4}}V^{\frac{1}{4}}H^{\frac{1}{3z}} + \bar\xi^{\frac{17}{12}+ \frac{\epsilon}{3}}H^{1/3z}V^{\frac{3}{4}}T^{\frac{1}{4} + \frac{\alpha}{3}} + \bar\xi^{1 + \frac{\epsilon}{2}}z^{\frac{1}{2}}T^{\frac{\alpha+1}{2}}
    \right).
    \end{split}
\end{equation}

\underline{Case 2}: $K^* \geq H$, then $V\leq T^{1-4\alpha}\bar\xi^{-1-4\epsilon}$. In this case, $M^\dagger = H$, and thus
\begin{equation}\label{equ:retailer-learning-K-BOB-appendix2}
    \begin{split}
    \reg(\policy{LUNA-N}, T) &= \Tilde{O}\left(
    \bar\xi T/H + \bar\xi^{\frac{4}{3}}V^{\frac{1}{3}} H^{\frac{1}{3}}T^{\frac{2}{3}} + \bar\xi^{\frac{4}{3}}V^{\frac{2}{3}} H^{-\frac{1}{3}}T^{\frac{1}{3}}H^{\frac{1}{3}} + \bar\xi\sqrt{|\set{J}|TH}
    \right)\\
    & = \Tilde{O}\left(
    \bar\xi^{1 - \alpha}T^{1 - \alpha} + \bar\xi^{\frac{4+\epsilon}{3}} V^{\frac{1}{3}}T^{\frac{2 + \alpha}{3}} + \bar\xi^{\frac{4}{3}}V^{\frac{2}{3}}T^{\frac{1}{3}} + z^{\frac{1}{2}}\bar\xi^{1+\frac{\epsilon}{2}}T^{\frac{1+\alpha}{2}}
    \right).
    \end{split}
\end{equation}
According to Eqs.~\eqref{equ:retailer-learning-K-BOB-appendix1} and \eqref{equ:retailer-learning-K-BOB-appendix2}, we can set $z = \floor{\ln{H}}$, $\epsilon = -\frac{1}{4}$ and $\alpha = \frac{1}{4}$ and the regret is 
\begin{align*}
    \reg(\policy{LUNA-N}, T) = \Tilde{O}\left(\bar\xi^{\frac{4}{3}}V^{\frac{3}{4}}T^{\frac{1}{3}} + \bar\xi^{\frac{5}{4}}V^{\frac{1}{3}}T^{\frac{3}{4}}\right).
\end{align*}
\Halmos
\endproof

\section{Additional materials for Section~\ref{sec:numerical}}
\label{appendix:LUNAF}
\subsection{Finite Decision Set}

We can modify $\policy{LUNA}$ to handle finite $\set{W}$, and we call this modified algorithm $\policy{LUNAF}$ (see the details in Algorithm~\ref{alg:retailer-learning-finite-w}).
Let $d\triangleq |\set{W}|$ be the number of admissible wholesale prices so $\set{W} = \{w_j\}^d_{j=1}$ where WLOG we assume $w_1 < w_2 < \cdots < w_d$. 
Let $\text{ceil}_{\set{S}}(x)$ be the smallest element in a set $\set{S}$ that is greater than or equal to $x$ and $\text{floor}_{\set{S}}(x)$ be the largest element in $\set{S}$ that is less than or equal to $x$. When $x\in\set{S}$, then $\text{ceil}_{\set{S}}(x)= \text{floor}_{\set{S}}(x) = x$. In the exploration phase of $\policy{LUNAF}$, the policy simply prices at each price in $\set{W}$. Let $j^*$ be the index of the optimal wholesale price $w_j^* \in \set{W}$. Then, in each period in the exploitation phase, $w^t_m$ for $m\in[M]$ and $w^t_0$ are computed according to
\begin{equation}\label{equ:luna-discrete-wm-appendix}
    (w^t_m - c)y_m = \varphi_{j^*} + \Delta_t + (w_{j^* + 1} - w_{j^*})y_{m^*}, \text{ which gives }w^t_m \triangleq \left(\varphi_{j^*} + (w_{j^*+1} - w_{j^*})y_{m^*} +  \Delta_t\right)/y_m + c,
\end{equation}
and
\begin{equation}\label{equ:luna-discrete-w0-appendix}
     (w_0 - c)y_{j^*} = \varphi_{j^*} - \Delta_t \text{ and }w^t_0 \geq 0, \text{ otherwise }w^0_t =0,
\end{equation}
which gives
\begin{equation*}
    w^t_0\triangleq \max\{w_{j^*} - \Delta_t/y_{j^*}, 0\}.
\end{equation*}
However, since we do not necessarily have $w^t_{m_t}\in\set{W}$ for $m_t\in[M]$, we need to project $w^t_{m_t}$ to $\set{W}$. In this case, the dynamic regret follows as
$$
\reg(\pi, T) \triangleq \max_{\mu \in\set{M}(V, T)}\E\left[\sum^T_{t=1} \left( \max_{w\in\set{W}}\varphi(w; \hat F^\mu_t) - (w_t - c)q(w_t; \hat F^\mu_t) \right) \right],
$$
where the clairvoyant benchmark optimizes over the prices in the finite admissable set $\set{W}$. We compare this regret for different algorithms numerically.

\begin{algorithm}[H]
	Input: Time horizon $T$, $c$, $s$, and admissible decisions $\set{W}$;\\
	Update current time $t\leftarrow 1$;\\
	Set epoch $i \leftarrow 1$ and $\tau^0_1\leftarrow 0$;\\
	\For{epoch $i = 1, 2, \ldots$}{
	\textbf{Exploration:}\\
	   Price at $w_j\in\set{W}, j\in[d]$ and observe $\varphi_j$ for the first $d$ periods in epoch $i$; \\
	   Let $j^*\in\arg\max_{j\in[J]}\varphi_j$ and $m^*$ be such that $y_{m^*} = q\left(w_{j^*}; F_{\tau^0_i + j^*}\right)$.\\
	   \textbf{Exploitation:}\\
	   In period $t$, set $\Delta_t \leftarrow \sqrt{M/(t-\tau^0_i)}$;\\
	   Compute prices $w_m$ for $m\in[M]$ and $w_0$ according to Eq.~\eqref{equ:luna-discrete-wm} and Eq.~\eqref{equ:luna-discrete-w0}, respectively;\\
	   Select $m_t$ according to the distribution
	   $$
        m_t = \begin{cases}
        0, &\text{ w.p. } 1 - \sqrt{\frac{M}{t-\tau^0_i}},\\
        \set{U}\{1,\ldots, M\}, &\text{ w.p. } \sqrt{\frac{M}{t-\tau^0_i}};
        \end{cases}
	   $$\\
	   \uIf{$m_t\geq 1$}{
    Set wholesale price at $w_t\leftarrow \text{ceil}_{\set{W}}(w_m)$ \;
  }
  \Else{
    Set wholesale price at $w_t\leftarrow \text{floor}_{\set{W}}(w_0)$ \;
  }
	   Observe retailer's order $q(w_t; F_t)$;\\
	   \If{
	   $q(w_{m_t}; F_t)\geq y_{m_t}$ for $m_t\in[M]$ or $q(w_{m_t}; F_t)< y_{m^*}$ for $m_t = 0$
	   }{
	   $\tau^0_{i+1}\leftarrow t$ and start the next epoch $i\leftarrow i+1$.
	   }
	}
	\caption{\textbf{L}earning \textbf{U}nder \textbf{N}on-stationary \textbf{A}gent with \textbf{F}inite decision set (LUNAF)}
	\label{alg:retailer-learning-finite-w}
\end{algorithm}

\end{document}